\documentclass[10pt]{article}
\usepackage{arxiv}
\usepackage[T1]{fontenc}
\usepackage[utf8]{inputenc}
\usepackage{amsmath,amsfonts,amssymb,amsthm}
\usepackage{bm,enumerate,graphicx,microtype,placeins}
\usepackage{algorithm,algorithmic}
\usepackage{etoolbox}
\usepackage{chapterbib}
\usepackage{xcolor}
\definecolor{LinkBlue}{RGB}{25,65,155}
\usepackage[colorlinks=true,citecolor=LinkBlue,linkcolor=LinkBlue,urlcolor=LinkBlue,
            bookmarksopen=true,bookmarksnumbered=true,pdfencoding=auto]{hyperref}
\usepackage[capitalise,nameinlink]{cleveref}
\DeclareGraphicsExtensions{.pdf,.png,.jpg}
\renewcommand{\shorttitle}{\footnotesize\scshape Uncertainty in mechanism--AI coupled dynamical systems}
\renewcommand{\headeright}{}
\date{}

\numberwithin{equation}{section}
\numberwithin{algorithm}{section}
\newtheorem{theorem}{Theorem}[section]

\newtheorem{proposition}[theorem]{Proposition}
\newtheorem{definition}[theorem]{Definition}

\theoremstyle{remark}
\newtheorem{remark}[theorem]{Remark}

\crefname{hypothesis}{Hypothesis}{Hypotheses}
\crefname{fact}{Fact}{Facts}
\newcommand{\MSCcodes}[1]{\par\noindent\textbf{MSC codes:} #1\par}
\newcommand{\email}[1]{\nolinkurl{#1}}
\newcommand{\funding}[1]{\textbf{Funding:} #1}
\newcommand{\bibprefix}{}
\CitationPrefix{\bibprefix}
\makeatletter
\patchcmd{\thebibliography}{\@biblabel{#1}}{\@biblabel{\bibprefix99}}{}{}
\AtBeginEnvironment{thebibliography}{\small}
\def\@cite#1#2{\textcolor{LinkBlue}{[{#1\if@tempswa , #2\fi}]}}
\makeatother
\newcommand{\StartSupplement}{%
  \clearpage
  \renewcommand{\bibprefix}{SM}%
  \CitationPrefix{\bibprefix}%
  \setcounter{section}{0}%
  \setcounter{equation}{0}%
  \setcounter{theorem}{0}%
  \setcounter{algorithm}{0}%
  \setcounter{figure}{0}%
  \setcounter{table}{0}%
  \renewcommand{\thesection}{SM\arabic{section}}%
  \renewcommand{\theHsection}{SM.\arabic{section}}%
  \renewcommand{\theHequation}{SM.\arabic{section}.\arabic{equation}}%
  \renewcommand{\theHtheorem}{SM.\arabic{section}.\arabic{theorem}}%
  \renewcommand{\theHalgorithm}{SM.\arabic{section}.\arabic{algorithm}}%
  \renewcommand{\thefigure}{SM\arabic{figure}}%
  \renewcommand{\thetable}{SM\arabic{table}}%
  \renewcommand{\shorttitle}{\footnotesize\scshape Supplementary materials: Distribution learning}%
}
\newcommand{\SupplementTitle}{%
  \pdfbookmark[0]{Supplementary Materials}{supplement-title}%
  \begin{center}
    {\Large\bfseries Supplementary Materials\par}
    \vspace{0.7em}
    {\large\bfseries Confronting Uncertainty in Mechanism-AI Coupled Dynamical Systems via Optimal Transport and Information Geometry: Part I. Distribution Learning\par}
    \vspace{0.7em}
    Pengfei Song\quad Yanni Xiao\quad Shuangshuang Yin
  \end{center}\vspace{0.5em}
}
\title{Confronting Uncertainty in Mechanism-AI Coupled Dynamical Systems via Optimal Transport and Information Geometry: Part I. Distribution Learning\thanks{\funding{Research partially supported  by
			the National Key R\&D Program of China (2025YFA1016502 (P.S.,Y.X.), 2022YFA1003704 (P.S.,Y.X.)), the National Natural Science Foundation of China (12220101001 (Y.X.), 12571544 (P.S., S.Y.)), and the China Postdoctoral Science Foundation (No. 2023T160504 (P.S.)), the Guangdong Provincial Key Laboratory of Mathematical and Neural Dynamical Systems (2024B1212010004 (P.S.)), the Shaanxi Institute of Fundamental Science Project (No. 23JSQ028 (P.S.)), and the Xi'an Jiaotong University Basic Research Operating Expenses (No. xyz012023045 (P.S.)).}}}
\author{Pengfei Song\thanks{School of Mathematics and Statistics, Xi'an Jiaotong University, Xi'an, Shaanxi, 710049, China. (\email{song921012@xjtu.edu.cn, yxiao@xjtu.edu.cn, yinss2014@stu.xjtu.edu.cn})}
\qquad Yanni Xiao\footnotemark[2]\qquad Shuangshuang Yin\footnotemark[2]}
\hypersetup{pdftitle={Confronting Uncertainty in Mechanism-AI Coupled Dynamical Systems via Optimal Transport and Information Geometry: Part I. Distribution Learning},pdfauthor={Pengfei Song, Yanni Xiao, Shuangshuang Yin}}
\begin{document}
\maketitle

\begin{abstract}
Mechanism-AI coupled dynamical systems are increasingly used to study complex biological systems. However, uncertainty associated with noisy, sparse, and partially observed data limits their reliability. This paper presents a systematic mathematical formulation for treating these uncertainties from the perspective of optimal transport and information geometry. We formulate the uncertainty quantification problem as the minimization of an energy functional on a probability density space equipped with a Riemannian metric, with the loss evaluated through a differential equation, and give a unified description of the gradient flows of  the energy functional under specific metrics, such as the Wasserstein and Fisher-Rao metrics. Furthermore, we present corresponding numerical methods for solving these gradient flows using abstract adjoint sensitivity analysis. The formulation connects a range of Bayesian inference methods for dynamical systems; numerical experiments compare posterior approximation and predictive uncertainty under a common computational budget. Moreover, it also provides a geometric perspective on dynamical modeling and the study of heterogeneity evolution.

\end{abstract}


\MSCcodes{92B05,58D25,62F15,68T07,49Q22}

\section{Introduction}

Mathematical modeling of dynamical systems in biology, exemplified by infectious disease dynamics and ecological dynamics, has long served as a fundamental tool for deciphering complex biological mechanisms and enabling precise intervention (\cite{murray2002mathematical,heesterbeek2015modeling}). While traditional mechanistic models, typically expressed as systems of differential equations, have provided valuable insights, they face significant challenges in precise modeling and efficient computation due to unknown mechanisms in biological modelling and accurate specification of model structures (\cite{metcalf2017opportunities}). To address these limitations, the integration of artificial intelligence with mechanistic modeling has recently gained prominence as a promising approach (\cite{kraemer2025artificial,wong2023leveraging}). 
This coupled framework (mechanism and data dual driven, AI for Science) combines the interpretability of mechanistic models with the flexible pattern recognition capabilities of deep learning (\cite{weinan2021dawning}). A particularly representative formulation in this domain is the Universal Differential Equations (UDEs) framework, which is mathematically represented as an initial value problem (\cite{rackauckas2020universal,yin2023optimal}):
\begin{equation}\label{eq0}
	x' = f_{\eta}(x, t, \mathrm{UniversalApprox}_{\phi}(x, t)),
\end{equation}
where \( f \) encapsulates prior mechanistic knowledge (e.g., from physics or biology) and \( \mathrm{UniversalApprox} \) is some universal AI learner (e.g., neural networks (NN)~\cite{lecun2015deep}, Kolmogorov-Arnold Networks (KAN)~\cite{liu2024kan}, Gaussian processes (GP)~\cite{rasmussen2003gaussian}) that can learn unknown mechanisms. 
By preserving known mechanisms, UDEs enhance interpretability and generalization compared to purely data-driven models, offering a powerful methodology for biological system identification (\cite{rackauckas2020universal,yin2023optimal}).

However, the application of Mechanism-AI coupled framework such as UDEs to biological dynamical systems confronts significant challenges, primarily stemming from the inherent heterogeneity and poor quality of experimental data. Biological data are characteristically partial, sparse, and contaminated by substantial measurement noise. Observations are often limited to a subset of system variables, collected at irregular and infrequent time points, and obscured by technical artifacts. Furthermore, constraints related to data privacy and acquisition costs further exacerbate these issues, leading to severely information-limited scenarios. It is precisely this data-induced uncertainty that represents the most critical bottleneck, severely restricting the reliability and generalizability of UDEs (\cite{wang2025discovering}). While the combination of a flexible neural component and a mechanistic model is powerful, it also becomes highly susceptible to these data-induced uncertainties. The learning process can easily converge to biologically implausible solutions or exhibit pathological sensitivity to initial conditions and noise realizations, as the neural network may overfit to artifacts or compensate for noise in unobservable dimensions. Consequently, without a principled framework to explicitly characterize and manage this uncertainty, the Mechanism-AI coupled paradigm risks generating models that are neither predictive nor trustworthy, ultimately limiting their translational potential in critical biological applications.

The challenge of uncertainty quantification in complex models has traditionally been addressed through several established methodological frameworks, each with distinct characteristics and limitations. Markov Chain Monte Carlo (MCMC) methods represent a gold standard for asymptotic accuracy (\cite{ma2015complete,brooks2011handbook,girolami2011riemann}). Techniques such as Hamiltonian Monte Carlo (HMC) (\cite{bou2018geometric,betancourt2017conceptual}) and its adaptive variant, the No-U-Turn Sampler (NUTS) (\cite{hoffman2014no}), leverage geometric information to efficiently explore the posterior distribution but require repeated gradients and Metropolis-Hastings corrections, making them computationally demanding for large-scale or stochastic models. Langevin-based methods use gradients and stochastic perturbations to explore the posterior. Stochastic gradient MCMC (SGMCMC) variants (\cite{ma2015complete}), including Stochastic Gradient Langevin Dynamics (SGLD) (\cite{welling2011bayesian}), reduce the gradient cost through minibatches, whereas the Metropolis-Adjusted Langevin Algorithm (MALA) (\cite{xifara2014langevin}) uses a gradient-informed proposal with a Metropolis correction. Their performance can be limited by slow mixing, step-size sensitivity, and complex correlation structures in high dimensions. On a different track, Variational Inference (VI) recasts the inference problem as an optimization, seeking the best approximation from a tractable family of distributions (\cite{zhang2018advances,blei2017variational}). While computationally efficient, classical mean-field VI can underestimate posterior uncertainty because of its factorization assumption. More advanced Particle-Based VI (ParVI) (\cite{wang2021particle,liu2019understanding,frank2009particle}) and Stein Variational Gradient Descent (Stein VI) methods mitigate this by using interacting particles to represent the posterior, but they can be sensitive to kernel choice and particle initialization, and their performance degrades when the variational family fails to capture the true posterior geometry (\cite{liu2016stein,liu2017stein,duncan2023geometry}). The field of Simulation-Based Inference (SBI) (\cite{martin2024approximating,lueckmann2021benchmarking,sisson2018handbook,cranmer2020frontier}), including Approximate Bayesian Computation (ABC) (\cite{beaumont2010approximate}) and Bayesian Synthetic Likelihood (BSL) (\cite{price2018bayesian}), bypasses the need for explicit likelihood evaluations, making them suitable for complex simulator-based models. Nonetheless, they typically require an enormous number of simulations to achieve reasonable accuracy, especially in high-dimensional parameter spaces, and their performance is heavily reliant on the choice of summary statistics. Filtering methods, such as particle filters and the ensemble Kalman filter (EnKF), are designed for sequential state estimation in dynamical systems (\cite{chopin2020introduction,sarkka2023bayesian,del2004feynman}). Although particle filters are consistent under suitable assumptions as the particle count increases, finite ensembles suffer from the well-known "curse of dimensionality" and weight degeneracy problem in high-dimensional state spaces (\cite{arulampalam2002tutorial}). The EnKF is more scalable but makes Gaussian assumptions that are often violated in biological systems with nonlinear dynamics (\cite{evensen2003ensemble}). A fundamental limitation common to many of these approaches, particularly in high dimensions, is their inefficient navigation of the parameter space. This inefficiency largely stems from the difficulty of using the underlying geometric structure efficiently, including the curvature of the loss landscape or posterior distribution. Without explicitly accounting for this geometry, the sampling or optimization paths can be highly suboptimal, leading to slow convergence and poor uncertainty estimates in the complex, high-dimensional problems typical of biological dynamical systems.

In response to these limitations, a class of geometric methods grounded in Optimal Transport (OT) and Information Geometry has emerged, reframing the optimization of a loss function \(\mathcal{L}(\theta)\) as the minimization of an energy functional \(\mathcal{F}(\rho)\) defined over a space of probability measures \(\rho\) (\cite{jordan1998variational,chen2023gradient,frogner2020approximate,nurbekyan2023efficient,li2024differential,chen2025accelerating,trillos2023optimization,chen2024efficient,santambrogio2017euclidean,mielke2023introduction,ambrosio2007gradient,ambrosio2005gradient}). The solution to this variational problem is conceptualized as a gradient flow on a Riemannian manifold, where the choice of metric fundamentally dictates the trajectory and properties of the flow. The Wasserstein gradient flow, induced by the Wasserstein metric, is particularly notable for its ability to capture the underlying geometry of the sample space. It corresponds to the continuity equation and is intimately connected to gradient descent in the space of measures. Its strengths include a physical interpretation as an evolution governed by a potential and, for entropy-regularized energies, a balance between drift and diffusion (\cite{santambrogio2017euclidean,villani2003topics,villani2008optimal,santambrogio2015optimal,ambrosio2012user,figalli2021invitation,ambrosio2021lectures,ambrosio2007gradient,ambrosio2005gradient}). The Fisher-Rao gradient flow, defined by the Fisher-Rao metric (or the Fisher information matrix), acts on probability densities and induces the Fisher information metric on parametric families (\cite{chen2025accelerating,trillos2023optimization,chen2024efficient,amari2016information,ay2017information}). It is known for its efficiency in natural gradient descent, effectively pre-conditioning the optimization with local curvature information. The Stein gradient flow offers a non-parametric alternative by leveraging the Stein identity and kernelized Stein discrepancy. It allows for flexible particle-based evolution without requiring explicit knowledge of the normalization constant of the target distribution (\cite{liu2016stein,liu2017stein,duncan2023geometry}).  To combine the strengths of different metrics, coupled approaches have been developed. The Kalman-Wasserstein gradient flow incorporates ideas from ensemble Kalman methods into the Wasserstein framework, offering improved efficiency for certain filtering and inference problems with nearly-Gaussian structure (\cite{garbuno2020interacting}). The Wasserstein-Fisher-Rao (WFR) geometry, associated with unbalanced optimal transport, unifies the Wasserstein and Fisher-Rao geometries, allowing for simultaneous transport and creation/destruction of mass (\cite{gallouet2017jko,zhu2024inclusive,liero2018optimal,sun2025variational,chizat2018unbalanced}). This makes it particularly suited for problems where distributions do not have the same total mass or for modeling birth-death processes, yet its mathematical and numerical complexity is significantly increased.

While these geometric flows provide a powerful and unifying perspective on sampling and optimization, representing a significant theoretical advance, their application has remained largely generic. Their use in dynamical systems requires an explicit treatment of the forward solution map, the observation operator, and parameter sensitivities. Further opportunities arise from structures such as the symplectic geometry of Hamiltonian systems, Lie group configuration spaces, and parameter-induced pullback geometry. Their implementation in biological dynamical systems therefore calls for careful choices of geometry and numerical approximation. It remains useful to investigate how the rich geometric structure of dynamical systems—such as invariant manifolds, conserved quantities, and the geometry of the flow map—can be used to inform and constrain these gradient flows, thereby developing more efficient and physically consistent inference algorithms tailored to biological dynamical systems. These questions motivate a formulation that relates probability evolution to the underlying forward model.

To address this problem, we present a methodological framework that systematically transforms uncertainty quantification in Mechanism-AI coupled biological dynamical systems into a dynamical system-constrained optimization on a density manifold grounded in optimal transport and information geometry. The core of our approach is to rigorously formulate the uncertainty in the coupled system, particularly pertaining to the unknown parameters  of the Mechanism-AI coupled biological dynamical systems, as a distribution \(\rho\) evolving on a density manifold with some Riemannian metric. The formulation keeps the forward dynamical system explicit when defining the energy on probability densities. We define an energy functional \(\mathcal{F}(\rho)\) that incorporates both the data mismatch (e.g., via a log-likelihood or misfit term) and the physical constraint encoded by the UDE. This transforms the inference problem into finding the path of the steepest descent for \(\mathcal{F}(\rho)\) on the Riemannian manifold of probability measures, where the metric (such as Wasserstein, Fisher-Rao, or a coupled metric such as WFR) intrinsically defines the geometry of the path. The resulting dynamical system-constrained gradient flow, often taking the form of a partial differential equation like a Fokker-Planck-type equation or its particle-system equivalent, provides a coherent and principled description of how the belief state over plausible models evolves and concentrates. We subsequently develop computationally tractable schemes to solve this constrained flow. This involves deriving interacting particle systems or deterministic solvers for the probability flow equation, where the forward trajectory associated with each parameter particle satisfies the underlying dynamical equations. The solution map makes explicit how mechanistic laws enter the distributional learning problem and the gradient calculations. This provides a common description of posterior exploration and uncertainty estimation for sparse, noisy, and heterogeneous observations; accuracy and computational efficiency depend on the chosen geometry, discretization, and observation model.

\textbf{Summary.} Unlike purely data-driven AI applications, biological dynamical systems are inherently characterized by profound data heterogeneity (e.g., sparse, noisy, partially observed), and often incomplete observations that arise from complex experimental and environmental conditions. This heterogeneity makes uncertainty quantification not merely an optional enhancement but a fundamental necessity for reliable modeling, especially in the emerging paradigm of mechanism-AI integration. Historically, the lack of principled and computationally tractable frameworks has hindered rigorous uncertainty propagation in such coupled systems. However, the recent rise of geometric methods grounded in optimal transport and information geometry offers a promising pathway to bridge this gap. In this work, we strive to synthesize and systematize key prior developments, providing a self-contained exposition that makes the theory of dynamical system-constrained density gradient flows accessible and cohesive. We believe the proposed framework can:
\begin{itemize}
	\item Provide geometric methods for calibrating or performing Bayesian inference in traditional mechanistic models (without embedded AI components), where posterior exploration can be computationally demanding;
	
	\item Render uncertainty quantification for mechanism-AI coupled models both explicit in its mathematical assumptions and amenable to computation, while its efficiency must be evaluated for each application;
	
	\item Offer novel modeling perspectives for biological dynamical systems—for instance, in capturing trait evolution within epidemiological models (such as the dynamics of susceptibility and infectivity)—through geometric gradient flows (see Section~\ref{sec:heterogeneity_flow}).
\end{itemize}

\textbf{Focus and Scope of This Paper}

While the classification of uncertainty quantification challenges includes data uncertainty, model uncertainty and parametric uncertainty (see Appendix SM1), this work focuses on a basic scenario that can be extended to more complex settings. Our primary emphasis is on:

Deterministic dynamical systems with data noise and deterministic parameters (Bayesian posterior inference).
Specifically, we consider models of the form:
\[
\frac{\mathrm{d}x}{\mathrm{d}t} = f(x, \theta), \quad x(0) = x_0, \quad \theta \in \Theta \subset \mathbb{R}^d,
\]
with a specified observation model, for example additive Gaussian noise:
\[
y = h(x) + \varepsilon, \quad \varepsilon \sim \mathcal{N}(0, \Sigma),
\]
where the parameters \(\theta\) are deterministic but unknown. The inference task is to characterize the posterior distribution \(\rho(\theta \mid \mathcal{D})\) given data \(\mathcal{D} = \{y_i\}_{i=1}^M\).
\noindent\emph{Why this focus?}
This setting is common in practical biological modeling applications, where mechanistic knowledge often provides a deterministic skeleton (e.g., mass-action kinetics, ecological Lotka-Volterra equations), while observational limitations introduce inevitable measurement noise. Moreover, it provides a natural entry point for introducing geometric gradient flow methodology, as the mathematical structure is clean enough to reveal fundamental insights without being obscured by additional stochastic complexities.

Intrinsic model stochasticity (e.g., SDEs or CTMCs) often shifts the primary challenge to intractable likelihood evaluation rather than geometric posterior exploration; the gradient-flow framework naturally extends to such settings once likelihood approximations are incorporated. Intrinsically random parameters require essentially the same geometric tools—only the interpretation and forward map differ. While UDEs address model misspecification via neural components, we treat the network weights simply as additional deterministic parameters, deferring epistemic uncertainty about architecture choice.

Our aim is to build intuition for \emph{geometric gradient flows in density space} without the additional complications of stochastic simulators or doubly stochastic parameters. This focused scope allows us to present a self-contained, pedagogically effective exposition of dynamical system-constrained density gradient flows. 

The remainder of this paper is organized as follows. 
Section \ref{sec:dynamic_gradflow} introduces our core contribution: the framework of dynamical system-constrained density gradient flows, which rigorously integrates mechanistic model constraints into the geometric optimization process. 
Section \ref{sec:applications} illustrates the proposed methodology through several illustrative examples in biological modeling. Finally, Section \ref{sec:conclusion} concludes the paper with a discussion of the results.

\section{Dynamical System-Constrained Probability Gradient Flows on a Riemannian Manifold}
\label{sec:dynamic_gradflow}

In modern scientific computing and Bayesian inference, we encounter an interplay between two distinct temporal dynamics: the intrinsic evolution of physical systems and the learning dynamics of probability distributions. To avoid notational confusion, we explicitly distinguish between two time scales:
\begin{itemize}
	\item \textbf{Physical (Forward) Time $\tilde{t}$}: This governs the intrinsic evolution of the biological, physical, or engineered dynamical system under study. It represents the natural time scale at which the system state evolves according to its governing equations.
	\item \textbf{Learning (Inverse) Time $t$}: This governs the evolution of the probability distribution over model parameters during inference. It represents the artificial time scale of the gradient flow on the probability manifold, driven by data and prior information.
\end{itemize}

This dual-time perspective allows us to separate the dynamics of the physical system from the dynamics of Bayesian learning, while maintaining their coupling through the model constraints and observational data. Throughout this paper, we have emphasized physical intuition and clarity in problem formulation, method development, and theoretical exposition. Consequently, the problems considered are assumed to possess sufficient smoothness and regularity.

\subsection{Abstract Universal Differential Equations for General Dynamical Systems}

To establish a general mathematical framework that can accommodate a wide range of applications—from simple ODEs to complex mechanistic-AI coupled models—we formulate our approach in abstract functional spaces.
This abstraction not only enhances theoretical generality but also clarifies the essential mathematical structure common to many problem classes.

Consider a dynamical system whose state \(x(\tilde{t})\) evolves in a Banach space \(\mathbb{X}\) according to the abstract evolution equation:
\begin{equation}
	\frac{\mathrm{d} x}{\mathrm{d}\tilde{t}} = F(x, \theta, \tilde{t}), \quad x(0) = x_0,
	\label{eq:abstract_dynamics}
\end{equation}
where \(F: \mathbb{X} \times \Theta \times \mathbb{R} \to \mathbb{X}\) is a (possibly nonlinear) evolution operator, and \(\theta \in \Theta \subset \mathbb{R}^d\) represents unknown parameters. We assume sufficient regularity such that for each \(\theta\), there exists a unique solution \(x(\tilde{t}; \theta)\) in an appropriate function space (e.g., \(C([0,T]; \mathbb{X})\) for ODEs, or Sobolev spaces for PDEs). For PDEs, the evolution operator may be defined on a suitable dense domain in $\mathbb{X}$, and the equation is interpreted in a weak or mild sense; the differentiability assumptions below concern the resulting solution map.

The \emph{Abstract Universal Differential Equation} (UDE) framework
with system \eqref{eq0} as an example
\begin{equation}
	\frac{\mathrm{d} x}{\mathrm{d}\tilde{t}}=G_{\theta_{\text{mech}}}(\tilde{t}, x(\tilde{t}), \mathrm{UniversalApprox}_{\theta_{\text{AI}}}(\tilde{t},x(\tilde{t}))), \quad \tilde{t}\in [0, T],
\end{equation}
emerges naturally within this abstract setting \eqref{eq:abstract_dynamics},
where \(G_{\theta_{\text{mech}}}\) encodes prior mechanistic knowledge (e.g., from physics or biology), and \(\mathrm{UniversalApprox}_{\theta_{\text{AI}}}\) is a universal approximator (e.g., a neural network) that learns unknown or hard-to-model dynamics. The full parameter vector \(\theta = (\theta_{\text{mech}}, \theta_{\text{AI}})\) combines both mechanistic and AI components.

For example, Gaussian observations of the system \eqref{eq:abstract_dynamics} are captured through a (possibly nonlinear) observation operator \(h: \mathbb{X} \to \mathbb{Y}\):
\begin{equation}
	y = h(x(\tilde{t})) + \varepsilon, \quad \varepsilon \sim \mathcal{N}(0, \Sigma),
	\label{eq:abstract_observation}
\end{equation}
where \(\mathbb{Y}\) is the observation space (typically \(\mathbb{R}^m\)), and \(\varepsilon\) represents measurement noise. Other specified likelihoods, including the count model in Section~\ref{sec:applications}, enter through the corresponding loss function.

For clarity, we note that the classical ODE case corresponds to:
\begin{itemize}
	\item \(\mathbb{X} = \mathbb{R}^n\) (finite-dimensional state space).
	\item \(F(x, \theta, \tilde{t}) = f(x, \theta, \tilde{t})\) with \(f: \mathbb{R}^n \times \mathbb{R}^d \times \mathbb{R} \to \mathbb{R}^n\).
	\item The UDE form \eqref{eq0}.
	\item Observation operator: \(h(x) = C x\) (linear) or a more general nonlinear \(h\).
\end{itemize}
Thus, our abstract formulation \eqref{eq:abstract_dynamics}--\eqref{eq:abstract_observation} encompasses both traditional mechanistic models and modern AI-augmented dynamical systems.

\subsection{Formulation of Constrained Probability Gradient Flows}

Following the \emph{Rule of Three} framework for generalized Bayesian inference (\cite{knoblauch2022optimization}, Appendix SM2.2\&2.4), we define the posterior distribution as the minimizer of an energy functional that combines data fidelity with prior information through a divergence measure. Specifically, given an observed dataset $\mathcal{D}$, we consider the energy functional:
\begin{equation}
	\mathcal{F}(\rho) = \mathbb{E}_{\theta \sim \rho}[\ell_{\mathcal{D}}(\theta)] + \beta\,D(\rho \parallel \rho_0),
	\label{eq:energy_functional}
\end{equation}
where:
\begin{itemize}
	\item $\ell_{\mathcal{D}}(\theta)$ is a loss function quantifying the misfit between model predictions and data $\mathcal{D}$ (e.g., negative log-likelihood or a robust alternative),
	\item $D(\cdot \parallel \cdot)$ is a divergence measure that regularizes the solution towards the prior distribution $\rho_0$,
	\item $\beta > 0$ is a regularization parameter controlling the trade-off between data fidelity and prior adherence.
\end{itemize}

The generalized posterior $\rho^*$ is then defined as a minimizer of $\mathcal{F}(\rho)$ over an appropriate space of probability measures, when such a minimizer exists. This variational formulation subsumes many inference paradigms: standard Bayesian inference ($\ell = -\log p(\mathcal{D} \mid \theta)$, $D = \text{KL}$, $\beta=1$), variational inference (with restricted families), and generalized Bayesian methods (using alternative losses or divergences). More details on basics of Riemannian geometry and gradient flows in metric spaces can be seen in Appendix SM2.1 and SM3, respectively.

\subsubsection*{First Variation of the Energy Functional}

The following differential identities are stated for smooth, strictly positive densities for which the energy and displayed integrals are finite, with $\rho_0>0$ on the parameter domain. On a bounded smooth domain we impose no-flux boundary conditions; on $\mathbb{R}^d$ we assume sufficient decay to justify integration by parts. First variations on the probability manifold are defined modulo constants in $\theta$, since admissible perturbations have zero mass. These assumptions justify the formal calculations below; existence and long-time convergence require conditions specific to the chosen energy and metric.

To derive gradient flows that minimize $\mathcal{F}$, we need the first variation $\frac{\delta \mathcal{F}}{\delta \rho}$, defined such that for any admissible perturbation $\xi$ with $\int \xi(\theta) d\theta = 0$,
\begin{equation}
	\lim_{\epsilon \to 0} \frac{\mathcal{F}(\rho + \epsilon \xi) - \mathcal{F}(\rho)}{\epsilon} = \int \frac{\delta \mathcal{F}}{\delta \rho}(\theta) \,\xi(\theta) \, d\theta.
\end{equation}

From \eqref{eq:energy_functional}, we compute:
\begin{equation}
	\frac{\delta \mathcal{F}}{\delta \rho}(\theta) = \ell_{\mathcal{D}}(\theta) + \beta \frac{\delta D(\rho \parallel \rho_0)}{\delta \rho}(\theta).
	\label{eq:first_variation_general}
\end{equation}
The explicit form of the second term depends on the choice of divergence $D$. Below we provide several important examples.

\noindent\emph{Example 1: Kullback-Leibler Divergence.}
For $D(\rho \parallel \rho_0) = \text{KL}(\rho \parallel \rho_0)$,

we have:
\begin{equation}
	\frac{\delta \text{KL}(\rho \parallel \rho_0)}{\delta \rho}(\theta) = \log \frac{\rho(\theta)}{\rho_0(\theta)} + 1.
\end{equation}
Thus,
\begin{equation}
	\frac{\delta \mathcal{F}}{\delta \rho}(\theta) = \ell_{\mathcal{D}}(\theta) + \beta \left( \log \frac{\rho(\theta)}{\rho_0(\theta)} + 1 \right).
	\label{eq:first_var_KL}
\end{equation}
This corresponds to standard Bayesian updating when $\ell_{\mathcal{D}}(\theta) = -\log p(\mathcal{D} \mid \theta)$ and $\beta=1$.

\noindent\emph{Example 2: $\alpha$-Divergence.}
For the $\alpha$-divergence $$D_\alpha(\rho \parallel \rho_0) = \frac{1}{\alpha(\alpha-1)} \left( \int \rho(\theta)^\alpha \rho_0(\theta)^{1-\alpha} d\theta - 1 \right)$$ with $\alpha \neq 0,1$, we have:
\begin{equation}
	\frac{\delta D_\alpha(\rho \parallel \rho_0)}{\delta \rho}(\theta) = \frac{1}{\alpha-1} \left[ \left( \frac{\rho(\theta)}{\rho_0(\theta)} \right)^{\alpha-1} - 1 \right].
\end{equation}
Hence,
\begin{equation}
	\frac{\delta \mathcal{F}}{\delta \rho}(\theta) = \ell_{\mathcal{D}}(\theta) + \frac{\beta}{\alpha-1} \left[ \left( \frac{\rho(\theta)}{\rho_0(\theta)} \right)^{\alpha-1} - 1 \right].
	\label{eq:first_var_alpha}
\end{equation}
As $\alpha \to 1$, this recovers the KL first variation in \eqref{eq:first_var_KL} up to an additive constant, which has no effect on a mass-preserving gradient flow.

\noindent\emph{Example 3: $\chi^2$-Divergence.}
For $D_{\chi^2}(\rho \parallel \rho_0) = \frac{1}{2} \int \frac{(\rho(\theta) - \rho_0(\theta))^2}{\rho_0(\theta)} d\theta$, we have:
\begin{equation}
	\frac{\delta D_{\chi^2}(\rho \parallel \rho_0)}{\delta \rho}(\theta) = \frac{\rho(\theta) - \rho_0(\theta)}{\rho_0(\theta)}.
\end{equation}
Thus,
\begin{equation}
	\frac{\delta \mathcal{F}}{\delta \rho}(\theta) = \ell_{\mathcal{D}}(\theta) + \beta \frac{\rho(\theta) - \rho_0(\theta)}{\rho_0(\theta)}.
	\label{eq:first_var_chi2}
\end{equation}

\subsubsection*{Gradient Flow Equations}

Our objective is to estimate the posterior distribution $\rho_t(\theta)$ of the parameters $\theta$ given observational data $y$, where $t$ denotes the \emph{learning time} of the probability evolution. With the first variation in hand, the gradient flow on the probability manifold under a given metric tensor $\mathcal{M}_\rho$ is (see Appendix SM3 for more details on gradient flows in metric spaces):
\begin{equation}
	\partial_t \rho_t = \mathcal{M}_{\rho_t}^{-1} \left[ -\frac{\delta \mathcal{F}}{\delta \rho} \right],
	\label{eq:general_gradient_flow}
\end{equation}
where $\mathcal{M}_\rho$ encodes the geometry (e.g., Wasserstein, Fisher-Rao, Stein). Its inverse is the nonnegative Onsager operator from cotangent vectors, represented by functions modulo constants, to mass-preserving tangent vectors. For details on Wasserstein, Fisher-Rao, Wasserstein-Fisher-Rao, Sobolev, and Stein metrics and gradient flows, we refer to Appendix SM3.2--3.6.
In the Wasserstein case, $$\mathcal{M}_\rho^{-1}\phi = -\nabla_\theta \cdot (\rho \nabla_\theta \phi),$$ yielding:
\begin{equation}
	\partial_t \rho_t = \nabla_\theta \cdot \left( \rho_t \nabla_\theta \frac{\delta \mathcal{F}}{\delta \rho} \right).
	\label{eq:wasserstein_gradient_flow}
\end{equation}
Substituting the appropriate first variation \eqref{eq:first_var_KL}--\eqref{eq:first_var_chi2} gives concrete evolution equations for the posterior distribution.

Crucially, the loss \(\ell_{\mathcal{D}}(\theta)\) depends implicitly on the solution \(x(\tilde{t}; \theta)\) of the abstract dynamical system \eqref{eq:abstract_dynamics}
\begin{equation}
		\ell_{\mathcal{D}}(\theta)= \mathcal{E}(x(\cdot;\theta)).
\end{equation}
For instance, with Gaussian measurement noise, a typical choice is the squared misfit:
\begin{equation}
	\ell_{\mathcal{D}}(\theta) = \frac{1}{2} \sum_{i=1}^{M} \| y_i - h(x(\tilde{t}_i; \theta)) \|_{\Sigma^{-1}}^2,
\end{equation}
where \(\{y_i, \tilde{t}_i\}_{i=1}^M\) are observed data. Therefore,
\begin{equation}
	\boxed{
		\frac{\delta \mathcal{F}}{\delta \rho}(\theta)=\mathcal{E}(x(\cdot;\theta))+\mathcal{K}(\rho,\theta),
	}
\end{equation}
This gives the following formulation.
\begin{theorem}[Constrained probability gradient flow]
Assume that the forward solution map is well defined, the energy admits the first variation \eqref{eq:first_variation_general}, and $\mathcal{M}_\rho^{-1}$ is a symmetric nonnegative Onsager operator that annihilates constants. Then every sufficiently smooth descending gradient flow satisfies
	\begin{equation}
		\boxed{
			\begin{aligned}
				&\partial_t \rho_t = \mathcal{M}_{\rho_t}^{-1} \left[ -\frac{\delta \mathcal{F}}{\delta \rho} \right],\\
				&	\frac{\delta \mathcal{F}}{\delta \rho}(\theta)=\mathcal{E}(x(\cdot;\theta))+\mathcal{K}(\rho,\theta),\\
				& \frac{\mathrm{d} x}{\mathrm{d}\tilde{t}} = F(x, \theta, \tilde{t}), \quad x(0) = x_0,
		\end{aligned}}
	\end{equation}
	where $\mathcal{K}(\rho,\theta)=\beta\,\delta D(\rho\parallel\rho_0)/\delta\rho(\theta)$ and $\mathcal{E}(x(\cdot;\theta))=\ell_{\mathcal D}(\theta)$. Along such a solution, writing $\psi_t=\delta\mathcal F/\delta\rho|_{\rho_t}$, the energy satisfies
	\begin{equation}
		\frac{\mathrm d}{\mathrm dt}\mathcal F(\rho_t)
		=-\langle\psi_t,\mathcal M_{\rho_t}^{-1}\psi_t\rangle\leq 0.
	\end{equation}
\end{theorem}
\begin{proof}
The first variation follows from the linearity of the expected loss and differentiation of the divergence. Applying the negative Onsager operator gives the descending gradient flow. The chain rule yields
$\mathrm d\mathcal F(\rho_t)/\mathrm dt=\langle\psi_t,\partial_t\rho_t\rangle$, and symmetry and nonnegativity of the Onsager operator give the stated dissipation identity. Since the operator annihilates constants, the total mass is preserved.
\end{proof}

Evaluating \(\nabla_\theta \ell_{\mathcal{D}}(\theta)\) requires differentiating the solution map \(\theta\mapsto x(\cdot;\theta)\). For a state of dimension $n$ (after spatial discretization when needed), the sensitivity matrix $x_\theta$ has size $n\times d$. If $\theta$ includes neural-network weights, storing and evolving this matrix can be expensive. The abstract adjoint method in Section~\ref{sec:sensitivity} replaces these $d$ forward sensitivity solves by one backward adjoint solve for a scalar objective. It still requires parameter vector--Jacobian products and access to the forward trajectory, so its total cost depends on the model and parameter dimension.

\subsection{Abstract Adjoint Sensitivity Analysis}
\label{sec:sensitivity}

We now present a general mathematical framework for adjoint sensitivity analysis in abstract spaces, which avoids the expensive computation of the full parameter Jacobian \(x_\theta\). The ideas are motivated by \cite{cao2003adjoint}. Consider the constrained optimization problem:
\begin{equation}
	\min_{\theta \in \Theta} \; J(\theta) = g(x(\theta), \theta) \quad \text{subject to} \quad c(x, \theta) = 0,
	\label{eq:abstract_opt}
\end{equation}
where \(c(\cdot, \theta): \mathbb{X} \to \mathbb{Y}\) is a differentiable operator between Banach spaces \(\mathbb{X}\) and \(\mathbb{Y}\), and \(g: \mathbb{X} \times \Theta \to \mathbb{R}\) is a differentiable objective functional. The parameter space \(\Theta \subset \mathbb{R}^d\) is finite-dimensional. We assume that for each \(\theta\), there exists a unique solution \(x(\theta) \in \mathbb{X}\) satisfying the constraint \(c(x(\theta), \theta) = 0\), and that the mapping \(\theta \mapsto x(\theta)\) is differentiable. 

\begin{theorem}[Abstract Adjoint Sensitivity]
	\label{thm:adjoint}
	
	Assume, in addition, that the adjoint equation below admits a solution. A sufficient condition is that the Fr\'echet derivative $D_xc(x(\theta),\theta):\mathbb X\to\mathbb Y$ is a bounded isomorphism. Define the Lagrangian functional:
	\begin{equation}
		\mathcal{L}(x, \theta, \lambda) = g(x, \theta) + \langle \lambda, c(x, \theta) \rangle_{\mathbb{Y}^*, \mathbb{Y}},
		\label{eq:abstract_lagrangian}
	\end{equation}
	where \(\lambda \in \mathbb{Y}^*\) is the adjoint variable (Lagrange multiplier) in the dual space of \(\mathbb{Y}\), and \(\langle \cdot, \cdot \rangle_{\mathbb{Y}^*, \mathbb{Y}}\) denotes the duality pairing. Then, the gradient of \(J\) with respect to \(\theta\) is given by:
	\begin{equation}
		\nabla_\theta J(\theta) = \nabla_\theta \mathcal{L}(x, \theta, \lambda) = \nabla_\theta g(x, \theta) + \nabla_\theta c(x, \theta)^* \lambda,
		\label{eq:abstract_gradient}
	\end{equation}
	where \(\lambda\) satisfies the \emph{adjoint equation}:
	\begin{equation}
		\nabla_x g(x, \theta) + \nabla_x c(x, \theta)^* \lambda = 0,
		\label{eq:abstract_adjoint_eq}
	\end{equation}
	with \(\nabla_x c(x, \theta)^*: \mathbb{Y}^* \to \mathbb{X}^*\) denoting the adjoint operator of the Fréchet derivative \(\nabla_x c(x, \theta): \mathbb{X} \to \mathbb{Y}\).
\end{theorem}

\begin{proof}
	By the chain rule, the total derivative of \(J\) is:
	\begin{align*}
		\frac{\mathrm{d}J}{\mathrm{d}\theta} &= \nabla_x g(x, \theta) x_\theta + \nabla_\theta g(x, \theta) +  \langle \lambda, \nabla_x c(x, \theta) x_\theta + \nabla_\theta c(x, \theta)\rangle_{\mathbb{Y}^*, \mathbb{Y}}.
	\end{align*}

	Note that $\nabla_x g(x, \theta) x_\theta = \langle \nabla_x g, x_\theta \rangle_{\mathbb{X}^*, \mathbb{X}}$. For any \(\lambda \in \mathbb{Y}^*\), we have:
	\begin{align*}
		\langle \nabla_x g, x_\theta \rangle_{\mathbb{X}^*, \mathbb{X}} 
		&= \langle \nabla_x g + \nabla_x c^* \lambda, x_\theta \rangle - \langle \nabla_x c^* \lambda, x_\theta \rangle \\
		&= \langle \nabla_x g + \nabla_x c^* \lambda, x_\theta \rangle - \langle \lambda, \nabla_x c \, x_\theta \rangle,
	\end{align*}
	which implies
	\begin{align*}
		\frac{\mathrm{d}J}{\mathrm{d}\theta} &= \langle \nabla_x g + \nabla_x c^* \lambda, x_\theta \rangle+ \nabla_\theta g + \langle \lambda, \nabla_\theta c \rangle.
	\end{align*}
	Choosing \(\lambda\) such that \(\nabla_x g + \nabla_x c^* \lambda = 0\) (the adjoint equation) eliminates the term involving \(x_\theta\), yielding:
	\begin{equation*}
		\frac{\mathrm{d}J}{\mathrm{d}\theta} = \nabla_\theta g + \langle \lambda, \nabla_\theta c \rangle.
	\end{equation*}
	This completes the proof.
\end{proof}

The key advantage is that we avoid computing the \((\dim \mathbb{X}) \times d\) Jacobian \(x_\theta\), which is computationally prohibitive for high-dimensional state spaces (e.g., in PDEs) or large neural networks. Instead, we solve:
\begin{itemize}
	\item The \emph{forward equation} \(c(x, \theta) = 0\) for \(x\) (dimension \(\dim \mathbb{X}\)).
	\item The \emph{adjoint equation} \eqref{eq:abstract_adjoint_eq} for \(\lambda\) (dimension \(\dim \mathbb{Y}\)).
\end{itemize}
Both are typically of similar complexity to the forward problem. The gradient \(\nabla_\theta J\) is then computed via \eqref{eq:abstract_gradient}, which only requires inner products with the adjoint variable \(\lambda\).

\noindent\emph{Example 1: Ordinary Differential Equations (ODEs).}
Consider an ODE system with a fixed, parameter-independent initial state and a loss functional:
\begin{equation}
	\frac{\mathrm{d}x}{\mathrm{d}\tilde t} = m(x, \theta, \tilde t), \quad x(0) = x_0, \qquad
	J(\theta) = \int_0^T e(x(\tilde t), \tilde t) \, d\tilde t.
	\label{eq:ode_example}
\end{equation}
The state belongs to the affine space $\{x\in H^1([0,T];\mathbb R^n):x(0)=x_0\}$; its variations satisfy $\delta x(0)=0$. Define $c(x,\theta)=\dot x-m(x,\theta,\cdot)$ with values in $\mathbb Y=L^2([0,T];\mathbb R^n)$. Then:
\begin{itemize}
	\item $D_xc[\delta x]=\dot{\delta x}-m_x\delta x$. Integration by parts identifies its adjoint expression as $D_xc^*[\lambda]=-\dot\lambda-m_x^\top\lambda$, with the terminal condition below.
	\item $D_xJ[\delta x]=\int_0^T e_x\delta x\,d\tilde t$.
	\item The adjoint equation \eqref{eq:abstract_adjoint_eq} becomes
	\begin{equation}
		-\frac{\mathrm{d}\lambda}{\mathrm{d}\tilde t}-m_x^\top\lambda+e_x^\top=0,
		\qquad \lambda(T)=0.
	\end{equation}
	\item Since $D_\theta c=-m_\theta$, the gradient is
	\begin{equation}
		\boxed{\nabla_\theta J=-\int_0^T m_\theta^\top\lambda(\tilde t)\,d\tilde t.}
	\end{equation}
\end{itemize}
For the discrete observation loss $J=\sum_i q_i(x(\tilde t_i))$, the adjoint instead satisfies $-\dot\lambda-m_x^\top\lambda=0$ between observation times and
\begin{equation}
	\lambda(\tilde t_i^-)=\lambda(\tilde t_i^+)-\nabla_xq_i(x(\tilde t_i)),
	\qquad \lambda(T^+)=0.
\end{equation}
Here $T\geq\max_i\tilde t_i$, and the terminal jump is included if an observation occurs at $T$. For Gaussian observations, $\nabla_xq_i=h_x(x_i)^\top\Sigma^{-1}(h(x_i)-y_i)$, where $x_i=x(\tilde t_i)$. The same parameter-gradient formula applies. Direct parameter dependence in the loss adds its partial derivative, while a parameter-dependent initial state adds $-D_\theta x_0^\top\lambda(0)$.

\noindent\emph{Example 2: Partial Differential Equations (PDEs).}
Consider a stationary PDE and a quantity of interest:
\begin{equation}
	\mathcal{L}(u, \theta) = 0 \text{ in } \Omega, \quad u = 0 \text{ on } \partial\Omega, \qquad 
	J(\theta) = \int_\Omega e(u) \, dx,
	\label{eq:pde_example}
\end{equation}
where $\mathcal{L}$ is a differential operator (e.g., $-\Delta u - g(\theta)$). Let $\mathbb{X} = H_0^1(\Omega)$, $\mathbb{Y} = H^{-1}(\Omega)$. Then:
\begin{itemize}
	\item $\nabla_u \mathcal{L}[\delta u]$ is the linearized PDE operator, and its adjoint $\nabla_u \mathcal{L}^*$ is the formal adjoint.
	\item $\nabla_u J[\delta u] = \int_\Omega e_u \delta u \, dx$.
	\item The adjoint equation is: $\nabla_u \mathcal{L}^*[\lambda] = -e_u$, with homogeneous boundary conditions.
	\item The gradient is: $\nabla_\theta J = \langle \lambda, \nabla_\theta \mathcal{L} \rangle$.
\end{itemize}
For the specific case $\mathcal{L}(u, \theta) = -\Delta u - g(\theta)$, we have:
\begin{itemize}
	\item Adjoint equation: $-\Delta \lambda = -e_u$, $\lambda|_{\partial\Omega}=0$.
	\item Gradient: $\nabla_\theta J = -\int_\Omega \lambda \, \nabla_\theta g \, dx$.
\end{itemize}

For a UDE of the form $\frac{\mathrm{d}x}{\mathrm{d}t} = f(x, \mathrm{NN}_\theta(t,x), t)$, the abstract framework applies directly by treating the neural network as part of the parameter $\theta$. The adjoint equation incorporates derivatives with respect to both the state and the neural network output, enabling efficient gradient computation for training. The abstract adjoint method provides a systematic way to compute gradients for optimization problems constrained by differential equations, avoiding the high-dimensional parameter-state Jacobian. For a scalar objective, it requires one forward and one adjoint state solve, together with parameter derivative products. This avoids the full state--parameter sensitivity matrix; the resulting computational gain depends on the solver, the cost of differentiating the model, and the storage or recomputation of the forward trajectory.

\subsection{Computational Framework and Particle Implementation}

The theoretical framework of constrained gradient flows provides a principled foundation for Bayesian inference in dynamical systems, but its practical implementation requires efficient numerical methods. In this subsection, we give a computational framework. 
Particle-based approaches offer a natural and flexible computational paradigm, where the probability distribution $\rho_t$ is approximated by an ensemble of $N$ particles $\{\theta_t^i\}_{i=1}^N$. These particles evolve according to discretized versions of the gradient flow equations, effectively performing parallel, interacting Bayesian inference.

Given a gradient flow equation of the general form $\partial_t\rho_t=\mathcal M_{\rho_t}^{-1}[-\delta\mathcal F/\delta\rho]$, we seek a particle approximation $\rho_t^N=\sum_{i=1}^N w_t^i\delta_{\theta_t^i}$, where $w_t^i\geq0$ and $\sum_iw_t^i=1$. Transport-only schemes can use $w_t^i=1/N$, while reaction geometries evolve the weights. For the Wasserstein gradient flow \eqref{eq:wasserstein_gradient_flow}, the formal characteristic equations are

\begin{equation}
	\frac{\mathrm{d}\theta_t^i}{\mathrm{d}t} = -\nabla_\theta \frac{\delta \mathcal{F}}{\delta \rho}(\theta_t^i), \quad i=1,\dots,N.
	\label{eq:particle_ode}
\end{equation}
Here the first variation is evaluated at the smooth density $\rho_t$. If the energy contains a KL or another density-ratio divergence relative to a continuous prior, the atomic measure $\rho_t^N$ cannot be substituted directly: its divergence is generally infinite. A smooth positive density reconstruction, a Stein identity, or a stochastic diffusion representation is required. The central challenge is to evaluate

\begin{equation}
	\nabla_\theta \frac{\delta \mathcal{F}}{\delta \rho}(\theta) = \nabla_\theta \ell_{\mathcal{D}}(\theta) + \beta \nabla_\theta \frac{\delta D(\rho \parallel \rho_0)}{\delta \rho}(\theta).
	\label{eq:total_gradient}
\end{equation}
The first term $\nabla_\theta \ell_{\mathcal{D}}(\theta)$ requires differentiation through the dynamical system \eqref{eq:abstract_dynamics}, which we address via adjoint sensitivity analysis. The second term depends on the chosen divergence and the current distribution $\rho$.

Different metric geometries lead to distinct particle updates:
\begin{itemize}
	\item Wasserstein geometry: the characteristic velocity is $v_t=-\nabla_\theta(\delta\mathcal F/\delta\rho)$. Particle interactions enter through the dependence of this first variation on $\rho$, rather than through the fixed loss $\ell_{\mathcal D}$. For $D=\mathrm{KL}$, define $U(\theta)=\ell_{\mathcal D}(\theta)-\beta\log\rho_0(\theta)$. The same density flow admits the diffusion representation
	\begin{equation}
		\mathrm d\theta_t=-\nabla U(\theta_t)\,\mathrm dt+\sqrt{2\beta}\,\mathrm dW_t,
	\end{equation}
	whose target density is $\pi_\beta\propto\rho_0\exp(-\ell_{\mathcal D}/\beta)$, provided it is normalizable. This avoids evaluating the density score explicitly; time discretization introduces an error unless it is controlled or corrected as described in Appendix SM4.
	\item Stein geometry: for $D=\mathrm{KL}$, integration by parts gives the particle velocity and update
	\begin{equation}\label{eq:svgd_corrected}
	\begin{aligned}
		v_n^i&=\sum_{j=1}^Nw_n^j\left[-k(\theta_n^j,\theta_n^i)\nabla U(\theta_n^j)
		+\beta\nabla_{\theta_n^j}k(\theta_n^j,\theta_n^i)\right],\\
		\theta_{n+1}^i&=\theta_n^i+\tau_n v_n^i.
	\end{aligned}
	\end{equation}
	The kernel is differentiable and positive definite, and the integration-by-parts boundary term is assumed to vanish. The positive kernel-derivative term provides repulsion. For a general divergence, the kernelized velocity is $v_t=-\mathcal T_{k,\rho_t}\nabla_\theta(\delta\mathcal F/\delta\rho)$ and requires the corresponding first variation.
	\item Fisher--Rao geometry: positions remain fixed while weights satisfy $\dot w_t^i=-w_t^i(\psi_t^i-\sum_jw_t^j\psi_t^j)$, where $\psi_t^i$ approximates the first variation at $\theta_t^i$. Normalized exponential reweighting gives a positive update for frozen $\psi_t^i$.
	\item Mass-preserving Wasserstein--Fisher--Rao geometry: transport and centered reaction are combined, for example by alternating a position update and normalized weight update.
\end{itemize}

For each particle $\theta^i$, computing $\nabla_\theta \ell_{\mathcal{D}}(\theta^i)$ requires solving the forward dynamical system \eqref{eq:abstract_dynamics} and its associated adjoint equation. This two-step process can be implemented as follows: (1) Forward Solve: For particle $\theta^i$, compute the solution $x(\tilde{t}; \theta^i)$ of \eqref{eq:abstract_dynamics} on the physical time interval $[0,T]$. (2) Adjoint Solve: Solve the backward adjoint equation (as derived in Theorem \ref{thm:adjoint}) to obtain the sensitivity $\lambda(\tilde{t})$. (3) Gradient Assembly: Compute $\nabla_\theta \ell_{\mathcal{D}}(\theta^i)$ via its abstract counterpart \eqref{eq:abstract_gradient}.

Crucially, these computations are \emph{embarrassingly parallel} across particles, as each particle's gradient evaluation is independent of others (except for potential interactions in the metric term). This parallelism makes particle methods particularly suitable for modern high-performance computing architectures.
We now present a high-level algorithmic framework that synthesizes the particle approximation, gradient flow discretization, and adjoint sensitivity analysis:
\begin{algorithm}[H]
	\caption{Particle-Based Constrained Gradient Flow with Adjoint Sensitivities}
	\label{alg:constrained_particle_flow}
	\begin{enumerate}
		\item \textbf{Input:} Prior $\rho_0$, particle number $N$, geometry, divergence $D$, regularization parameter $\beta$, step sizes $\{\tau_k\}_{k=0}^{K-1}$, and stopping tolerance $\epsilon$.
		\item \textbf{Initialize:} Draw $\theta_0^i\sim\rho_0$ and set $w_0^i=1/N$.
		\item \textbf{For} $k=0,\ldots,K-1$:
		\begin{enumerate}
			\item Form $\rho_k^N=\sum_iw_k^i\delta_{\theta_k^i}$. Whenever a density-dependent first variation is needed, construct a smooth positive density $\widehat\rho_k$ from the particles.
			\item For each particle, solve the forward system \eqref{eq:abstract_dynamics} and the adjoint problem in Theorem~\ref{thm:adjoint}. Compute $\ell_{\mathcal D}(\theta_k^i)$ and its parameter gradient, including observation jumps when appropriate.
			\item As required by the selected geometry, evaluate
			\[
			\psi_k(\theta)=\ell_{\mathcal D}(\theta)+\beta\frac{\delta D(\widehat\rho_k\parallel\rho_0)}{\delta\rho}(\theta),
			\qquad g_k^i=\nabla_\theta\psi_k(\theta_k^i).
			\]
			For KL--Stein updates, use \eqref{eq:svgd_corrected} directly, which eliminates the density score.
			\item Compute the position velocities: $v_k^i=-g_k^i$ for Wasserstein or mass-preserving WFR, and $v_k^i=0$ for Fisher--Rao. For Stein with KL use \eqref{eq:svgd_corrected}; for Stein with other divergences use $v_k^i=-\sum_jw_k^j k(\theta_k^j,\theta_k^i)g_k^j$.
			\item Update positions by $\theta_{k+1}^i=\theta_k^i+\tau_kv_k^i$, enforcing the parameter-domain boundary condition if needed.
			\item For Fisher--Rao or WFR, set
			\[
			w_{k+1}^i=\frac{w_k^i\exp[-\tau_k\psi_k(\theta_k^i)]}
			{\sum_jw_k^j\exp[-\tau_k\psi_k(\theta_k^j)]}.
			\]
			Otherwise retain $w_{k+1}^i=w_k^i$. Optional resampling must preserve this weighted measure in expectation.
			\item Stop when the position velocities and, for reaction flows, the centered weight residuals are below $\epsilon$.
		\end{enumerate}
		\item \textbf{Output:} The final weighted empirical measure $\sum_iw^i\delta_{\theta^i}$.
	\end{enumerate}
\end{algorithm}
The density Onsager operator acts on functions to produce density rates; particle velocities and weight rates are derived from those rates, as above. The algorithm is a discretization template. Its finite-particle, density-reconstruction, and time-discretization errors must be assessed separately from the update residual used for stopping.

Several practical aspects merit attention in implementations:
\begin{itemize}
	\item Adaptive Time Stepping: The learning rate $\tau_k$ can be adapted based on gradient norms or energy decrease to improve stability and convergence.
	
	\item Kernel Selection: For Stein and kernel-based methods, the choice of kernel $k$ and its bandwidth significantly affects performance. Adaptive or learned kernels can enhance efficiency.
	
	\item Regularization: The divergence regularization parameter $\beta$ balances data fidelity and prior adherence. Cross-validation or empirical Bayes methods can guide its selection.
	
	\item High-Dimensional State Spaces: For PDE-constrained problems where $\dim(\mathbb{X})$ is large (e.g., after spatial discretization), model reduction techniques (proper orthogonal decomposition, neural operators) can accelerate forward and adjoint solves.
	
	\item Stochastic Gradients: When the dataset is large, stochastic approximations of $\nabla_\theta \ell_{\mathcal{D}}$ using mini-batches can reduce computational cost, though care must be taken to maintain convergence properties.
\end{itemize}

The formulation connects several existing computational approaches:
\begin{itemize}
	\item Stein Variational Gradient Descent (SVGD): Emerges as a special case of the Stein gradient flow discretization with specific kernel choices.
	
	\item Particle Filtering: Reweighting and resampling connect reaction-based particle schemes to particle-filter computations. Filtering additionally requires a sequential observation model and latent-state prediction.
	
	\item Variational Inference: When particles are replaced by parametric distributions and the metric is chosen appropriately, we recover variational inference schemes.
	
	\item Ensemble Kalman Methods: Covariance-based mobilities and suitable Gaussian approximations lead to related ensemble Kalman-type updates.
\end{itemize}
The particle implementation bridges theoretical gradient flows with practical Bayesian computation, providing a flexible, parallelizable, and geometrically principled approach to uncertainty quantification in dynamical systems. Combining particle methods with adjoint sensitivity analysis provides a practical route to inference in mechanistic and AI-augmented models.

The constrained gradient flow formulation links two temporal scales: the physical time $\tilde{t}$ governing the intrinsic evolution of the dynamical system \eqref{eq:abstract_dynamics}, and the learning time $t$ governing the evolution of the parameter distribution $\rho_t$ on the probability manifold. The adjoint sensitivity analysis serves as the crucial link between these scales, efficiently propagating gradient information from the physical system to the probability evolution. This gives a computational formulation for Bayesian inference in differential equation models.

For further numerical methods for gradient flows on density manifolds, we refer to Appendix SM4. We systematically explore three complementary approaches to probability computation: time-averaging methods (exploiting ergodicity and MCMC) (\cite{ma2015complete,brooks2011handbook,girolami2011riemann,bou2018geometric,betancourt2017conceptual,hoffman2014no,xifara2014langevin}), spatial-averaging methods (particle discretizations and variational inference) (\cite{liu2016stein,liu2017stein,duncan2023geometry}), and proximal methods (JKO schemes). Each approach offers distinct advantages and reveals connections to existing Bayesian inference algorithms, while maintaining the geometric structure essential for efficient exploration of high-dimensional parameter spaces under dynamical constraints.

\section{Applications}
\label{sec:applications}

We present two applications in biological dynamical systems: Bayesian inference in a mechanism-AI coupled model and the evolution of heterogeneity within a population. The first compares posterior approximation and predictive uncertainty on synthetic and Ontario case data. The second illustrates how selection and trait transport shape the distribution of infectivity. The experiments use Python and C++; the source code, data, and fixed configurations are supplied with the manuscript. Related implementations can also be built using Julia \cite{bezanson2017julia}, DifferentialEquations.jl \cite{rackauckas2017differentialequations}, and Turing.jl \cite{fjelde2025turing}.

\subsection{Bayesian Inference for Mechanism-AI Coupled Models}
\label{subsec:rt_nn}

We consider uncertainty in the estimation of a time-varying effective reproduction number from incidence data, extending the point-estimation problem in \cite{song2022estimating}. The transmission component is represented by a neural network whose inputs are time and the infectious population. We retain the mechanism-AI coupled system
\begin{equation}\label{eq:rt_nn_system}
\left\{\begin{aligned}
\frac{\mathrm{d}I}{\mathrm{d}t}&=\gamma\bigl(\mathcal{R}_{\theta}(t,I)-1\bigr)I,\\
\frac{\mathrm{d}H}{\mathrm{d}t}&=\gamma\mathcal{R}_{\theta}(t,I)I.
\end{aligned}\right.
\end{equation}
Here $H$ accumulates the modeled incidence and $\gamma>0$ is fixed. For reported-case data, this is a reduced observation-scale model; identifying its states with all infections would require additional assumptions about ascertainment and reporting delays.

We use a network with eight hidden units and a positive, bounded output:
\begin{equation}\label{eq:rt_positive_network}
\mathcal{R}_{\theta}(t,I)
=3\,\operatorname{sigmoid}\!\left[b+\sum_{j=1}^{8}v_j
\tanh\!\left(a_j\tau(t)+c_j u(I)+d_j\right)\right],
\end{equation}
where $\tau(t)=2t/n_{\mathrm{tr}}-1$, $u(I)=\log(I/100)/5$, time is measured in days, and $\operatorname{sigmoid}(z)=(1+e^{-z})^{-1}$. The training horizon $n_{\mathrm{tr}}$ fixes the time normalization; it is not recomputed for the held-out interval. The initial value of $I$ is positive, so the logarithmic input is well defined. This architecture has 33 parameters: 24 hidden-layer parameters $\theta_{\mathrm{det}}=(a_j,c_j,d_j)_{j=1}^{8}$ and nine output parameters $w=\theta_{\mathrm{last}}=(v_1,\ldots,v_8,b)$.

The observations are daily increments, rather than independently perturbed cumulative counts. With $y_k$ the count reported in day $k$, we use
\begin{equation}\label{eq:rt_daily_likelihood}
y_k\mid\theta\sim\operatorname{NegBin}(\mu_k(\theta),\kappa),
\qquad \mu_k(\theta)=H(k;\theta)-H(k-1;\theta),
\end{equation}
independently conditional on $\theta$. The parameterization has mean $\mu_k$ and variance $\mu_k+\mu_k^2/\kappa$. The shape $\kappa$ is fixed separately for each experiment below. All available training days, including days with zero reports, enter the likelihood.

Following the last-layer Bayesian approach \cite{harrison2024variational,sharma2023bayesian}, we first obtain a regularized point estimate of the full network. Writing $\mathcal{L}(\theta;y)=-\sum_{k=1}^{n_{\mathrm{tr}}}\log p(y_k\mid\theta)$, this stage solves
\begin{equation}\label{eq:det_training}
\underset{\theta}{\operatorname{minimize}}
\left\{\mathcal{L}(\theta;y)-\log p_0(\theta)\right\},
\qquad\text{subject to \eqref{eq:rt_nn_system}}.
\end{equation}
The independent priors are $a_j,d_j\sim\mathcal{N}(0,5^2)$, $c_j\sim\mathcal{N}(0,1)$, $v_j\sim\mathcal{N}(0,2^2)$, and $b\sim\mathcal{N}(\log(1/2),2^2)$. These priors are fixed before fitting. Three optimization initializations are compared using the training objective, followed by a gradient-based refinement. We denote the best solution reached by $\widehat\theta$; global optimality is not assumed. The hidden parameters are then fixed at $\widehat\theta_{\mathrm{det}}$. For Bayesian inference, we specialize the divergence to relative entropy and set its coefficient to one:
\begin{equation}\label{eq:lastlayer_bayes}
\underset{\rho\in\mathcal{P}(\Theta_{\mathrm{last}})}{\operatorname{minimize}}
\left\{\int\mathcal{L}(\widehat\theta_{\mathrm{det}},w;y)\,\rho(\mathrm{d}w)
+\operatorname{KL}(\rho\Vert p_{0,\mathrm{last}})\right\},
\quad\text{subject to \eqref{eq:rt_nn_system}}.
\end{equation}
Its minimizer is the conditional posterior
$\pi(\mathrm{d}w\mid y,\widehat\theta_{\mathrm{det}})\propto
e^{-\mathcal{L}(\widehat\theta_{\mathrm{det}},w;y)}p_{0,\mathrm{last}}(\mathrm{d}w)$.
The output prior remains the fixed prior above; it is not recentered at the fitted weights. The resulting credible intervals describe uncertainty conditional on the learned representation, the initial state, $\gamma$, and $\kappa$. They do not include uncertainty in the fitted hidden layers.

We compare three discretizations on this same conditional target: the Metropolis-adjusted Langevin algorithm (MALA), Stein variational gradient descent (SVGD), and the unadjusted Langevin algorithm (ULA). ULA includes the Brownian increment required by the Wasserstein gradient flow of relative entropy. SVGD includes both the posterior-score term and the kernel-repulsion term. All methods use the same fixed affine coordinates obtained from the Hessian of the negative log posterior at the output-layer mode. In these coordinates, SVGD uses the Gaussian kernel $k(z,z')=\exp(-\|z-z'\|^2/h)$, the median-distance bandwidth $h=\operatorname{median}_{i<j}\|z_i-z_j\|^2/\log(65)$, and step size $0.06$; ULA uses step size $0.02$.

Each comparison run has a budget of 32,768 single-particle likelihood-gradient evaluations, including initialization and MALA warmup. Each evaluation computes the full nine-dimensional gradient at one parameter vector. SVGD and ULA use 64 particles for 512 updates and are evaluated through their terminal ensembles. MALA uses four chains, each with 1,000 warmup iterations and 7,191 retained iterations; its step size is adapted only during warmup. Thus the time-averaged MALA measure contains 28,764 retained states. The full-network fit and Hessian calculation are common preprocessing, excluded from the comparison budget. We repeat each method with three random seeds on the same data. No MCMC effective sample size is assigned to the deterministic SVGD ensemble.

An independently initialized, longer MALA run supplies the reference posterior: four chains, 2,500 warmup iterations per chain, and 40,000 retained states in total. Rank-normalized split and folded $\widehat R$ and bulk/tail effective sample sizes are computed as in \cite{vehtari2021rank}. Across both problems and both the original and affine output coordinates, the largest $\widehat R$ is below $1.001$, and the smallest bulk effective sample size exceeds $10{,}600$. These diagnostics support use of this numerical reference within the conditional model.

\begin{figure}[htbp]
\centering
\includegraphics[width=\linewidth]{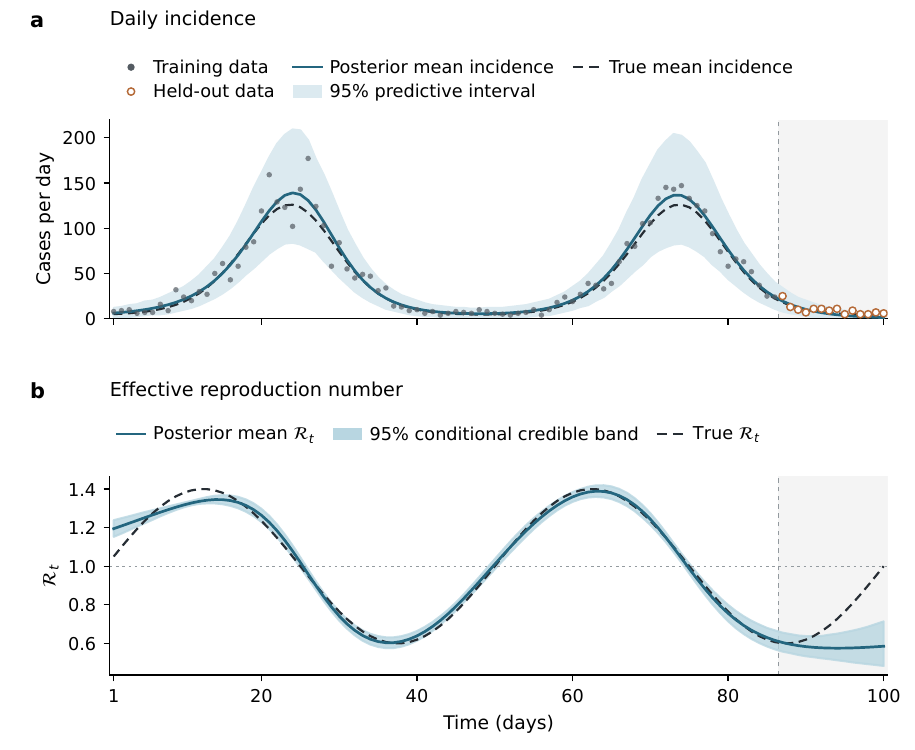}
\caption{Conditional last-layer inference from synthetic daily counts. (a) Observations, the true generating mean incidence, and the posterior mean incidence. The shaded band is the central 95\% posterior predictive interval, including negative-binomial observation noise. (b) The true effective reproduction number $\mathcal{R}_t$, its posterior mean, and its pointwise 95\% conditional credible band. The horizontal dotted line marks $\mathcal{R}_t=1$. Filled and open circles denote training and held-out observations, respectively; the grey background marks days 87--100, excluded from both network fitting and posterior inference. All posterior summaries use the independently initialized, longer MALA reference run and condition on the fitted hidden layers, initial state, $\gamma$, and $\kappa$.}
\label{fig:last_layer_results}
\end{figure}

For the synthetic experiment, we retain the time-dependent transmission law
\[
\beta_{\mathrm{true}}(t)=0.5+0.2\sin(2\pi t/50),\qquad
\mathcal{R}_{\mathrm{true}}(t)=\beta_{\mathrm{true}}(t)/\gamma,
\]
and set $\gamma=0.5$, $I(0)=10$, and $H(0)=0$ in \eqref{eq:rt_nn_system}. This generates data from the same reduced system used for inference. We observe 100 daily counts with negative-binomial shape $\kappa=25$, fit days 1--86, and reserve days 87--100 for prediction. Figure~\ref{fig:last_layer_results} shows the reference posterior's reconstruction of mean daily incidence and of the effective reproduction number $\mathcal{R}_t$. The training-interval root mean square error for $\mathcal{R}_t$ is 0.032; it increases to 0.211 over the held-out interval as the true transmission law turns upward. The 95\% posterior predictive intervals contain 85 of 86 training observations and 10 of 14 held-out observations. The corresponding conditional credible band for $\mathcal{R}_t$ contains the truth at 4 of the 14 held-out daily times. This single-trajectory time-grid coverage is descriptive, not a repeated-sampling calibration guarantee. The deterioration under extrapolation shows why conditional last-layer uncertainty should not be interpreted as uncertainty in the entire learned dynamical model.

For the real-data example, we use the Ontario Ministry of Health's public daily case series \cite{ontario2020cases}. Daily incidence is the first difference of its \emph{Total Cases} column, with the value on February 24, 2020 serving as the baseline. February 25 through July 25 contains 152 daily observations and 38,540 newly reported cases. The raw file and extraction code are supplied with the reproducibility materials. We use February 25--July 11 for fitting and July 12--25 as a 14-day holdout. The working model fixes $\gamma=1/7$, $\kappa=20$, $H(0)=3$, and $I(0)=1$, the reported unresolved-case count at the baseline. No held-out observations are used to choose features, priors, or tuning parameters.

\begin{figure}[htbp]
\centering
\includegraphics[width=\linewidth]{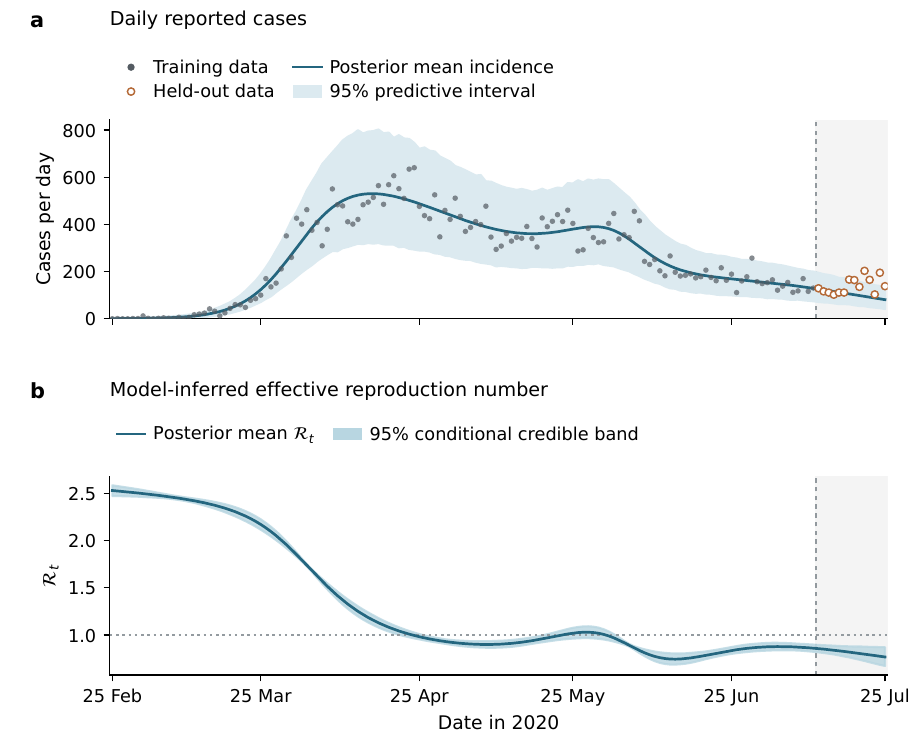}
\caption{Conditional last-layer inference from Ontario daily reported cases, February 25--July 25, 2020. (a) Reported daily counts, posterior mean incidence, and the central 95\% posterior predictive interval, including negative-binomial observation noise. (b) The model-inferred effective reproduction number: posterior mean $\mathcal{R}_t$ and pointwise 95\% conditional credible band. The horizontal dotted line marks $\mathcal{R}_t=1$. Filled and open circles denote training and held-out observations; the grey background marks July 12--25, excluded from both network fitting and posterior inference. No ground-truth reproduction-number curve is available. All posterior summaries use the independently initialized, longer MALA reference run and condition on the fitted hidden layers, initial state, $\gamma$, and $\kappa$.}
\label{fig:last_layer_results2}
\end{figure}

Figure~\ref{fig:last_layer_results2} shows that the model-inferred effective reproduction number decreases through much of the first wave. There is no observed ``true'' $\mathcal{R}_t$ against which to validate its credible band. The posterior-mean incidence has a root mean square error of 49.8 cases per day on the training interval and 54.9 on the holdout; the 95\% predictive intervals contain 133 of 138 training observations and 9 of 14 held-out observations. The reduced model misses part of the late increase in reported cases. These are retrospective holdout results from the downloaded data vintage. Changing ascertainment, delays, observation dependence, or uncertainty in the fixed quantities would require a richer observation model and can change the inferred $\mathcal{R}_t$.

\begin{figure}[htbp]
\centering
\includegraphics[width=\linewidth]{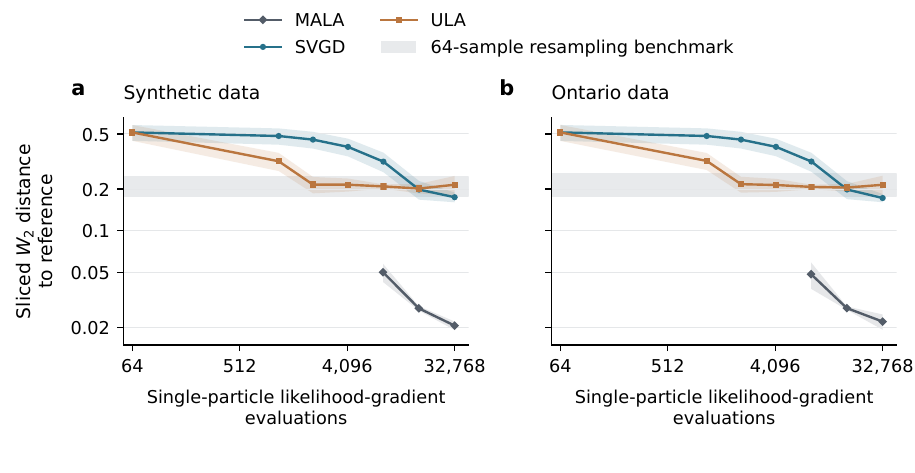}
\caption{Accuracy of conditional-posterior approximation at increasing computational budgets: (a) synthetic data and (b) Ontario data. The vertical axis is the sliced $W_2$ distance to the empirical distribution from an independently initialized, longer MALA reference run, evaluated in the common nine-dimensional affine coordinates. The horizontal axis counts full likelihood-gradient evaluations at individual parameter vectors. Curves show means and coloured bands show $\pm$ one sample standard deviation across three algorithmic seeds on the same dataset. MALA uses all retained post-warmup states at each budget; SVGD and ULA use their 64 current particles. The horizontal grey band is the central 95\% range of distances from 100 independently resampled 64-point reference sets to the full empirical reference. It measures finite-ensemble sampling variation, not posterior uncertainty or uncertainty in an algorithm's mean curve. Comparison budgets include initialization and MALA warmup but exclude common preprocessing and the reference run.}
\label{fig:posterior_comparison}
\end{figure}

Figure~\ref{fig:posterior_comparison} and Table~\ref{tab:lastlayer_posteriors} assess posterior-approximation accuracy separately from predictive-model error. The sliced $W_2$ discrepancy is evaluated in the common affine coordinates with 64 fixed projection directions and 999 quantile levels. The mean error is the root mean squared componentwise mean difference scaled by reference marginal standard deviations; the covariance error is the relative Frobenius error of the covariance matrix. At this budget, the time-averaged MALA measure has the smallest discrepancy in both nine-dimensional conditional problems. To quantify finite-ensemble effects, we also draw 100 independent 64-point samples with replacement from the empirical reference. Their mean sliced $W_2$ discrepancies are 0.206 and 0.206 for the synthetic and Ontario problems, respectively. This resampling benchmark, shown as the grey band in Figure~\ref{fig:posterior_comparison}, places the terminal SVGD and ULA errors in the context of the error expected from a 64-point empirical measure. It does not identify discretization bias separately from finite-particle error. The comparison supports a common geometric formulation while making no general efficiency claim across posterior dimensions or geometries.

\begin{table}[htbp]
\centering
\small

\begin{tabular}{llccc}
\hline
Dataset & Method & Sliced $W_2$ & Mean error & Covariance error \\
\hline
Ontario & MALA & $0.022 \pm 0.003$ & $0.013 \pm 0.004$ & $0.037 \pm 0.003$ \\
Ontario & SVGD & $0.172 \pm 0.012$ & $0.054 \pm 0.009$ & $0.274 \pm 0.027$ \\
Ontario & ULA & $0.215 \pm 0.035$ & $0.138 \pm 0.040$ & $0.338 \pm 0.024$ \\
Synthetic & MALA & $0.021 \pm 0.001$ & $0.013 \pm 0.001$ & $0.036 \pm 0.001$ \\
Synthetic & SVGD & $0.175 \pm 0.014$ & $0.055 \pm 0.008$ & $0.275 \pm 0.027$ \\
Synthetic & ULA & $0.215 \pm 0.034$ & $0.140 \pm 0.039$ & $0.338 \pm 0.027$ \\
\hline
\end{tabular}

\caption{Errors relative to the independent conditional-posterior reference at 32,768 likelihood-gradient evaluations per run. Values are means $\pm$ sample standard deviations over three algorithmic seeds. The mean and covariance errors are defined in the text. MALA uses its 28,764 retained states; SVGD and ULA each use 64 terminal particles.}
\label{tab:lastlayer_posteriors}
\end{table}

The computations use double precision, classical fourth-order Runge--Kutta integration of $\log I$ and daily incidence with a $0.25$-day step, and the associated tangent sensitivity equations. Direct sensitivities are inexpensive for this nine-parameter conditional problem and differentiate the same constrained likelihood as the adjoint formulation. Directional finite-difference checks of both the full-network and last-layer gradients have maximum relative errors below $4.3\times10^{-7}$. Halving the ODE step changes the negative log posterior by less than $2.6\times10^{-7}$ at the tested parameter values. The supplied Python/C++ code records the fixed configurations, raw observations, random seeds, gradients, posterior draws, diagnostics, and numerical tables from which all figures are generated.

\subsection{Heterogeneity Evolution as Gradient Flows on Density Manifolds}
\label{sec:heterogeneity_flow}

We consider an SIR-type epidemic model in which the infected population is structured by a heterogeneity parameter $a\in A=[a_-,a_+]\subseteq\mathbb R_+$, representing an inherited infectivity or activity trait. Here $t$ denotes physical time. Let $I(t,a)$ denote the density of infected individuals with trait $a$ and define $\bar I(t)=\int_A I(t,a)\,da$. As long as $\bar I(t)>0$, the distribution within the infected class is the probability density
\begin{equation}
	\rho(t,a) = \frac{I(t,a)}{\bar I(t)}, \qquad \int_A \rho(t,a)\,da = 1.
\end{equation}
The epidemic dynamics in the absence of trait evolution are given by the heterogeneous SIR model
\begin{equation}\label{eq:SIR_hetero}
	\left\{
	\begin{aligned}
		&	\frac{\mathrm{d}S}{\mathrm{d}t} = -\frac{S(t)}{N}\int_A \beta(a) I(t,a)\,da, \\
		&	\frac{\partial I(t,a)}{\partial t} = \frac{\beta(a) S(t) I(t,a)}{N} - \gamma(a) I(t,a), \\
		&	\frac{\mathrm{d}R}{\mathrm{d}t} = \int_A \gamma(a) I(t,a)\,da,
	\end{aligned}
	\right.
\end{equation}
where $\beta(a)$ and $\gamma(a)$ are trait-dependent transmission and recovery rates. The infection term assumes that newly infected individuals inherit the trait of the transmitting class. In particular, $N=S+\bar I+R$ is conserved.

When heterogeneity also changes within the infected class, add the drift--diffusion term $-\partial_a(\mu_b I)+\tfrac12\partial_{aa}(\sigma^2 I)$ to its equation in \eqref{eq:SIR_hetero}, with zero flux at $a_-$ and $a_+$. Here $\mu_b$ describes baseline trait changes and $\sigma$ their stochastic variation. Writing
\[
 m(t,a)=\frac{S(t)}N\beta(a)-\gamma(a),\qquad
 \bar m(t)=\int_A m(t,a)\rho(t,a)\,da,
\]
integration gives $\dot{\bar I}=\bar m\bar I$, and normalization yields
\begin{equation}\label{eq:normalized_heterogeneity}
 \partial_t\rho=\rho(m-\bar m)-\partial_a(\mu_b\rho)
 +\frac12\partial_{aa}(\sigma^2\rho).
\end{equation}
Thus epidemic selection is a reaction term in the normalized density. In one trait dimension it also admits a transport representation. For a smooth positive density, a reflected diffusion realizing that representation has the form
\begin{equation}\label{eq:a_SDE}
 da_t=[\mu_{\mathrm{ep}}(t,a_t)+\mu_b(t,a_t)]\,dt
 +\sigma(t,a_t)\,dW_t+dK_t,
\end{equation}
where $K_t$ is the inward reflection term supported on the endpoints of $A$. The drift $\mu_{\mathrm{ep}}$ depends on the current density and represents selection at the level of its law. Its Fokker--Planck equation is
\begin{equation}\label{eq:FP_full}
 \partial_t\rho=-\partial_a[(\mu_{\mathrm{ep}}+\mu_b)\rho]
 +\frac12\partial_{aa}(\sigma^2\rho),
\end{equation}
with zero total flux at the endpoints. The required selection flux is
\begin{equation}\label{eq:mu_ep}
 \mu_{\mathrm{ep}}(t,a)=-\frac{S(t)}N
 \frac{\int_{a_-}^a[\beta(b)-\bar\beta(t)]\rho(t,b)\,db}{\rho(t,a)}
 +\frac{\int_{a_-}^a[\gamma(b)-\bar\gamma(t)]\rho(t,b)\,db}{\rho(t,a)},
\end{equation}
where $\bar\beta(t)=\int_A\beta\rho\,da$ and $\bar\gamma(t)=\int_A\gamma\rho\,da$. Differentiating $\mu_{\mathrm{ep}}\rho$ gives $-\partial_a(\mu_{\mathrm{ep}}\rho)=\rho(m-\bar m)$; the flux vanishes at both endpoints because this reaction has zero integral. Hence \eqref{eq:FP_full} is equivalent to \eqref{eq:normalized_heterogeneity}. The flux remains meaningful when a density vanishes, but the velocity in \eqref{eq:mu_ep} then requires care. It replaces the normalized selection term and must not be added again to an infected-density equation that already contains $mI$.

The transport and selection terms in \eqref{eq:FP_full} have complementary geometric interpretations. We now specialize to constant $\sigma>0$ and a potential drift $\mu_b=-\partial_aU$ in order to identify their energies explicitly.

\noindent\emph{1. Wasserstein (transport) geometry. Micro evolution dynamics.} The diffusion and baseline drift terms together can be interpreted as a gradient flow with respect to the Wasserstein metric. Consider the case without epidemic forces, where the trait evolves according to the Langevin dynamics
\begin{equation}
	da_t = -\partial_a U(a_t)\,dt + \sigma\,dW_t+dK_t, \qquad \mu_b(t,a)=-\partial_a U(a),
\end{equation}
with a potential function $U(a)$ and constant diffusion coefficient $\sigma$. The corresponding Fokker-Planck equation for the probability density $\rho(t,a)$ is
\begin{equation}
	\frac{\partial \rho}{\partial t} = \nabla\cdot\big(\rho\,\nabla U\big) + \frac{\sigma^2}{2}\Delta \rho.
\end{equation}
This equation is precisely the Wasserstein gradient flow of the \emph{free energy} functional
\begin{equation}
	\boxed{\mathcal{F}_1(\rho) = \frac{\sigma^2}{2}\int_A \rho(a)\log \rho(a)\,da + \int_A U(a) \rho(a)\,da.}
\end{equation}
Indeed, the first variation of $\mathcal{F}_1$ is $\delta\mathcal{F}_1/\delta\rho=(\sigma^2/2)(\log\rho+1)+U$, and the Wasserstein gradient flow $\partial_t\rho = \nabla\cdot(\rho\nabla(\delta\mathcal{F}_1/\delta\rho))$ reproduces the above Fokker-Planck equation. The associated Onsager operator (metric isomorphism) is
\begin{equation}
	\mathcal{M}_W^{-1}(\Phi) = -\nabla\cdot(\rho\nabla\Phi).
\end{equation}
Thus, in the absence of epidemic forces, the zero-flux equilibrium is
\[
\rho_\infty=Z^{-1}\exp(-2U/\sigma^2),\qquad
Z=\int_A\exp(-2U/\sigma^2)\,da.
\]
The normalization constant is finite for a continuous potential on the bounded trait interval. This density minimizes $\mathcal F_1$. The Wasserstein geometry captures both the diffusive spreading (entropy increase) and the drift toward lower potential (energy decrease).

\noindent\emph{2. Fisher-Rao (reaction) geometry. Macro selection dynamics.} The epidemic-induced drift term $-\partial_a(\mu_{\text{ep}} \rho)$ can be reformulated as a gradient flow with respect to the Fisher-Rao (Hellinger) metric. Define the "epidemic potential"
\begin{equation}\boxed{
		\mathcal{F}_2(t,\rho) = -\int_A m(t,a)\rho(a)\,da, \qquad
		m(t,a) = \frac{S(t)}{N}\,\beta(a) - \gamma(a).}
\end{equation}
Then the replicator-type equation
\begin{equation}
	\frac{\partial \rho}{\partial t} = \rho\big(m(t,a)-\langle m\rangle_\rho\big),
\end{equation}
where $\langle m\rangle_\rho=\int_A m\rho\,da$, is the descending Fisher--Rao gradient flow of $\mathcal F_2(t,\rho)$ with $S(t)$ held fixed when taking the variation. Its first variation is $-m(t,a)$. The corresponding Onsager operator is
\begin{equation}
	\mathcal{M}_{FR}^{-1}(\Phi) = \rho\big(\Phi - \langle\Phi\rangle_\rho\big).
\end{equation}
This component drives the distribution toward higher values of the epidemic fitness $m(t,a)$; it selects for traits that increase transmission or reduce recovery.

\noindent\emph{3. Combined gradient flow.} For the constant-diffusion, potential-drift setting above, the complete heterogeneity dynamics \eqref{eq:FP_full} can be written as a sum of descending gradient flows under different metrics:
\begin{equation}\label{eq:combined_gradient_flow}
	\boxed{\;
		\frac{\partial \rho}{\partial t} = -\nabla_{W}\mathcal{F}_1(\rho) \;-\; \nabla_{FR}\mathcal{F}_2(t,\rho),
		\;}
\end{equation}
where $\nabla_W$ and $\nabla_{FR}$ denote the gradients with respect to the Wasserstein and Fisher--Rao metrics, respectively. The two components use different energies, and $\mathcal F_2$ depends on the evolving susceptible population. A common Lyapunov functional or a single WFR energy therefore requires additional compatibility conditions. The reformulation \eqref{eq:combined_gradient_flow} shows how heterogeneity evolves through competing transport and selection effects during an epidemic. The Fisher-Rao component tends to concentrate probability mass on traits that maximize epidemic fitness (a form of \emph{natural selection}), while the Wasserstein component spreads probability mass diffusively, promoting diversity and entropy.  The geometric viewpoint connects heterogeneity models to a broad literature on gradient flows in probability space, opening avenues for analysis of long-time behavior, stability, and model reduction.

Notably, this framework extends beyond infectivity to other heterogeneities. For instance, the evolution of susceptibility in an epidemic model can be described by an analogous equation \cite{berestycki2023epidemic}, where the Fisher-Rao potential now depends on the force of infection and the trait-dependent susceptibility. Many trait-structured models in epidemiology, ecology, or cell population dynamics admit related transport and selection decompositions on the space of probability densities. This geometric perspective unifies macroscopic deterministic selection with microscopic stochastic variation, revealing a deep mathematical structure that underlies a wide class of models. The key insight is that trait distributions (e.g., infectivity, susceptibility, resource uptake rate, phenotypic state) evolve under two competing forces: (i) a \emph{diffusive} force that spreads probability mass and maximizes entropy, captured by the Wasserstein gradient of a free energy; and (ii) a \emph{selective} force that concentrates probability on traits with higher fitness, captured by the Fisher-Rao gradient of a potential function. From a first-principles physics viewpoint, density gradient flows arise naturally as the large-scale limit of individual-based stochastic processes. They offer a powerful language for bridging scales, from microscopic agent-based rules to macroscopic PDE descriptions.

\medskip
\noindent\emph{Numerical illustration.}
To illustrate the two contributions in \eqref{eq:combined_gradient_flow}, we solve the structured SIR system on $A=[0,1]$ with
\[
\beta(a)=0.18+0.50a,\qquad \gamma(a)=0.14,\qquad
U(a)=0.02(a-0.25)^2,\qquad \frac{\sigma^2}{2}=4\times10^{-4}.
\]
Time is measured in days and the trait is dimensionless. We normalize the population to $N=1$ and set $S(0)=0.99$, $\bar I(0)=0.01$, $R(0)=0$, with $\rho(0,a)$ proportional to $\exp[-(a-0.30)^2/(2\times0.10^2)]$ on $[0,1]$. The combined case uses the drift $-U'(a)$ and reflecting diffusion; the selection-only case sets both drift and diffusion to zero. The transmission and recovery rates, initial state, and observation times are identical in the two simulations.

\begin{figure}[!tbp]
\centering
\includegraphics[width=\textwidth]{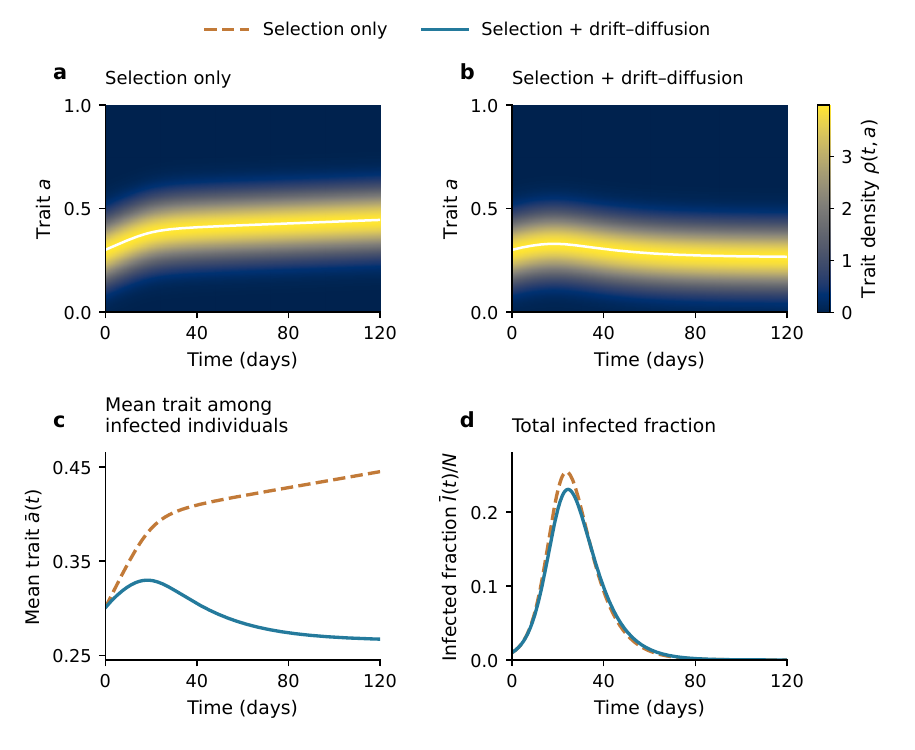}
\caption{Selection and drift--diffusion in the trait-structured SIR model. (a,b) The normalized trait density among infected individuals, $\rho(t,a)=I(t,a)/\bar I(t)$, under selection only and selection plus drift--diffusion, respectively. The heatmaps share a density scale; white curves mark the mean trait $\bar a(t)=\int_A a\rho(t,a)\,da$. (c) The corresponding mean traits $\bar a(t)$ within the infected population. (d) Total infected fractions $\bar I(t)/N$. In the combined case, trait transport consists of the drift $-U'(a)$ and diffusion with coefficient $\sigma^2/2$, with zero flux at the trait boundaries; both terms are absent in the selection-only case. Transmission and recovery rates and initial conditions are identical. All curves are deterministic numerical solutions, with no statistical uncertainty bands.}
\label{fig:heterogeneity_numerics}
\end{figure}

We discretize the unnormalized infected density $I(t,a)$ on 201 uniform finite-volume cells. For the combined case, the interior flux is
\[
\begin{aligned}
J_{j+1/2}&=\frac{D}{\Delta a}
\left[B(z_j)I_j-B(-z_j)I_{j+1}\right],\\
z_j&=\frac{U(a_{j+1})-U(a_j)}{D},\qquad
B(z)=\frac{z}{e^z-1},\qquad D=\frac{\sigma^2}{2},
\end{aligned}
\]
where $B(0)=1$ and the endpoint fluxes vanish. This exponentially fitted flux preserves the discrete Gibbs equilibrium of the transport operator. An adaptive backward differentiation formula integrates the coupled equations for $S$, $I_j$, and $R$, with relative and absolute tolerances $2\times10^{-9}$ and $2\times10^{-12}$, respectively. The density $\rho$ is recovered by normalization at each output time.

Figure~\ref{fig:heterogeneity_numerics} shows the infected-trait distributions and their means in panels (a)--(c), and the total infected fractions in panel (d). Epidemic selection shifts the distribution toward higher transmission traits, whereas the baseline drift and diffusion moderate this shift. At day 120, the mean trait is $0.4451$ under selection alone and $0.2671$ with drift--diffusion; the peak infected fractions are $0.2541$ and $0.2307$, respectively. These are illustrative consequences of the stated coefficients, rather than a general comparison theorem. Increasing the grid from 201 to 401 cells changes the combined-case mean trait by at most $1.82\times10^{-5}$ and the infected fraction by at most $7.18\times10^{-6}$ over the simulated interval. The maximum population-conservation error is $1.6\times10^{-15}$. In a separate transport-only check, the free energy decreases at every recorded time and the density at day 300 is within $1.61\times10^{-6}$ in discrete $L^1$ distance of the normalized Gibbs density $e^{-U/D}$.

\FloatBarrier
\section{Conclusion and Discussion}
\label{sec:conclusion}

In this paper, we have developed a unified mathematical and computational framework for quantifying and managing uncertainty in mechanism-AI coupled dynamical systems. By reformulating the uncertainty quantification problem as the minimization of an energy functional on a probability manifold subject to dynamical constraints, we established a rigorous connection between optimal transport, information geometry, and gradient flows. We systematically derived gradient flow formulations under various metrics---including Wasserstein, Fisher-Rao, Wasserstein--Fisher--Rao, Sobolev, and Stein geometries---and provided corresponding numerical schemes for their solution. This framework not only generalizes existing Bayesian inference methods but also provides a principled approach to learning distributions in dynamical systems with uncertain parameters and learned model components.

The main formulation of this work lies in the integration of dynamical system constraints directly into the geometric flow framework. By defining an energy functional that incorporates both data fidelity and the physical laws encoded by universal differential equations (UDEs), we introduced the concept of \emph{dynamical system-constrained probability gradient flows}. This approach ensures that the evolution of the probability distribution over model parameters respects the underlying biological or physical dynamics, leading to more interpretable and physically consistent uncertainty estimates. Furthermore, we developed efficient particle-based implementations leveraging adjoint sensitivity analysis, providing a route to gradient evaluation for large deterministic models.

The numerical comparisons show that a common geometric formulation does not by itself imply uniform computational superiority. At the tested budget, MALA gives the most accurate time-averaged approximation in the two conditional last-layer problems. The prediction experiments also expose uncertainty omitted by fixing the hidden representation, and the heterogeneity example separates selection from trait transport.

Despite these advances, several limitations warrant consideration. The computational cost of solving constrained gradient flows, especially in high dimensional parameter spaces, remains significant and may limit real-time applications. The performance of kernel-based methods such as SVGD is sensitive to the choice of kernel and bandwidth, which poses challenges in poorly understood systems without prior geometric knowledge. Additionally, the current framework assumes known noise models and observation structures; uncertainty in the observation and noise models themselves requires further investigation. Available convergence and consistency results depend on assumptions about the energy, geometry, and discretization that must be checked for highly nonlinear, non-identifiable, or chaotic systems.

Beyond the limitations outlined above, this paper itself raises numerous questions that merit further exploration. First, we have not performed rigorous theoretical analysis of the algorithms, such as convergence rates, computational complexity, consistency, or upper and lower complexity bounds. Second, the present work focuses on posterior or generalized-posterior learning; point optimization serves only as a preliminary step in the numerical examples, and classical parameter-estimation theory is not developed. Third, parameter identifiability (i.e., whether model parameters can be uniquely determined from the data) and model selection or averaging \cite{miao2011identifiability,hong2020global,claeskens2008model} are not addressed, despite their importance for model interpretation and robust inference. Fourth, we have not covered simulation-based inference methods \cite{martin2024approximating,lueckmann2021benchmarking,sisson2018handbook,cranmer2020frontier} nor generative models (e.g., diffusion models, flow matching \cite{lai2025principles}) or conditional optimal transport \cite{hosseini2025conditional,trigila2016data}, which are rapidly advancing frontiers. Fifth, the case of function-valued parameters (e.g., time-varying parameters in Bayesian inversion \cite{stuart2010inverse,idier2013bayesian}) is not considered; although such problems can be transformed into our framework via neural representations, a dedicated investigation is planned for future work. Finally, we have not explored other observation settings, including privacy-constrained data, chaotic dynamics, or streaming data. These directions all merit systematic study in subsequent research.

\bigskip
\section*{Acknowledgements}

The authors thank the referees for their comments. 

\bibliographystyle{doi_only}
\bibliography{references}

\StartSupplement
\SupplementTitle

\section{Problem Formulation}
\label{formulation}

To establish a rigorous foundation for uncertainty quantification in mechanism-AI coupled dynamical systems, we first systematically dissect the sources and mathematical representations of uncertainty. We then illustrate these concepts with concrete examples from biological modeling, clarifying the exact challenges that each type of uncertainty poses. 

\subsection{Uncertainties in Biological Modeling}
\label{subsec:uncertainty_types}

A principled treatment of uncertainty begins with its systematic classification. In the context of mechanism-AI coupled dynamical systems, we categorize uncertainty into three primary types, each with distinct origins and implications for model reliability.

\textbf{Data uncertainty}, also termed aleatoric uncertainty, arises from the inherent randomness, noise, and variability in the observational process itself. This type of uncertainty is irreducible, meaning it cannot be eliminated by collecting more data from the same experimental process; it is a property of the system under study \cite{hullermeier2021aleatoric}. In biological systems, this manifests profoundly. The data acquisition process is a primary source, susceptible to instrument errors, inappropriate sampling frequencies, and data transmission artifacts. For example, data uncertainty can be mathematically represented as follows:
\begin{equation}
	\left\{
	\begin{aligned}
		& \frac{\mathrm{d}x}{\mathrm{d}t}=f(t,x),\\
		& y=h(x)+\epsilon,
	\end{aligned}
	\right.
\end{equation}
where $x \in \mathbb{R}^{n}, y \in \mathbb{R}^{m}$, $\epsilon \sim \mathcal{N}(0,\sigma^2 I)$.
A particularly representative formulation in this domain is the Universal Differential Equations (UDEs) framework,
\begin{equation}\label{sm:eq0}
	x' = f_{\eta}(x, t, \mathrm{UniversalApprox}_{\phi}(x, t)),
\end{equation}
where \( f \) encapsulates prior mechanistic knowledge (e.g., from physics or biology) and \( \mathrm{UniversalApprox} \) is a flexible approximator (e.g., a neural network (NN)~\cite{lecun2015deep}, Kolmogorov-Arnold Network (KAN)~\cite{liu2024kan}, or Gaussian process (GP)~\cite{rasmussen2003gaussian}) used to learn unknown mechanisms.

\textbf{Model uncertainty}, or epistemic uncertainty, stems from a lack of knowledge about the correct model structure or functional form. It reflects our ignorance about the fundamental mechanisms governing the system. This is reducible in principle with more informative data or improved scientific understanding. In biological dynamics, this uncertainty arises when critical biochemical pathways, regulatory feedback loops, or cellular interactions are unknown or oversimplified in the mechanistic component $f_{\eta}(u,t)$ of UDE \eqref{sm:eq0}. The Universal Differential Equation framework partially addresses this by using the universal approximator $\mathrm{UniversalApprox_{\phi}}(u,t)$ of UDE \eqref{sm:eq0} to represent the unknown mechanisms. However, this introduces a new form of epistemic uncertainty: which architecture is appropriate for $\mathrm{UniversalApprox}_{\phi}$, and is the chosen structure capable of capturing the true missing mechanism? Lack of knowledge from model structures is often handled by model selection, model averaging or stacking methods (\cite{claeskens2008model}), focusing on model likelihood (such as Bayes factors) or posterior
\[\rho(D|\mathrm{Model}), \quad \rho(\mathrm{Model}|D).\]

Another more fundamental formulation of model uncertainty involves treating the model itself as stochastic. We call it \textbf{Random Model}. It will involve
\begin{itemize}
	\item Discrete-time, discrete-state Markov chain, random difference equations
	\[
	X_{t+1} \sim P(X_{t+1} \mid X_t, \theta), \quad X_t \in \mathcal{X} \text{ (finite or countable)}, \quad t \in \mathbb{N},
	\]
	where \(P(\cdot \mid x, \theta)\) is a transition kernel parameterized by \(\theta\). Such models are common in population genetics, epidemic models with discrete generations, and cellular state machines. For example, consider a stochastic difference equation
	\[
	X_{t+1} = X_t + \sum_{j=1}^m \nu_j \, N_{j,t}\big(\Lambda_j(X_t, \theta)\big), \quad X_t \in \mathbb{Z}_{\ge 0}^n,
	\]
	where \(N_{j,t}\) are independent unit-rate Poisson processes, so the conditional event-count means are \(\Lambda_j(X_t,\theta)\), and \(\nu_j\in\mathbb{Z}^n\) are stoichiometric vectors. Fresh counts are drawn at each time step. This is a Poisson tau-leap approximation; it does not guarantee nonnegative populations unless the event counts are controlled.
	
	Example (Stochastic difference SIR model in discrete time): Let \(S_t,I_t,R_t\) denote susceptible, infected, and recovered populations. A nonnegative finite-population discretization uses conditionally independent binomial event counts. Over one step of length \(\Delta t\), new infections satisfy
	\[
	\Delta I_t\mid(S_t,I_t)\sim\mathrm{Binomial}\left(S_t,1-e^{-\beta I_t\Delta t/N}\right),
	\]
	and recoveries follow:
	\[
	\Delta R_t\mid I_t\sim\mathrm{Binomial}\left(I_t,1-e^{-\gamma\Delta t}\right).
	\]
	The state update is:
	\[
	\begin{cases}
		S_{t+1} = S_t - \Delta I_t,\\
		I_{t+1} = I_t + \Delta I_t - \Delta R_t,\\
		R_{t+1} = R_t + \Delta R_t.
	\end{cases}
	\]
	This update preserves \(S_t+I_t+R_t=N\) and nonnegativity. Uncertainty arises from the event counts and unknown parameters \(\beta,\gamma\), compounded by partial and noisy observations. Inference requires propagating the distribution of the discrete population state.
	
	\item Discrete-time, continuous-state model, state space models (\cite{sarkka2023bayesian,chopin2020introduction})
	\[
	X_{t+1} = f(X_t, \theta) + \xi_t, \quad \xi_t \sim \mathcal{N}(0, \Sigma), \quad X_t \in \mathbb{R}^n.
	\]
	often observed through \(Y_t = h(X_t) + \varepsilon_t\). Used for time-series data with fixed sampling intervals, e.g., viral load dynamics, ecological abundance time series. Filtering theory provides a principled approach to sequentially estimate the posterior distribution \(\rho(X_t \mid Y_{1:t})\) as new data arrive. 
	Classical methods such as the Kalman filter (for linear Gaussian models), particle filters (for nonlinear/non-Gaussian models), and the ensemble Kalman filter (for high-dimensional systems) have been widely employed. 
	However, these methods often struggle with high-dimensional parameter spaces, strongly nonlinear dynamics, or when the model itself contains uncertain AI-augmented components (\cite{sarkka2023bayesian,chopin2020introduction}).
	
	\item Continuous-time, discrete-state Markov chain (CTMC)
	Governed by the chemical master equation (\cite{gardiner2004handbook,qian2021stochastic}):
	\[
	\frac{\partial p(x,t)}{\partial t} = \sum_{j=1}^R \big[ a_j(x-\nu_j) p(x-\nu_j,t) - a_j(x) p(x,t) \big], \quad x \in \mathbb{Z}_{\ge 0}^n,
	\]
	where \(a_j(x)\) are propensity functions and \(\nu_j\) stoichiometric vectors. Propensities vanish for jumps that would leave the nonnegative lattice, and terms evaluated outside that lattice are set to zero. The exact process can also be written using Poisson random measures (Kurtz representation):
	\[
	\mathbf{X}(t) = \mathbf{X}(0) + \sum_{j=1}^R \bm{\nu}_j \int_0^t \int_0^{\infty} \mathbb{1}_{\{0 \leq u \leq a_j(\mathbf{X}(s^-))\}} N_j(ds, du),
	\]
	where $N_j(ds, du)$ are independent Poisson random measures with intensity $ds\, du$. CTMC models are well known in stochastic biochemical reaction networks, gene expression bursts, and ion-channel dynamics.
	
	\item Continuous-time, continuous-state diffusion (SDE) (\cite{evans2012introduction,sarkka2019applied,oksendal2003stochastic}).
	\[
	dX_t = f(X_t, \theta) dt + \sigma(X_t, \theta) dW_t, \quad X_t \in \mathbb{R}^n
	\]
	with \(W_t\) a Wiener process. Provides a coarse-grained description of population dynamics, neural firing rates, or trait evolution under random environmental fluctuations.
\end{itemize}

\textbf{Parametric uncertainty} concerns the imperfect knowledge of the parameters within a chosen model structure. Even if the model form 
in UDE \eqref{sm:eq0} is correct, the precise values of the parameters $\eta$ (mechanistic) and $\phi$ (AI) may be unknown. This uncertainty arises from several factors: i) \textit{Identifiability} (\cite{miao2011identifiability,hong2020global}): Parameters may be non-identifiable from the available data, meaning different combinations of parameter values can yield identical model outputs. ii) \textit{Inference from finite data}: With limited and noisy datasets, we can only estimate a distribution over plausible parameter values rather than a single point estimate. This is often formalized in a Bayesian framework, where the parameters $\eta$ and $\phi$ are treated as random variables with a prior distribution $\rho(\eta, \phi)$ that is updated by data to a posterior distribution $\rho(\eta, \phi | \mathcal{D})$. The spread of this posterior quantifies the parametric uncertainty. In high-dimensional coupled models, this posterior is often complex and correlated, making its characterization a significant computational challenge.

Another parametric uncertainty, we call it \textbf{Random Parameter}, refers to the scenario where the model parameters themselves are intrinsically random variables. This concept moves beyond the Bayesian interpretation of uncertainty as a state of knowledge, and instead posits that the parameters \(\eta\) and \(\phi\) possess an inherent stochasticity. Consequently, even with a perfectly specified, deterministic model structure 
\[
\frac{\mathrm{d}x}{\mathrm{d}t} = f_{\eta}(x,t) + \mathrm{UniversalApprox}_{\phi}(x,t), \quad \text{random parameters},
\]
and noiseless observational data,
\[y=h(x), \quad \text{deterministic observation,}\]
the model output would still exhibit variability due to the random fluctuations in the parameters themselves, i.e., the laws of $x$ and $y$ are pushforwards of $\mu_{\eta,\phi}$, where $\Phi_t$ is the parameter-to-state map
\[\mu_{x(t)}=(\Phi_t)_{\#}\mu_{\eta,\phi},\qquad \mu_{y(t)}=(h\circ\Phi_t)_{\#}\mu_{\eta,\phi}.\]
This phenomenon is prevalent in biological systems where parameters often represent physical quantities that are subject to temporal variation, spatial heterogeneity, or unobserved external stimuli. For example, a kinetic rate constant might fluctuate due to stochastic cellular environments, or a connection weight in the neural network might represent a process with inherent randomness. Formally, this is characterized by treating the parameters as stochastic variables or stochastic processes, 
$\eta(t, \omega)$ and $\phi(t, \omega)$, where \(\omega\) denotes an elementary event in a probability space. The resulting system dynamics then become a conditional process, and the objective shifts from learning a fixed posterior distribution \(\rho(\eta, \phi | \mathcal{D})\) to characterizing the law of the time-varying random parameters that govern the system's behavior. This intrinsic randomness at the parameter level presents a distinct and profound challenge for uncertainty quantification, as it necessitates methods that can disentangle and propagate this specific source of variability through the coupled mechanistic-AI dynamical system.

\subsection{Formulation of Probability Energy Functional Minimization Problems}
\label{subsec:energy-formulation}

To systematically address the various uncertainty scenarios outlined above, we cast each inference task as a minimization problem over probability measures. Let \(\Theta\) denote the parameter space, typically \(\Theta = \mathbb{R}^d\) or a subset thereof. The forward model is described by a general dynamical system
\[
\frac{\mathrm{d}x}{\mathrm{d}t} = f(x, \theta), \quad x(0) = x_0, \quad \theta \in \Theta,
\]
The displayed equation is the deterministic ODE case; an SDE or other stochastic forward model is specified by its solution law. The observation process is \(y=h(x)+\varepsilon\), where \(\varepsilon\) captures measurement noise. If the unknown parameter is a time-dependent random process, its law is defined on a suitable path space rather than on \(\mathbb{R}^d\).

We distinguish four classes of problems based on whether the \emph{parameters are inherently random} and whether the \emph{data/model are subject to uncertainty}. In each case, the goal is to minimize an energy functional of the form
\[
\mathcal{E}(\rho) = D\big( \rho,\, \rho^* \big),
\]
where \(D\) is a suitable divergence or distance between probability measures (\cite{amari2016information,ay2017information,villani2003topics,villani2008optimal,basu2011statistical,nielsen2022many,cichocki2010families,birrell2022f}), \(\rho\) is a candidate distribution over parameters (or over trajectories), and \(\rho^*\) is a target distribution dictated by the problem setting.

\subsubsection*{Case 1: Parameter estimation under data uncertainty (frequentist viewpoint)}
Here the parameter \(\theta\) is deterministic without inherent randomness but unknown, while observations are noisy and the model is deterministic or random. The task reduces to minimizing a discrepancy between the observed data distribution and the model-predicted distribution:
\[
\min_{\theta \in \Theta} \; D\Big( P_{\mathrm{data}}(y),\; P(y \mid \theta) \Big),
\]
where \(P(\cdot\mid\theta)\) is the observation law conditional on \(\theta\). For a deterministic model with additive noise, it is the pushforward of the noise law by \(\varepsilon\mapsto H(\Phi_\theta)+\varepsilon\). Population KL minimization, when defined, is equivalent to minimizing expected negative log likelihood; its empirical counterpart is the sample negative log likelihood, not the KL divergence between an atomic empirical measure and a continuous density. Gaussian observation errors yield a least-squares loss. Wasserstein discrepancies provide another choice (\cite{frogner2015learning,adler2017learning}).

\subsubsection*{Case 2: Bayesian inference under data uncertainty}
The parameter is again treated as deterministic in reality, but within a Bayesian framework it is modeled as a random variable with a prior \(\rho_0(\theta)\). The model may be deterministic or stochastic. Given noisy data \(y_{\mathrm{obs}}\), the goal is to approximate the posterior \(\rho(\theta \mid y_{\mathrm{obs}})\) or \textbf{generalized posterior} \(\rho^{*}\) (\cite{knoblauch2022optimization}). This can be formulated as minimizing a divergence between a variational distribution \(\rho(\theta)\) and the (generalized) posterior:
\[
\min_{\rho \in \mathcal{Q}} \; D\Big( \rho,\; \rho(\theta \mid y_{\mathrm{obs}}) \;(\text{or}\; \rho^{*}) \Big),
\]
where \(\mathcal{Q}\) is a tractable family of distributions. When \(D\) is the KL divergence, this recovers standard variational inference. For other divergences (e.g., \(\alpha\)-, \(\beta\)-, or Stein discrepancies), one obtains robust or geometry-aware generalizations (\cite{amari2016information,ay2017information,villani2003topics,villani2008optimal,basu2011statistical,nielsen2022many,cichocki2010families,birrell2022f}).

\subsubsection*{Case 3: Recovering a random parameter from perfect data and model}
Now the parameter \(\theta\) is intrinsically random (e.g., a random variable or a time-varying stochastic process), but the model \(f(x,\theta)\) and the observation operator \(H\) are known exactly, and data are noise-free. The problem becomes one of recovering the ground-truth law \(\rho_{\mathrm{true}}\) of \(\theta\) from perfect trajectory observations, described as a least-squares problem over a space of probability measures in \cite{li2025least}. Let \(\Psi: \Theta \to \mathcal{Y}\) be the deterministic map from a parameter realization to the corresponding noise-free observation. Then the ideal data distribution is the push-forward \(\Psi_\# \rho_{\mathrm{true}}\). Given empirical observations \(y_i = \Psi(\theta_i)\) with \(\theta_i \sim \rho_{\mathrm{true}}\), we seek
\[
\min_{\rho \in \mathcal{P}(\Theta)} \; D\Big( \Psi_\# \rho,\; \widehat{P}_{\mathrm{data}} \Big),
\]
where \(\widehat{P}_{\mathrm{data}}\) is the empirical distribution of the \(y_i\). This is an inverse problem in distribution space, often tackled via optimal transport or moment-matching methods (\cite{li2024differential,li2025least}).

\subsubsection*{Case 4: Recovering a random parameter under data/model uncertainty}
The most challenging scenario combines intrinsic parameter randomness with imperfect data and/or model misspecification. Suppose the true parameter follows an unknown law \(\rho_{\mathrm{true}}\), the forward model is only approximately known (e.g., contains an AI-augmented component with its own uncertainties), and observations are contaminated by noise. The energy functional must simultaneously account for all sources of discrepancy:
\[
\min_{\rho \in \mathcal{P}(\Theta)} \; \underbrace{D\Big( \mathcal{G}\rho,\; P_{\mathrm{data}} \Big)}_{\text{data mismatch}} \;+\; \underbrace{\lambda_1 R_1(\rho)}_{\text{prior/regularization}}.
\]
Here \(\mathcal{G}\) maps a parameter law to an observation law through a specified conditional kernel \(Q(dy\mid\theta)\): \((\mathcal{G}\rho)(B)=\int_\Theta Q(B\mid\theta)\rho(d\theta)\). The kernel includes the stochastic dynamics and observation noise; \(P_{\mathrm{data}}\) is the noisy observation law, and \(R_1\) imposes prior knowledge or regularity. Recoverability depends on the identifiability of this map and on how model uncertainty is specified. Generative modeling, including diffusion models and flow matching, offers possible approaches to such problems \cite{lai2025principles}.

\subsubsection*{Connections between Distribution Learning and Parameter Estimation}
\label{subsubsec:dist-param-connection}

Mathematically, handling uncertainties is concerned with two main problems, \textbf{Distribution Learning} and \textbf{Parameter Estimation}. The problem of learning a distribution \(\rho\) over parameters can be recast as a parameter estimation problem through two primary approaches:

Restriction to a parametric density submanifold: Constrain \(\rho\) to belong to a finite-dimensional parametric family \(\{\rho_\xi : \xi \in \Xi \subset \mathbb{R}^m\}\), such as Gaussian, exponential family, or mixture models. The distribution learning problem then reduces to estimating the parameter \(\xi\):
\[
\min_{\xi \in \Xi} \; D\Big( \rho_\xi,\; \rho^{*} \Big) + \text{regularization}.
\]
However, a crucial geometric subtlety arises: the parametric model typically does not form a \emph{totally geodesic submanifold} of the full density manifold \(\mathcal{P}(\Theta)\). The notable exception is the Gaussian case under the Wasserstein metric, where the family of Gaussian distributions is geodesically convex, enabling closed-form updates and exact computations of the Wasserstein distance (\cite{takatsu2011wasserstein,malago2018wasserstein,lambert2022variational}). For other parametric families or different metrics (Fisher-Rao, Stein), the submanifold structure is generally not preserved under the Riemannian geometry induced by the metric, complicating both theoretical analysis and algorithmic design.

Pushforward parameterization via transport maps (We call it \emph{second parameterization}): Represent the target distribution as the pushforward of a simple reference measure \(\rho_{\text{ref}}\) (e.g., standard Gaussian) through a parametric transformation \(T_\eta : \Theta \to \Theta\):
\[
\rho_\eta = (T_\eta)_\# \rho_{\text{ref}}.
\]
The parameters \(\eta\) then govern the shape of \(\rho\). This framework encompasses:
\begin{itemize}
	\item Normalizing flows (\cite{papamakarios2021normalizing}): \(T_\eta\) is an invertible neural network with tractable Jacobian.
	\item Generative models (\cite{lai2025principles}): \(T_\eta\) can be the decoder of a VAE or the reverse process of a diffusion model.
	\item Optimal transport maps (\cite{villani2003topics,villani2008optimal,santambrogio2015optimal}): \(T_\eta\) approximates the Brenier map in Wasserstein geometry.
	\item Reparameterization trick: The widely used \emph{reparameterization trick} in variational autoencoders and Bayesian deep learning provides a concrete instantiation of the second parameterization (\cite{kingma2015variational}). Instead of directly parameterizing the variational distribution \(\rho(\theta)\) (first parameterization), one parameterizes a deterministic transformation \(T_\phi\) that maps samples from a fixed base distribution \(\rho(\epsilon)\) to the target distribution:
	\[
	\rho(\theta) = (T_\phi)_\# \rho(\epsilon), \quad \theta = T_\phi(\epsilon), \ \epsilon \sim \rho(\epsilon).
	\]
\end{itemize}

This dual viewpoint clarifies that ``learning distributions'' and ``estimating parameters'' are not fundamentally distinct—rather, they represent different levels of abstraction in modeling uncertainty, with the former emphasizing the geometric structure of the probability space and the latter focusing on computational tractability through finite-dimensional approximations.

\section{Preliminaries}
\label{sec:preliminary}

To establish a rigorous foundation for the geometric gradient flow framework developed in subsequent sections, we first review essential mathematical concepts and notation. This section serves three purposes: (1) to make the presentation self-contained for readers from diverse backgrounds (mathematical biology, dynamical systems, statistics, machine learning); (2) to fix notation and terminology that will be used throughout the paper; and (3) to highlight the specific mathematical structures—Riemannian geometry, optimal transport, and generalized Bayesian inference—that underpin our constrained gradient flow methodology. While some of this material is standard, we emphasize the connections and interpretations most relevant to uncertainty quantification in mechanism--AI dynamical systems.

The exposition proceeds from geometry to probability: we begin with Riemannian manifold essentials, then introduce distances and divergences between probability measures, followed by optimal transport formulations, and conclude with a modern perspective on generalized Bayesian inference via the ``Rule of Three.'' Readers familiar with these topics may skip directly to Section~\ref{sec:gradient-flow}.

\subsection{Basics of Riemannian Geometry}
\label{subsec:riemannian-basics}

Riemannian geometry provides the natural language for describing gradient flows on curved spaces. In this subsection, we outline the essential concepts; a more comprehensive introduction can be seen  in \cite{petersen2006riemannian} for example.

\begin{definition}
	A \textbf{Riemannian manifold} $(M,g)$ is a smooth manifold $M$ equipped with a \textbf{Riemannian metric} $g$, which is a smoothly varying family of positive-definite inner products on tangent spaces:
	\[
	g_p: T_p M \times T_p M \rightarrow \mathbb{R}, \quad \forall p \in M.
	\]
	In local coordinates $(x^1,\dots,x^n)$, the metric is expressed as:
	\[
	g = g_{ij}dx^i \otimes dx^j, \quad g_{ij} = g(\partial_i,\partial_j)
	\]
	with $g_{ij}$ symmetric and positive-definite.
\end{definition}

The \textbf{tangent bundle} $TM = \bigcup_{p\in M} T_p M$ and \textbf{cotangent bundle} $T^*M = \bigcup_{p\in M} T_p^* M$ provide the spaces for vector fields and differential forms. These spaces describe how a smooth function $f: M \rightarrow \mathbb{R}$ changes along a direction $\nu \in T_{p}M$. Let \(\gamma: (-\epsilon,\epsilon)\to M\) be a smooth curve with \(\gamma(0)=p\) and \(\gamma'(0)=\nu\). Then
\begin{equation}\label{eq1}
	\nu(f)=D_{\nu}f=df_p(\nu)
	=\left.\frac{d}{dt}f(\gamma(t))\right|_{t=0}
	=g_{p}(\nabla_{g}f,\nu).
\end{equation}

The metric induces an isomorphism between the tangent and cotangent spaces, often called the \emph{musical isomorphisms}:

\begin{itemize}
	\item \textbf{Flat (\(\flat\)) or metric tensor $\mathcal{M}_{p}$:} \(T_pM \to T_p^*M\) sends a vector \(v\) to the covector \(v^\flat=\mathcal{M}_{p}v\) defined by
	\begin{equation}\label{eq2}
		v^\flat(w) = g_p(v,w) = \langle \mathcal{M}_{p}v, w \rangle\quad \text{for all } w \in T_pM.
	\end{equation}
	
	\item \textbf{Sharp (\(\sharp\)) or Onsager operator $\mathcal{M}_{p}^{-1}$ (\cite{mielke2016generalization}):} \(T_p^*M \to T_pM\) is the inverse map. For a covector \(\alpha\), the vector \(\alpha^\sharp=\mathcal{M}_{p}^{-1}\alpha\) satisfies
	\begin{equation}\label{eq3}
		g_p(\alpha^\sharp, w) = \langle \alpha, w \rangle
		\quad \text{for all } w \in T_pM.
	\end{equation}
\end{itemize}

These isomorphisms allow us to identify vectors and covectors, a crucial step in writing gradient flow equations.
For a smooth function \(F: M \to \mathbb{R}\), its differential \(dF_p \in T_p^*M\) is defined by
\[
dF_p(v) = v(F) \quad \text{for } v \in T_pM.
\]
The \emph{Riemannian gradient} \(\nabla_g F(p) \in T_pM\) is then obtained by applying the sharp operator to the differential:
\[
\nabla_g F(p) = (dF_p)^\sharp=\mathcal{M}_{p}^{-1}dF_p.
\]
When \(M = \mathbb{R}^n\) with the Euclidean metric \(g_{ij} = \delta_{ij}\), this reduces to the usual gradient \(\nabla F\).

\noindent\emph{Onsager operator: from cotangent to tangent space.}
In non-equilibrium thermodynamics and information geometry, the evolution of a system is often driven by a \emph{thermodynamic force} (a covector) that produces a flux (a vector). The linear map relating force to flux is called the \emph{Onsager operator} (or mobility operator). We consistently denote it by \(\mathcal{M}_p^{-1}:T_p^*M\to T_pM\). It is symmetric and positive under the dual pairing: \(\langle\alpha,\mathcal{M}_p^{-1}\beta\rangle=\langle\beta,\mathcal{M}_p^{-1}\alpha\rangle\) and \(\langle\alpha,\mathcal{M}_p^{-1}\alpha\rangle>0\) for nonzero \(\alpha\) (\cite{mielke2016generalization}). The musical isomorphism satisfies \(\mathcal{M}_p^{-1}(v^\flat)=v\). More generally, the mobility may incorporate physical constants or state dependence.

\subsection{Distances and Divergences between Probability Measures}
\label{subsec:prob_distances}

Distances and divergences quantify discrepancies between probability measures and may be used to define an energy functional. The geometry that determines the gradient is a separate choice: for example, KL energy can be minimized using either the Wasserstein or the Fisher-Rao metric. Some divergences also induce local metric tensors under suitable regularity assumptions (\cite{amari2016information,ay2017information,villani2003topics,villani2008optimal,basu2011statistical,nielsen2022many,cichocki2010families,birrell2022f}). We briefly review the quantities used in this paper.

A critical component of many machine learning and statistical inference tasks is quantifying the difference between two probability measures \(P\) and \(Q\). While often colloquially referred to as ``distances,'' these measures have distinct properties and motivations. We review the key concepts used in this work. A \emph{distance} (or \emph{metric}) \(d\) must satisfy: (1) non-negativity: \(d(P, Q) \ge 0\); (2) identity of indiscernibles: \(d(P, Q) = 0 \iff P = Q\); (3) symmetry: \(d(P, Q) = d(Q, P)\); and (4) triangle inequality. A \emph{divergence} is a weaker concept, typically requiring only non-negativity and identity, but not symmetry or the triangle inequality.

Given a convex function \(f: \mathbb{R}_+ \to \mathbb{R}\) with \(f(1) = 0\), the \(f\)-divergence is defined as:
\[
D_f(P \parallel Q) = \int_{\Omega} f\left(\frac{p(\theta)}{q(\theta)}\right) q(\theta) d\theta,
\]
where \(P\ll Q\), and \(p\) and \(q\) are densities with respect to a common dominating measure. Singular parts require the standard extended-value convention. Convexity and \(f(1)=0\) ensure nonnegativity; an additional strictness condition, such as strict convexity at \(1\), is needed for \(D_f(P\parallel Q)=0\) to imply \(P=Q\).

Kullback-Leibler (KL) Divergence: \(f(t) = t \log t\).
\[
\text{KL}(P \parallel Q) = \int p(\theta) \log \frac{p(\theta)}{q(\theta)} d\theta.
\]

Wasserstein Distance (Earth Mover's): Let \(\Gamma(P, Q)\) be the set of all couplings of \(P\) and \(Q\). For \(p \ge 1\), the \(p\)-Wasserstein distance is:
\[
W_p(P, Q) = \left( \inf_{\gamma \in \Gamma(P, Q)} \int d(\theta, \theta')^p d\gamma(\theta, \theta') \right)^{1/p},
\]
where \(d(\cdot, \cdot)\) is a metric on the sample space \(\Omega\).

Stein Discrepancy: A specific integral probability metric where the function class is derived from a \emph{Stein operator} \(\mathcal{A}_P\) such that \(\mathbb{E}_{\theta \sim P}[\mathcal{A}_P f(\theta)] = 0\) for all \(f\) in a Stein class.
\[
\mathbb{S}(Q, P) = \sup_{f \in \mathcal{G}} \left| \mathbb{E}_{\theta \sim Q}[\mathcal{A}_P f(\theta)] \right|.
\]

\subsection{Definition of Optimal Transport}

In this subsection, we present the essential definitions of optimal transport (OT) that will be used throughout the paper. For a more comprehensive treatment, including historical development, alternative formulations, and extensive references, we refer readers to the classic texts \cite{santambrogio2015optimal,villani2003topics,villani2008optimal}.

The definitions below make explicit the transport and duality structures used in the gradient-flow formulas. Proofs and detailed hypotheses can be found in the cited references.

Optimal transport (OT) compares probability measures by minimizing the cost of rearranging mass (\cite{santambrogio2017euclidean,villani2003topics,villani2008optimal,santambrogio2015optimal,ambrosio2012user,figalli2021invitation,ambrosio2021lectures,ambrosio2007gradient,ambrosio2005gradient,peyre2019computational}). Let $X,Y$ be Polish spaces, $\mu,\nu$ Borel probability measures, and $c:X\times Y\to[0,\infty]$ a lower semicontinuous cost. When discussing finite optimal values, assume that at least one coupling has finite cost.

The oldest and most intuitive formulation, due to Gaspard Monge (1781), seeks a deterministic mass-preserving map.

\begin{definition}[Monge Formulation]
	The \emph{Monge optimal transport problem} is to find a Borel map $T: X \to Y$ that pushes $\mu$ forward to $\nu$, denoted $T_\# \mu = \nu$, and minimizes the total transportation cost:
	\begin{equation}
		\inf_{T} \left\{ \int_X c(x, T(x)) \, d\mu(x) \ : \ T_\# \mu = \nu \right\}.
	\end{equation}
	A map $T$ achieving the infimum is called an \emph{optimal transport map}.
\end{definition}

The constraint $T_\# \mu = \nu$ means that for any Borel set $B \subset Y$, $\nu(B) = \mu(T^{-1}(B))$, ensuring the map $T$ rearranges the mass from $\mu$ into $\nu$. While geometrically natural, this formulation is inherently nonlinear and an admissible map may not exist (e.g., when $\mu$ is a Dirac mass and $\nu$ is not), limiting its analytic tractability because the constraint on $T$ is not closed under weak convergence (see \cite[Chapter 1]{santambrogio2015optimal} for example).

Kantorovich's relaxation allows mass to be split, leading to a linear minimization problem with an optimal plan under the stated assumptions.
\begin{definition}[Kantorovich Formulation]
	The \emph{Kantorovich optimal transport problem} is to find a probability measure $\pi$ on $X \times Y$ (called a \emph{coupling} or \emph{transport plan}) with marginals $\mu$ and $\nu$, i.e., $\pi(A \times Y) = \mu(A)$ and $\pi(X \times B) = \nu(B)$ for all measurable sets $A, B$, which minimizes the total cost:
	\begin{equation}
		\inf_{\pi} \left\{ \int_{X \times Y} c(x, y) \, d\pi(x, y) \ : \ \pi \in \Pi(\mu, \nu) \right\},
	\end{equation}
	where $\Pi(\mu, \nu)$ denotes the set of all such couplings.
\end{definition}

Any transport map $T$ induces a plan $\pi=(\mathrm{id},T)_\#\mu$ concentrated on its graph. Weak compactness of the coupling set and lower semicontinuity of the cost ensure an optimal plan; they do not in general ensure an optimal map. The linear structure also supports a dual formulation.

\begin{definition}[Kantorovich Duality]
	The \emph{dual Kantorovich problem} is to find a pair of functions $(\varphi, \psi) \in L^1(\mu) \times L^1(\nu)$, called \emph{Kantorovich potentials}, that maximize the total payoff subject to a constraint preventing arbitrage:
	\begin{equation}
		\sup \left\{ \int_X \varphi(x) \, d\mu(x) + \int_Y \psi(y) \, d\nu(y) \ : \ \varphi(x) + \psi(y) \leq c(x, y) \right\}.
	\end{equation}
\end{definition}

Under the stated assumptions, the primal infimum and dual supremum agree. Attainment of the dual supremum requires additional conditions. Whenever optimal integrable potentials exist, $\varphi(x)+\psi(y)=c(x,y)$ holds $\pi$-almost everywhere for any finite-cost optimal plan $\pi$. The potentials can be interpreted as prices assigned to departure and arrival locations.

For the quadratic cost $c(x, y) = |x - y|^2$, Benamou and Brenier recast the problem as one of fluid flow, providing a dynamic interpretation of the Wasserstein distance.

\begin{definition}[Benamou-Brenier Formulation]\label{def:benamou-brenier}
	Let $\mu, \nu \in \mathcal{P}_2(\mathbb{R}^d)$. The \emph{dynamic formulation} defines the squared 2-Wasserstein distance as:
	\begin{equation}
		W_2^2(\mu,\nu)=\inf_{(\rho,v)}\int_0^1\int_{\mathbb{R}^d}|v_t(x)|^2\,d\rho_t(x)\,dt,
	\end{equation}
	where the infimum is taken over all pairs $(\rho_t, v_t)$ satisfying:
	\begin{itemize}
		\item The continuity equation (Eulerian perspective, density): $\partial_t \rho_t + \nabla \cdot (\rho_t v_t) = 0$ on $[0,1] \times \mathbb{R}^d$; 
		
		For sufficiently regular velocities, the corresponding Lagrangian representation is $\frac{dX}{dt}=v_t(X)$, with $\rho_t=\mathrm{Law}(X(t))$.
		\item The boundary conditions: $\rho_{0} = \mu$, $\rho_1 = \nu$; $X(0)\sim \mu, X(1) \sim \nu$.
	\end{itemize}
\end{definition}	

Here $\rho_t$ is a narrowly continuous curve of measures, and the continuity equation is understood weakly. If $\rho_t$ has a density, then $d\rho_t(x)=\rho_t(x)\,dx$. The action is also $\int_0^1\mathbb{E}|v_t(X(t))|^2dt$ whenever the Lagrangian representation is available. A minimizing path is a constant-speed Wasserstein geodesic; on the smooth density class, its minimal velocity has a potential representation $v_t=\nabla\Phi_t$.

These formulations provide complementary descriptions of transport. For quadratic cost on $\mathbb{R}^d$, finite second moments, and an absolutely continuous source, the Monge, Kantorovich, and dynamic formulations attain the same value. Without an appropriate source assumption, an optimal plan need not be induced by a map.

\subsection{Generalized Bayesian Inference and the Rule of Three}
\label{subsec:gbi_rot}

Bayesian inference updates prior beliefs \(\rho_0(\theta)\) through a likelihood \(p(x\mid\theta)\). Misspecification and computational constraints motivate broader optimization formulations. A generalized posterior can be defined, when a minimizer exists, through the variational problem (\cite{knoblauch2022optimization}):
\begin{equation}
	\rho^* \in \arg\min_{\rho \in \Pi} \left\{ \mathbb{E}_{\rho(\theta)} \left[ \ell(\theta, x_{1:n}) \right] + D(\rho, \rho_0) \right\},
	\label{eq:generalized_posterior}
\end{equation}
where \(\ell\) represents a general loss function, \(D\) denotes a divergence measure between distributions, and \(\Pi\) defines the space of admissible posterior distributions. \cite{knoblauch2022optimization} provides a \emph{Rule of Three (ROT)} formulation that formalizes this perspective by explicitly modularizing the three components that define the inference procedure:
\begin{itemize}
	\item Loss Function (\(\ell\)): Determines how model parameters relate to observed data. While traditional Bayes uses the negative log-likelihood \(\ell(\theta, x_i) = -\log p(x_i \mid \theta)\), the framework accommodates robust alternatives, scoring rules, or task-specific losses.
	
	\item Divergence (\(D\)): Controls the regularization strength and nature of prior influence. The Kullback-Leibler divergence recovers standard Bayesian updating, while alternative divergences (e.g., \(\alpha\)-, \(\beta\)-, or \(\gamma\)-divergences) provide robustness to prior misspecification.
	
	\item Feasible Set (\(\Pi\)): Encodes computational constraints. With negative log likelihood as loss and KL regularization relative to the prior, the full space \(\mathcal{P}(\Theta)\) recovers the usual posterior, provided its normalizing constant is finite and positive. Restricted families (e.g., mean-field approximations) yield variational approximations.
\end{itemize}

This tripartite structure provides both conceptual clarity and practical flexibility, allowing practitioners to tailor inference procedures to specific challenges.

\section{Review of Gradient Flows in Metric Spaces}
\label{sec:gradient-flow}

This section systematically introduces the theory of gradient flows across three nested settings: Euclidean space, general Riemannian manifolds, and infinite-dimensional spaces of probability measures (density manifolds). We begin with the familiar Euclidean case to build intuition, then extend to Riemannian manifolds to incorporate curvature effects, and finally specialize to probability density spaces (\cite{lafferty1988density}) equipped with various metrics (Wasserstein, Fisher-Rao, Stein, etc.). This progression reveals how the choice of metric fundamentally shapes the dynamics of optimization and sampling.

Our exposition emphasizes four related descriptions of gradient flows: (1) the gradient flow equation, (2) energy dissipation, (3) the evolution variational inequality (EVI), and (4) the minimizing movement (JKO) scheme (\cite{jordan1998variational}) or Moreau-Yosida regularization (\cite{lemarechal1997practical}). Their precise relationship depends on regularity, geodesic convexity, and compactness assumptions; energy decrease alone does not characterize a gradient flow.

For readers seeking more comprehensive treatments, we recommend the following references: the foundational work of Ambrosio, Gigli, and Savar\'e \cite{ambrosio2005gradient} for gradient flows in metric spaces; the reviews by Santambrogio \cite{santambrogio2017euclidean} comparing different gradient flow geometries; and the monographs by Villani \cite{villani2003topics,villani2008optimal} on optimal transport and its connections to gradient flows (\cite{mielke2023introduction}). 

We now proceed from the simplest to the most general setting, with the ultimate goal of applying these geometric flows to uncertainty quantification in dynamical systems.

\subsection{Gradient Flows in Euclidean Space}

Let \(F: \mathbb{R}^d \to \mathbb{R}\) be a continuously differentiable function.
The gradient flow of \(F\) in Euclidean space \(\mathbb{R}^d\) is described by the ordinary differential equation:
\begin{equation}
	\dot{\theta}(t) = -\nabla F(\theta(t)), \quad \theta(0) = \theta_0,
	\label{eq:euclidean_gradient_flow}
\end{equation}
where \(\nabla F\) denotes the classical gradient of \(F\) and \(\dot{\theta}(t) = \frac{\mathrm{d}\theta}{\mathrm{d}t}(t)\).

This formulation arises naturally from the steepest descent principle: at each point \(\theta\), the flow follows the direction of fastest decrease of \(F\). The energy dissipation along solutions takes the particularly elegant form:
\begin{equation}
	\frac{\mathrm{d}}{\mathrm{d}t}F(\theta(t)) = \nabla F(\theta(t)) \cdot \dot{\theta}(t) = -|\nabla F(\theta(t))|^2 \leq 0,
	\label{eq:euclidean_energy_dissipation}
\end{equation}
where \(|\cdot|\) denotes the Euclidean norm. This shows that \(F\) is strictly decreasing along the flow except at critical points where \(\nabla F = 0\).

For \(\lambda\)-convex functions \(F\) (satisfying \(F(\eta) \geq F(\theta) + \nabla F(\theta)\cdot(\eta-\theta) + \frac{\lambda}{2}|\eta-\theta|^2\)), the gradient flow can be characterized by the Euclidean evolution variational inequality:
\begin{equation}
	\frac{1}{2}\frac{\mathrm{d}}{\mathrm{d}t}|\theta(t) - \eta|^2 \leq F(\eta) - F(\theta(t)) - \frac{\lambda}{2}|\theta(t) - \eta|^2,
	\label{eq:euclidean_EVI}
\end{equation}
for all \(\eta \in \mathbb{R}^d\) and almost every \(t > 0\).

The discrete scheme in Euclidean space takes the form:
\begin{equation}
	\theta_{n+1}^\tau \in \arg\min_{\eta \in \mathbb{R}^d} \left\{ F(\eta) + \frac{1}{2\tau} |\theta_n^\tau - \eta|^2 \right\},
	\label{eq:euclidean_JKO}
\end{equation}
which is exactly the proximal point algorithm. When \(F\) is convex, the solution is unique and is characterized by the resolvent associated with the Moreau-Yosida regularization (\cite{lemarechal1997practical}):
\begin{equation}
	\theta_{n+1}^\tau = (I + \tau \nabla F)^{-1}(\theta_n^\tau).
\end{equation}

Let \((M, g)\) be a smooth Riemannian manifold and \(F: M \to \mathbb{R}\) a smooth function. 
A curve \(z: [0,T] \to M\) is called a \emph{gradient flow} of \(F\) if it satisfies the 
gradient flow equation:
\begin{equation}
	\dot{z}(t) = -\nabla_g F(z(t)) = -\mathcal{M}_{z}^{-1}\nabla F(z) \quad \text{for all } t \in [0,T],
	\label{eq:gradient_flow}
\end{equation}
where 
\(\nabla_g F\) denotes the Riemannian gradient of \(F\), defined by the relation (see \eqref{eq1} for more details)
\begin{equation}
	g(\nabla_g F, Y) = dF(Y),
\end{equation} for all vector fields \(Y\) on \(M\). Here \(\nabla F(z) = \frac{\delta F}{\delta z} \in T_{z}^{*}M\) is the first variation and \(\mathcal{M}_{z}^{-1}\) is the sharp (\(\sharp\)) or Onsager operator (\cite{mielke2016generalization}): 
\[ \mathcal{M}_{z}^{-1} : T_z^*M \to T_zM. \]

Along a solution, the gradient flow equation implies the energy dissipation identity:
\begin{equation}\label{eq4}
	\frac{\mathrm{d}}{\mathrm{d}t}F(z(t)) = \langle \nabla F(z), \dot{z}(t) \rangle = g(\nabla_g F(z(t)), \dot{z}(t)) = - \|\nabla_g F(z(t))\|^2_g, \quad  t \in [0,T],
\end{equation}
where \(\|V\|_g = \sqrt{g(V,V)}\) denotes the Riemannian norm.

When \(F\) is \(\lambda\)-convex, i.e., there exists \(\lambda \in \mathbb{R}\) such that 
\(F(\gamma(t)) \leq (1-t)F(z_1) + tF(z_2) - \frac{\lambda}{2}t(1-t)d^2(z_1,z_2)\) for all 
geodesics \(\gamma: [0,1] \to M\) connecting \(z_1\) and \(z_2\), the gradient flow is also equivalent to the evolution 
variational inequality:
\begin{equation}
	\frac{1}{2}\frac{\mathrm{d}}{\mathrm{d}t}d^2(z(t), \eta) \leq F(\eta) - F(z(t)) - \frac{\lambda}{2}d^2(z(t), \eta),
	\label{eq:EVI}
\end{equation}
for all \(\eta \in M\) and almost every \(t \in [0,T]\), where \(d\) denotes the Riemannian distance.

On a complete manifold, with \(F\) bounded below and having compact sublevel sets, the smooth gradient flow can also be obtained as the continuous limit of the discrete
minimizing movement scheme: for a fixed time step \(\tau > 0\), define the discrete sequence
\(\{z_n^\tau\}_{n=0}^N\) by the minimizing movement scheme:
\begin{equation}
	z_{n+1}^\tau \in \arg\min_{\eta \in M} \left\{ F(\eta) + \frac{1}{2\tau} d^2(z_n^\tau, \eta) \right\}, 
	\quad n = 0, 1, \dots, N-1,
	\label{eq:JKO_scheme}
\end{equation}
with \(z_0^\tau = z_0\). Defining the piecewise constant interpolant 
\(\bar{z}^\tau(t) = z_n^\tau\) for \(t \in [n\tau, (n+1)\tau)\), we have that as \(\tau \to 0\), 
the curves \(\bar{z}^\tau(t)\) converge to the unique gradient flow \(z(t)\) of \(F\) starting 
from \(z_0\).

\begin{remark}
	These related descriptions provide complementary ways to understand gradient
	flows: the gradient flow equation offers an intuitive dynamical perspective, the energy 
	dissipation emphasizes variational structure, the EVI enables robust estimates in 
	non-smooth settings, and the JKO scheme provides both existence proofs and computational 
	algorithms through discrete approximation. Moreover, gradient flows have close relation to large deviation principles connecting microscopic and macroscopic systems \cite{mielke2014relation}.
\end{remark}

\subsection{Wasserstein Gradient Flow}
\label{subsec:wasserstein-gf}

\begin{definition}[The Wasserstein space]
	Let \(\Omega \subset \mathbb{R}^d\) be a convex domain (or \(\Omega = \mathbb{R}^d\)). We consider probability measures with finite second moments:
	\[
	\mathcal{P}_2(\Omega) = \left\{\mu\in\mathcal{P}(\Omega):\int_\Omega |\theta|^2\,d\mu(\theta)<\infty\right\}.
	\]
\end{definition}
Endowed with \(W_2\) (see Definition~\ref{def:benamou-brenier}), this is the Wasserstein metric space. Its smooth, strictly positive densities admit a formal Riemannian description, which we use below. All differential identities in the following geometry subsections are understood for sufficiently regular densities and first variations, with no-flux boundary conditions on bounded smooth domains, periodic boundary conditions on a flat torus, or sufficient decay on \(\mathbb{R}^d\) to justify integration by parts. These formulas do not assert a smooth manifold structure at arbitrary, possibly singular, measures.

\noindent\emph{Tangent and cotangent spaces of the density manifold.}
At a point \(\rho \in \mathcal{P}_2(\Omega)\), the tangent space \(T_\rho\mathcal{P}_2(\Omega)\) consists of infinitesimal variations that preserve mass and positivity (to first order). Formally, consider a smooth curve \(\rho_\epsilon\) with \(\rho_0 = \rho\) and \(\left.\frac{\mathrm{d}}{\mathrm{d}\epsilon}\rho_\epsilon\right|_{\epsilon=0} = \xi\). The condition \(\int_\Omega \rho_\epsilon(\theta)\,d\theta = 1\) for all \(\epsilon\) implies
\[
\int_\Omega \xi(\theta)\,d\theta = 0.
\]
In this smooth formal description, tangent perturbations are represented as
\[
T_\rho\mathcal{P}_2(\Omega) = \left\{ \xi \in \mathcal{D}'(\Omega) : \int_\Omega \xi = 0,\ \xi = - \nabla \cdot (\rho \nabla \phi) \text{ for some } \phi \in C^\infty(\Omega)/\mathbb{R} \right\},
\]
where the potential solves a weighted Poisson equation with the stated boundary conditions. On a bounded connected domain, smooth strictly positive \(\rho\) and smooth zero-mean \(\xi\) give this representation under the usual elliptic regularity assumptions; the potential is unique up to a constant. Here \(\mathcal{D}'(\Omega)\) denotes distributions; the full metric tangent space requires the corresponding weighted Sobolev completion.

Smooth cotangent representatives can be identified with potentials modulo additive constants. More concretely, given a smooth functional \(\mathcal{F} : \mathcal{P}_2(\Omega) \to \mathbb{R}\), its first variation at \(\rho\) is the function \(\frac{\delta \mathcal{F}}{\delta \rho}\) satisfying
\[
\left.\frac{\mathrm{d}}{\mathrm{d}\epsilon} \mathcal{F}(\rho + \epsilon \xi)\right|_{\epsilon=0} = \int_\Omega \frac{\delta \mathcal{F}}{\delta \rho}(\theta) \, \xi(\theta)\,d\theta \quad \text{for all } \xi \in T_\rho\mathcal{P}_2(\Omega).
\]
Thus, the differential \(d\mathcal{F}_\rho\) is the linear functional
\[
d\mathcal{F}_\rho(\xi) = \int_\Omega \frac{\delta \mathcal{F}}{\delta \rho} \, \xi \, d\theta,
\]
and we identify the covector \(d\mathcal{F}_\rho\) with the function \(\frac{\delta \mathcal{F}}{\delta \rho}\) (defined up to a constant, because \(\int \xi = 0\)). Hence,
\[
T_\rho^*\mathcal{P}_2(\Omega) \cong C^\infty(\Omega) / \mathbb{R} \quad \text{(smooth functions modulo constants)}.
\]

\noindent\emph{The metric isomorphism.}
The Wasserstein metric provides an isomorphism between tangent and cotangent spaces via the Onsager operator \(\mathcal{M}_\rho^{-1}\):
\[
\flat : T_\rho\mathcal{P}_2(\Omega) \to T_\rho^*\mathcal{P}_2(\Omega), \quad \xi \mapsto \phi \ \text{such that } \xi = -\nabla \cdot (\rho \nabla \phi),
\]
\[
\sharp : T_\rho^*\mathcal{P}_2(\Omega) \to T_\rho\mathcal{P}_2(\Omega), \quad \phi \mapsto -\nabla \cdot (\rho \nabla \phi).
\]
In coordinates, if we write a tangent vector as \(\xi = - \nabla \cdot (\rho \nabla \phi)\), then the corresponding covector is precisely \(\phi\) (up to an additive constant). This duality is the infinite-dimensional analogue of the musical isomorphisms in finite-dimensional Riemannian geometry.

\noindent\emph{Onsager operator and Wasserstein metric tensor.}
At a density \(\rho \in \mathcal{P}_2(\Omega)\), the tangent space \(T_\rho\mathcal{P}_2(\Omega)\) consists of mass-preserving perturbations \(\xi\) satisfying \(\int_\Omega \xi(\theta)\,d\theta = 0\). The \emph{Onsager operator} (or mobility operator) \(\mathcal{M}_\rho^{-1} : T_\rho^*\mathcal{P}_2(\Omega) \to T_\rho\mathcal{P}_2(\Omega)\) is defined by
\[
\mathcal{M}_\rho^{-1}(\phi) = - \nabla \cdot (\rho \nabla \phi), \qquad \phi \in T_\rho^*\mathcal{P}_2(\Omega),
\]
where \(\phi\) is a smooth function (a co-tangent vector, interpreted as a chemical potential). The corresponding \emph{Wasserstein metric tensor} \(g_W\) is given by the duality
\[
g_W(\xi_1, \xi_2) = \langle \xi_1, \mathcal{M}_\rho \xi_2 \rangle_{L^2}=\langle \phi_1, \mathcal{M}_\rho^{-1} \phi_2 \rangle_{L^2} = \int_\Omega \nabla \phi_1 \cdot \nabla \phi_2 \; \rho\,d\theta,
\]
where \(\xi_i = - \nabla \cdot (\rho \nabla \phi_i)\) for \(i=1,2\). The operator \(\mathcal{M}_\rho\) is positive-definite on the subspace of gradients, reflecting the fact that optimal transport trajectories are driven by gradient fields.

\noindent\emph{Gradient flow definition.}
For a smooth energy functional \(\mathcal{F} : \mathcal{P}_2(\Omega) \to \mathbb{R}\), its \emph{Wasserstein gradient} \(\nabla_W \mathcal{F}(\rho) \in T_\rho\mathcal{P}_2(\Omega)\) is defined via the metric duality:
\[
g_W\!\big(\nabla_W \mathcal{F}(\rho), \xi\big) = \left. \frac{\mathrm{d}}{\mathrm{d}\epsilon} \mathcal{F}(\rho + \epsilon \xi) \right|_{\epsilon=0} = \left\langle \frac{\delta \mathcal{F}}{\delta \rho}, \xi \right\rangle_{L^2},
\]
for all tangent vectors \(\xi\). Using the Onsager operator, this yields the explicit expression
\[
\nabla_W \mathcal{F}(\rho) = \mathcal{M}_\rho^{-1}\!\left( \frac{\delta \mathcal{F}}{\delta \rho} \right) = - \nabla \cdot \left( \rho \,\nabla \frac{\delta \mathcal{F}}{\delta \rho} \right).
\]
The associated \emph{Wasserstein gradient flow} is the evolution equation
\begin{equation}
	\label{eq:wasserstein-flow}
	\boxed{
		\partial_t \rho_t = - \nabla_W \mathcal{F}(\rho_t) = \nabla \cdot \left( \rho_t \,\nabla \frac{\delta \mathcal{F}}{\delta \rho}[\rho_t] \right).}
\end{equation}

\noindent\emph{Common energy functionals and their flows (\cite{santambrogio2015optimal,santambrogio2017euclidean}).}
\begin{enumerate}
	\item Negative entropy: \(\mathcal{F}(\rho) = \int_\Omega \rho \log \rho\,d\theta\).  
	Then \(\frac{\delta \mathcal{F}}{\delta \rho} = \log \rho + 1\) and \eqref{eq:wasserstein-flow} becomes the \emph{heat equation}
	\[
	\partial_t \rho = \Delta \rho.
	\]
	
	\item Potential energy: \(\mathcal{F}(\rho) = \int_\Omega V(\theta) \rho(\theta)\,d\theta\) with \(V : \Omega \to \mathbb{R}\).  
	Here \(\frac{\delta \mathcal{F}}{\delta \rho} = V\) and \eqref{eq:wasserstein-flow} reads
	\[
	\partial_t \rho = \nabla \cdot (\rho \,\nabla V),
	\]
	a drift-dominated Fokker-Planck equation.
	
	\item Interaction energy: \(\mathcal{F}(\rho) = \frac12 \int_\Omega \int_\Omega W(\theta-\theta') \rho(\theta) \rho(\theta')\,d\theta\,d\theta'\) for a symmetric kernel \(W\).  
	Then \(\frac{\delta \mathcal{F}}{\delta \rho}(\theta) = \int_\Omega W(\theta-\theta') \rho(\theta')\,d\theta'\) and
	\[
	\partial_t \rho = \nabla \cdot \big( \rho \,\nabla (W \ast \rho) \big),
	\]
	which models aggregation or repulsion depending on \(W\).
	
	\item Free energy (combination): \(\mathcal{F}(\rho) = \int_\Omega \rho \log \rho\,d\theta + \int_\Omega V \rho\,d\theta + \frac12 \int_\Omega \int_\Omega W(\theta-\theta') \rho(\theta) \rho(\theta')\,d\theta\,d\theta'\).  
	The resulting flow
	\[
	\partial_t \rho = \nabla \cdot \big( \rho \,\nabla (\log \rho + V + W \ast \rho) \big)
	\]
	is a general aggregation-diffusion equation.
	
	\item Langevin dynamics as Wasserstein gradient flow.
	Consider the free energy functional
	\[
	\mathcal{F}(\rho) = \int_{\mathbb{R}^d} \rho(\theta) \log \rho(\theta) \, d\theta + \int_{\mathbb{R}^d} V(\theta) \rho(\theta) \, d\theta = \text{KL}(\rho\|\rho^{*})+\text{Constant},
	\]
	where \(\rho^*(\theta)=Z^{-1}e^{-V(\theta)}\), assuming \(0<Z=\int e^{-V(\theta)}d\theta<\infty\). The first term is the negative entropy and the second is the potential energy. Its first variation is
	\[
	\frac{\delta \mathcal{F}}{\delta \rho}(\theta) = \log \rho(\theta) + 1 + V(\theta).
	\]
	
	The Wasserstein gradient flow of \(\mathcal{F}\) is given by
	\[
	\partial_t \rho = \nabla \cdot \left( \rho \nabla \frac{\delta \mathcal{F}}{\delta \rho} \right) = \nabla \cdot \left( \rho \nabla (\log \rho + V) \right).
	\]
	Since \(\nabla \log \rho = \frac{\nabla \rho}{\rho}\), this simplifies to the Fokker-Planck equation
	\[
	\partial_t \rho = \Delta \rho + \nabla \cdot (\rho \nabla V).
	\]
	
	The Fokker-Planck equation above describes the evolution of the probability density for the \emph{overdamped Langevin diffusion}
	\[
	d \Theta_t = -\nabla V(\Theta_t) \, dt + \sqrt{2} \, dW_t,
	\]
	where \(W_t\) is a standard Brownian motion. Thus, the stochastic Langevin SDE induces a deterministic gradient flow in Wasserstein space for its time-marginal distributions \(\rho_t = \mathrm{Law}( \Theta_t)\).
\end{enumerate}

\subsection{Fisher-Rao Gradient Flow}

The Fisher-Rao metric, related to Hellinger geometry through the square-root density map \cite{lu2019accelerating}, is defined at a positive probability density \(\rho\) by (\cite{chen2025accelerating,trillos2023optimization,chen2024efficient,amari2016information,ay2017information}):
\begin{equation}
	g_{FR}(\xi, \eta) = \int_\Omega \frac{\xi(\theta)\eta(\theta)}{\rho(\theta)} d\theta,
	\label{eq:fisher_rao_metric}
\end{equation}
where \(\xi, \eta\) are tangent vectors at \(\rho\) (satisfying \(\int_\Omega \xi d\theta = \int_\Omega \eta d\theta = 0\)), and the corresponding metric tensor \(\mathcal{M}_{\rho}^{FR}: T_{\rho}\mathcal{P}_2(\Omega)\rightarrow T_{\rho}^{*}\mathcal{P}_2(\Omega)\) satisfies
\begin{equation}	\mathcal{M}_{\rho}^{FR}(\xi) = \frac{\xi}{\rho} \quad \text{(with zero mean constraint)},
\end{equation}
and the Onsager operator \(\mathcal{M}_{\rho}^{FR,-1}: T_{\rho}^{*}\mathcal{P}_2(\Omega)\rightarrow T_{\rho}\mathcal{P}_2(\Omega)\) satisfies
\begin{equation}
	\mathcal{M}_{\rho}^{FR,-1}(\phi) = \rho \left( \phi - \int_\Omega \phi \rho \, d\theta \right).
\end{equation}

Given an energy functional \(\mathcal{F}(\rho)\), its gradient with respect to the Fisher-Rao metric is given by:
\begin{equation}
	\nabla_{FR} \mathcal{F}(\rho) = \rho \left( \frac{\delta \mathcal{F}}{\delta \rho} - \int_\Omega \frac{\delta \mathcal{F}}{\delta \rho} \rho \, d\theta \right),
	\label{eq:fisher_rao_gradient}
\end{equation}
where \(\frac{\delta \mathcal{F}}{\delta \rho}\) denotes the first variation of \(\mathcal{F}\). The gradient flow equation 
then becomes:
\begin{equation}
	\boxed{	\partial_t \rho = -\nabla_{FR} \mathcal{F}(\rho) = -\rho \left( \frac{\delta \mathcal{F}}{\delta \rho} - \int_\Omega \frac{\delta \mathcal{F}}{\delta \rho} \rho \, d\theta \right).}
	\label{eq:fisher_rao_gradient_flow}
\end{equation}

Here we give several examples.
\begin{enumerate}
	\item For the negative entropy \(\mathcal{F}(\rho) = \int_\Omega \rho \log \rho \, d\theta\):
	\[
	\frac{\delta \mathcal{F}}{\delta \rho} = \log \rho + 1, \quad \partial_t \rho = -\rho \left(\log \rho - \int_\Omega \rho \log \rho \, d\theta \right),
	\]
	which is the \emph{replicator equation} from evolutionary dynamics.
	
	\item For a potential energy \(\mathcal{F}(\rho) = \int_\Omega V(\theta) \rho(\theta) \, d\theta\):
	\[
	\partial_t \rho = -\rho \left(V - \int_\Omega V \rho \, d\theta \right)
	\]
	which increases the relative weight of smaller values of \(V\) within the initial support.
\end{enumerate}

The Fisher-Rao metric measures local changes in relative density. Its Onsager operator multiplies the centered first variation by \(\rho\), changing density values without transporting mass between locations; the centering term enforces total mass conservation. A KL-based proximal discretization is
\[
\rho_{k+1} \in \arg\min_{\rho} \left\{ \mathcal{F}(\rho) + \frac{1}{\tau} \operatorname{KL}(\rho \mid \rho_k) \right\},
\]
whose formal small-step limit is \eqref{eq:fisher_rao_gradient_flow}, assuming the required positivity and regularity. It is a Bregman proximal step, rather than the exact minimizing movement step based on the squared Fisher-Rao geodesic distance. The two agree to second order locally because the Fisher-Rao metric is the Hessian of the KL divergence
at the diagonal \(\rho = \nu\). More precisely, for \(\rho_\varepsilon = \rho + \varepsilon \xi\) 
with \(\int_\Omega \xi d\theta = 0\), we have:
\begin{equation}
	\left.\frac{\mathrm{d}^2}{\mathrm{d}\varepsilon^2} \operatorname{KL}(\rho_\varepsilon \mid \rho)\right|_{\varepsilon=0} 
	= \int_\Omega \frac{\xi(\theta)^2}{\rho(\theta)} d\theta = g_{FR}(\xi, \xi).
\end{equation}

\begin{remark}
	These examples highlight the contrasting behaviors induced by different geometries. The Fisher-Rao and Wasserstein geometries provide complementary perspectives:
	\begin{itemize}
		\item Fisher-Rao: Local interactions, preserves support, multiplicative structure. 
		Suitable for models with birth-death processes or chemical reactions.
		\item Wasserstein: Non-local transport, can expand support, additive structure. 
		Suitable for models with spatial diffusion or optimal transport phenomena.
	\end{itemize}
	The choice between geometries depends on the physical modeling assumptions and the nature 
	of the energy functional \(\mathcal{F}(\rho)\).
\end{remark}

\begin{remark}
	For the Boltzmann entropy \(\mathcal{F}(\rho) = \int_\Omega \rho \log \rho \, d\theta\), we obtain:
	\begin{itemize}
		\item Fisher-Rao gradient flow: \(\partial_t \rho = -\rho\left(\log \rho - \int_\Omega \rho \log \rho \, d\theta\right)\) (replicator dynamics)
		\item Wasserstein gradient flow: \(\partial_t \rho = \Delta \rho\) (heat equation)
	\end{itemize}
	This illustrates how the same functional evolves differently under distinct geometric structures.
\end{remark}

\subsection{Wasserstein-Fisher-Rao Gradient Flow}

The Wasserstein-Fisher-Rao (WFR) geometry combines transport with local growth and decay (\cite{gallouet2017jko,zhu2024inclusive,liero2018optimal,sun2025variational,chizat2018unbalanced}). The unbalanced WFR, or conic Hellinger--Kantorovich, distance allows the total mass to vary. Since our unknown is a probability distribution, we use its mass-constrained, spherical version here and denote it by \(d_{WFR}\) (\cite{kondratyev2019spherical}).

The WFR metric combines the continuity equation of optimal transport with a source term:
\begin{equation}
	\partial_t \rho + \nabla \cdot (\rho v) = \rho \alpha,
\end{equation}
where \(v\) is the transport velocity and \(\alpha\) is the relative growth rate, constrained by \(\int_\Omega\alpha\rho\,d\theta=0\). The metric is defined through the following minimization:
\begin{equation}
	d_{WFR}^2(\rho_0, \rho_1) = \inf_{(\rho,v,\alpha)} \left\{ 
	\int_0^1 \int_\Omega \left[ \rho_t(\theta)|v_t(\theta)|^2 + \rho_t(\theta)\alpha_t(\theta)^2 \right] d\theta dt \right\},
\end{equation}
subject to \(\partial_t\rho+\nabla\cdot(\rho v)=\rho\alpha\), \(\rho_t\in\mathcal{P}(\Omega)\), and \(\int_\Omega\alpha_t\rho_t\,d\theta=0\) for almost every \(t\), with \(\rho(0)=\rho_0\), \(\rho(1)=\rho_1\), and the stated boundary conditions.

Let \(\mathcal{A}_\rho(\xi)=\{(v,\alpha):-\nabla\cdot(\rho v)+\rho\alpha=\xi,\ \int_\Omega\rho\alpha\,d\theta=0\}\), with the stated boundary conditions. The squared tangent norm is
\begin{equation}
	g_{WFR}(\xi,\xi)=\inf_{(v,\alpha)\in\mathcal{A}_\rho(\xi)}
	\int_\Omega\rho(\theta)\bigl(|v(\theta)|^2+\alpha(\theta)^2\bigr)\,d\theta,
\end{equation}
and the bilinear metric follows by polarization. The corresponding Onsager operator is the sum of the Wasserstein and mass-preserving Fisher-Rao operators.

The gradient flow of an energy functional \(\mathcal{F}(\rho)\) with respect to the WFR metric 
takes the form:
\begin{equation}
	\boxed{
		\partial_t \rho = \nabla \cdot \left( \rho \nabla \frac{\delta \mathcal{F}}{\delta \rho} \right) 
		- \rho \left( \frac{\delta \mathcal{F}}{\delta \rho} - \int_\Omega \frac{\delta \mathcal{F}}{\delta \rho} \rho \, d\theta \right).}
	\label{eq:wfr_gradient_flow}
\end{equation}
This equation combines the diffusion structure of Wasserstein gradient flow with 
the multiplicative structure of Fisher-Rao gradient flow.

The minimizing movement discretization uses this combined distance:
\begin{equation}
	\rho_{n+1}^\tau \in \arg\min_{\rho \in \mathcal{P}(\Omega)} \left\{ 
	\mathcal{F}(\rho) + \frac{1}{2\tau} d_{WFR}^2(\rho_n^\tau, \rho) \right\}.
	\label{eq:wfr_jko}
\end{equation}

\begin{remark}
	The parameter \(\tau\) in \eqref{eq:wfr_jko} is the time step. Relative transport and reaction strengths are set by the weights in the metric, not by \(\tau\). Adding \(W_2^2\) and KL penalties evaluated at the same pair of measures does not give the combined metric above; transport--reaction splitting is a different discretization (\cite{gallouet2017jko}).
\end{remark}

\subsection{Sobolev Gradient Flow}

Sobolev metrics provide an alternative geometric structure on probability space 
that emphasizes smoothness and regularity (\cite{bauer2016use,nurbekyan2023efficient}). Unlike the Wasserstein metric which 
measures transportation cost, or the Fisher-Rao metric which measures local relative-density changes, the metric considered here measures the \(H^{-1}\) norm of perturbations.

The Sobolev metric (specifically, the \(H^{-1}\) metric) is defined through the 
negative Sobolev norm:
\begin{equation}
	g_S(\xi, \eta) = \int_\Omega \nabla \phi_\xi \cdot \nabla \phi_\eta \, d\theta,
\end{equation}
where \(\phi_\xi\) and \(\phi_\eta\) solve the Poisson equations:
\begin{equation}
	-\Delta \phi_\xi = \xi, \quad -\Delta \phi_\eta = \eta,
\end{equation}
with Neumann or periodic boundary conditions on a bounded connected domain or a flat torus. The tangent vectors have zero integral, and the potentials are normalized to have zero integral as well.

\begin{proposition}
	The \(H^{-1}\) metric on \(\mathcal{P}(\Omega)\) can be equivalently expressed as:
	\begin{equation}
		g_S(\xi, \eta) = \int_\Omega \int_\Omega \xi(\theta) G(\theta,\theta') \eta(\theta') \, d\theta d\theta',
	\end{equation}
	where \(G(\theta,\theta')\) is the Green's function for the inverse of \(-\Delta\) on the mean-zero subspace, with the chosen boundary conditions.
\end{proposition}

The gradient of an energy functional \(\mathcal{F}(\rho)\) with respect to the Sobolev metric 
is given by:
\begin{equation}
	\nabla_S \mathcal{F}(\rho) = -\Delta \frac{\delta \mathcal{F}}{\delta \rho}.
\end{equation}
The resulting Sobolev gradient flow takes the form:
\begin{equation}
	\boxed{
		\partial_t \rho = -\nabla_S \mathcal{F}(\rho) = \Delta \frac{\delta \mathcal{F}}{\delta \rho}.}
	\label{eq:sobolev_gradient_flow}
\end{equation}

\begin{remark}
	Sobolev gradient flows have several distinctive properties:
	\begin{itemize}
		\item The order and smoothing properties of the flow depend on both the metric and the energy.
		\item The metric involves a non-local inverse Laplacian, although the resulting evolution can be a local differential equation.
		\item For the Dirichlet energy \(\mathcal{F}(\rho) = \frac{1}{2}\int_\Omega |\nabla \rho|^2 \, d\theta\), 
		the Sobolev gradient flow yields \(\partial_t \rho=-\Delta^2\rho\), with periodic conditions or \(\partial_n\rho=\partial_n\Delta\rho=0\).
		\item Sobolev metrics are particularly useful in image processing and pattern formation models.
	\end{itemize}
These equations preserve the integral under the stated boundary conditions, but they do not generally preserve nonnegativity. Their interpretation as flows of probability densities therefore requires an additional positivity argument or a positivity-constrained formulation; the biharmonic example alone does not supply one.
\end{remark}

The classical Sobolev metric can be generalized to a family of metrics based on 
elliptic differential operators. Let \(L\) be a self-adjoint operator that preserves the mean-zero subspace \(L^2_0(\Omega)\), and assume its restriction to that subspace is positive and invertible. The generalized Sobolev metric is defined by:
\begin{equation}
	g_S^L(\xi, \eta) = \int_\Omega \xi(\theta) (L^{-1} \eta)(\theta) \, d\theta,
	\label{eq:general_sobolev_metric}
\end{equation}
where \(L^{-1}\) denotes the inverse on \(L^2_0(\Omega)\).

\begin{proposition}
	The metric \(g_S^L\) defines a positive inner product on tangent vectors in its domain. When \(L = -\Delta\) with Neumann boundary conditions,
	we recover the classical \(H^{-1}\) metric. For \(L = (I - \alpha\Delta)^k\) with \(\alpha > 0\), 
	we obtain negative Sobolev metrics of order \(H^{-k}\).
\end{proposition}

Write \(P_0\phi=\phi-|\Omega|^{-1}\int_\Omega\phi\,d\theta\) for the projection onto mean-zero functions. The mass-preserving formal gradient flow is
\begin{equation}
	\boxed{
		\partial_t \rho = -L P_0\left( \frac{\delta \mathcal{F}}{\delta \rho} \right).}
	\label{eq:general_sobolev_gradient_flow}
\end{equation}

Special cases include:
\begin{itemize}
	\item \(L = -\Delta\): Classical \(H^{-1}\) gradient flow: \(\partial_t \rho = \Delta \frac{\delta \mathcal{F}}{\delta \rho}\).
	\item \(L = (I - \alpha\Delta)\): Bessel potential flow: \(\partial_t \rho = -(I - \alpha\Delta)P_0 \frac{\delta \mathcal{F}}{\delta \rho}\).
	\item \(L = (-\Delta)^k\): Higher-order diffusion: \(\partial_t \rho = -(-\Delta)^k \frac{\delta \mathcal{F}}{\delta \rho}\).
\end{itemize}

\subsection{Stein Metric and Stein Gradient Flow}
\label{subsec:stein-gf}

The Stein metric is a geometric structure on the space of probability densities induced by Stein Variational Gradient Descent (SVGD) (\cite{liu2016stein,liu2017stein,duncan2023geometry}). It relies on a reproducing kernel Hilbert space (RKHS) defined by a positive definite kernel \(k\), and captures the interaction structure of particle systems constrained by the RKHS norm.
Let \(k:\mathbb{R}^d\times\mathbb{R}^d\to\mathbb{R}\) be a sufficiently differentiable, symmetric, integrally strictly positive definite kernel, and let \(\mathcal{H}_k^d\) be the corresponding vector-valued RKHS. We use \(\mathcal{P}_k(\mathbb{R}^d)\) for smooth positive densities satisfying the integrability conditions needed below. For such a density \(\rho\), define the integral operator
\[
\mathcal{T}_{k,\rho}\phi = \int_{\mathbb{R}^d} k(\cdot,\theta') \phi(\theta') \, d\rho(\theta'), \qquad \phi\in L^2(\rho).
\]

The \emph{tangent space} of the Stein manifold at \(\rho\) is
\[
T_\rho\mathcal{P}_k(\mathbb{R}^d)=\Bigl\{\xi=-\nabla\!\cdot\!(\rho v) \; \big|\; v\in\overline{\mathcal{T}_{k,\rho}\nabla C_c^\infty(\mathbb{R}^d)}^{\mathcal{H}_k^d}\Bigr\},
\]
where the closure is taken in \(\mathcal{H}_k^d\). Smooth cotangent representatives are potentials modulo constants; the full dual uses the corresponding completion.

The metric tensor \(\mathcal{M}_{\rho}^{\text{stein}}:T_\rho\mathcal{P}_k(\mathbb{R}^d)\to T_\rho^*\mathcal{P}_k(\mathbb{R}^d)\) satisfies
\[
\xi=-\nabla\!\cdot\!\bigl(\rho\,\mathcal{T}_{k,\rho}\nabla\phi\bigr) \;\Longleftrightarrow\; \phi=\mathcal{M}_{\rho}^{\text{stein}}(\xi).
\]
The Onsager operator (or \emph{metric isomorphism}) of the Stein geometry is
\[
\mathcal{M}_{\rho}^{\text{stein},-1}(\phi)=-\nabla\!\cdot\!\bigl(\rho\,\mathcal{T}_{k,\rho}\nabla\phi\bigr), \qquad \phi\in T_\rho^*\mathcal{P}_k(\mathbb{R}^d).
\]
The corresponding \emph{Stein metric tensor} \(g_k\) is given by
\[
g_{k,\rho}(\xi_1,\xi_2)=\langle v_1, v_2 \rangle_{\mathcal{H}_k^d},
\]
where \(\xi_i=-\nabla\!\cdot\!(\rho v_i)\) with \(v_i\in\overline{\mathcal{T}_{k,\rho}\nabla C_c^\infty(\mathbb{R}^d)}^{\mathcal{H}_k^d}\).

For a smooth energy functional \(\mathcal{F}:\mathcal{P}_k(\mathbb{R}^d)\to\mathbb{R}\), its \emph{Stein gradient} is
\[
\nabla_{\text{stein}}\mathcal{F}(\rho)=\mathcal{M}_{\rho}^{\text{stein},-1}\!\left(\frac{\delta\mathcal{F}}{\delta\rho}\right)
=-\nabla\!\cdot\!\Bigl(\rho\,\mathcal{T}_{k,\rho}\nabla\frac{\delta\mathcal{F}}{\delta\rho}\Bigr).
\]
The associated \emph{Stein gradient flow} is the evolution equation
\begin{equation}
	\boxed{\,
		\partial_t\rho_t=-\nabla_{\text{stein}}\mathcal{F}(\rho_t)=\nabla\!\cdot\!\Bigl(\rho_t\,\mathcal{T}_{k,\rho_t}\nabla\frac{\delta\mathcal{F}}{\delta\rho}[\rho_t]\Bigr).
		\,}
	\label{eq:stein-flow}
\end{equation}

The Stein geometry replaces the Wasserstein velocity gradient \(\nabla\phi\) with \(\mathcal{T}_{k,\rho}\nabla\phi\). This introduces non-local particle interactions through \(k\). For a fixed kernel, the resulting continuum flow has a different Onsager operator from the Wasserstein flow; it is not its time or space discretization.

\begin{remark}
	The choice of kernel \(k\) critically influences the performance of Stein gradient flows. 
	\begin{itemize} 
		\item Smoothness: Reduced kernel smoothness can change relaxation rates (\cite{duncan2023geometry}). Kernels such as \(k(\theta,\theta')=e^{-|\theta-\theta'|^p}\), \(0<p<1\), have singular derivatives at coincident particles and fall outside the smooth setting above; their use requires additional regularization or analysis.
		\item Tails: Kernels adapted to the target distribution (e.g., weighted Matérn kernels) can improve tail exploration.
		\item Non-translation-invariance: Breaking translation-invariance can enhance convergence rates in certain settings.
	\end{itemize}
	These observations provide principled guidelines for kernel selection in SVGD-type algorithms.
\end{remark}

\noindent\emph{Kalman-Wasserstein metric: a brief note.}
While we do not develop it in detail here, the Kalman--Wasserstein metric combines ensemble covariance information with optimal transport. It preconditions the transport mobility by the covariance of the evolving distribution, rather than interpolating the Wasserstein and Fisher-Rao distances. This covariance dependence can be useful for correlated inference problems and affine-invariant particle methods. We refer to \cite{garbuno2020interacting,chen2024efficient} for its formulation and applications.

\section{Numerical Methods for Gradient Flows on Density Manifolds}
\label{sec:numerical_methods}

The preceding sections describe uncertainty quantification in mechanism-AI coupled dynamical systems through gradient flows on probability spaces. While the available geometries include Fisher-Rao, Wasserstein-Fisher-Rao, Sobolev, and Stein metrics, our numerical exposition emphasizes Wasserstein transport and closely related sampling methods. Wasserstein geometry captures the spatial structure of parameter spaces, while particle-based variational methods (\cite{wang2021particle,liu2019understanding,frank2009particle}) and proximal algorithms offer different ways to approximate distributional evolution. SVGD (\cite{liu2016stein,liu2017stein,duncan2023geometry}) is treated separately as a particle approximation of a Stein gradient flow.

It is crucial to emphasize that the dynamical system constraints, encoded through adjoint sensitivity analysis, modularly integrate into these numerical schemes. Rather than presenting algorithms with embedded adjoint computations, we treat the gradient computation via adjoint methods as a pluggable component that can be naturally incorporated into any particle-based or variational discretization. This modularity allows practitioners to adapt the numerical methods to specific differential equation constraints without altering the core geometric structure of the gradient flow.

In what follows, we explore three complementary approaches to distribution learning: time averaging through ergodic sampling (\cite{ma2015complete,brooks2011handbook,girolami2011riemann,bou2018geometric,betancourt2017conceptual,hoffman2014no,xifara2014langevin}), spatial approximation by particles (\cite{liu2016stein,liu2017stein,duncan2023geometry}), and proximal methods (JKO schemes). Their relation to a continuum gradient flow, invariant target, and finite-step approximation must be considered separately.

\subsection{Time-Averaging Methods: From Ergodicity to Practical Algorithms}
\label{subsec:time_averaging}

While particle discretizations of gradient flows offer a geometric perspective, many established sampling algorithms can be reinterpreted or derived from discretizations of these flows, particularly through the lens of time-averaging (ergodicity). This connection bridges our geometric framework with practical MCMC methods.

\subsubsection{The Ergodic Foundation of MCMC Sampling}
\label{subsubsec:ergodic_foundation}

The theoretical justification for using Markov Chain Monte Carlo (MCMC) methods to approximate posterior distributions rests on the profound concept of \emph{ergodicity} (see \cite[Chapter 5]{da2006introduction} for example). In the context of Bayesian inference, we seek to compute expectations with respect to a posterior distribution \(\pi(\theta)\):
\begin{equation}
	\label{eq:expectation}
	\mathbb{E}_{\pi}[f] = \int f(\theta)\pi(\theta)d\theta,
\end{equation}
The density also has the following representation, interpreted as an identity of distributions:
\begin{equation}
	\mathbb{E}_{\pi}[\delta(\Theta - \eta)] = \pi(\eta),
\end{equation}
The fundamental insight, dating back to von Neumann and Birkhoff, is that under ergodicity conditions, time averages converge to spatial averages. For a positive Harris recurrent Markov chain \(\{\Theta_n\}_{n\geq0}\) with invariant probability \(\pi\), and \(f\in L^1(\pi)\), the Markov chain ergodic theorem gives
\begin{equation}
	\label{eq:ergodic_theorem}
	\lim_{N\to\infty} \frac{1}{N}\sum_{n=1}^{N} f(\Theta_n) = \mathbb{E}_{\pi}[f] \quad \text{almost surely},
\end{equation}
where almost sure convergence is with respect to the law of the chain.

In the context of Wasserstein gradient flows, this ergodic framework provides the bridge between the continuous-time dynamics and their discrete-time approximations. The Langevin dynamics:
\begin{equation}
	\label{eq:langevin}
	d\Theta_t = -\nabla V(\Theta_t)dt + \sqrt{2}\,dW_t,
\end{equation}
with \(V(\theta)=-\log\pi(\theta)\), has invariant density \(\pi\) under the usual integrability and no-flux conditions. Under additional assumptions ensuring ergodicity, its time averages converge to expectations under \(\pi\). More generally, for inverse temperature \(\kappa>0\), noise amplitude \(\sqrt{2\kappa^{-1}}\) with the same drift corresponds to the invariant density proportional to \(e^{-\kappa V}\).

Metropolis-corrected discretizations preserve the target invariant measure, provided their proposal and acceptance steps are implemented consistently. Unadjusted methods generally retain a step-size bias; their asymptotic accuracy requires a separate argument.

\subsubsection{Metropolis-Hastings: Ensuring Exactness through Rejection}
\label{subsubsec:mh}

Discretizing a diffusion in time can change its invariant measure, and finite-particle approximations introduce a separate approximation error (\cite{sinho2025log}). For a Markov proposal with an evaluable transition density, the Metropolis-Hastings correction removes invariant-measure bias by enforcing detailed balance. This does not imply that every deterministic interacting-particle method admits an independent per-particle correction; its joint proposal and target must be specified. Metropolis correction is one route to exact invariance, while other methods require their own convergence analysis.

The Metropolis-Hastings algorithm (\cite{brooks2011handbook,livingstone2022barker}) provides an elegant solution: augment the proposal with a rejection step that exactly preserves the target measure through the detailed balance condition.
Given a proposal kernel \(q(\theta'|\theta)\), the acceptance probability
\begin{equation}
	\label{eq:mh_acceptance}
	\alpha(\theta, \theta') = \min\left\{1, \frac{\pi(\theta')q(\theta|\theta')}{\pi(\theta)q(\theta'|\theta)}\right\}
\end{equation}
ensures that the resulting Markov chain satisfies detailed balance with respect to \(\pi\), guaranteeing \(\pi\) as its stationary distribution.

An alternative is the Barker acceptance rule (\cite{livingstone2022barker})
\begin{equation}
	\label{eq:barker}
	\alpha_{\text{Barker}}(\theta, \theta') = \frac{\pi(\theta')q(\theta|\theta')}{\pi(\theta)q(\theta'|\theta) + \pi(\theta')q(\theta|\theta')},
\end{equation}
which also satisfies detailed balance for the stated proposal.

Ensemble-dependent MCMC methods allow a proposal to depend on the current states of several chains (\cite{del2011nonlinear,vuckovic2022nonlinear,luengo2020survey}). For a current ensemble \(X=(\theta^{(1)},\ldots,\theta^{(N)})\), consider the mixture proposal
\begin{equation}
	\label{eq:nonlinear_proposal}
	q_X(\theta'|\theta) = \frac{1}{N}\sum_{j=1}^N K(\theta'|\theta, \theta^{(j)}),
\end{equation}
where \(K\) is a proposal density. Update one coordinate \(\theta=\theta^{(i)}\) at a time, and let \(Y\) be \(X\) with that coordinate replaced by \(\theta'\). The acceptance probability is
\begin{equation}
	\label{eq:nonlinear_mh}
	\alpha_{\text{NL}}(X,Y) = \min\left\{1, \frac{\pi(\theta')q_Y(\theta|\theta')}{\pi(\theta)q_X(\theta'|\theta)}\right\}.
\end{equation}

The reverse proposal must use the proposed ensemble \(Y\) because the proposal includes the updated particle itself. With this correction, each sequential coordinate update preserves the product target \(\pi^{\otimes N}\). Ensemble dependence alone does not ensure invariance or convergence.

\begin{algorithm}
	\caption{Nonlinear Metropolis-Hastings}
	\label{alg:nonlinear_mh}
	\begin{algorithmic}
		\STATE{Initialize current ensemble \(X=(\theta^{(1)},\ldots,\theta^{(N)})\)}
		\FOR{\(t = 0, 1, 2, \ldots\)}
		\FOR{\(i = 1\) to \(N\)}
		\STATE{Set \(\theta=\theta^{(i)}\); sample \(j\sim\text{Uniform}(\{1,\ldots,N\})\)}
		\STATE{Propose \(\theta'\sim K(\cdot\mid\theta,\theta^{(j)})\)}
		\STATE{Form \(Y\) by replacing coordinate \(i\) of \(X\) with \(\theta'\)}
		\STATE{Compute \(q_X(\theta'\mid\theta)\) and \(q_Y(\theta\mid\theta')\) using \eqref{eq:nonlinear_proposal}}
		\STATE{Compute \(\alpha\) using \eqref{eq:nonlinear_mh}; set \(X\leftarrow Y\) with probability \(\alpha\), and retain \(X\) otherwise}
		\ENDFOR
		\ENDFOR
	\end{algorithmic}
\end{algorithm}

\subsubsection{Langevin-based Methods}
\label{subsubsec:langevin_based}

While random-walk proposals suffer from the curse of dimensionality, the Metropolis-Adjusted Langevin Algorithm (MALA) harnesses gradient information to guide exploration. By using the Euler-Maruyama discretization as a proposal (\cite{xifara2014langevin}):
\begin{equation}
	\theta' = \theta_k - h \nabla V(\theta_k) + \sqrt{2h} \xi_k,
\end{equation}
where \(\xi_k\sim\mathcal{N}(0,I)\), the proposal has density \(q_h(\cdot\mid\theta)=\mathcal{N}(\theta-h\nabla V(\theta),2hI)\). The Metropolis correction uses this same density in both directions to preserve the target invariant measure.

\begin{algorithm}
	\caption{Metropolis-Adjusted Langevin Algorithm (MALA)}
	\label{alg:mala}
	\begin{algorithmic}
		\STATE{Initialize \(\theta_0\), step size \(h > 0\)}
		\FOR{\(k = 0, 1, 2, \ldots\)}
		\STATE{Sample \(\xi_k \sim \mathcal{N}(0,I)\)}
		\STATE{Propose: \(\theta' \leftarrow \theta_k - h \nabla V(\theta_k) + \sqrt{2h} \xi_k\)}
		\STATE{Compute acceptance probability \(\alpha\) using \eqref{eq:mh_acceptance}}
		\STATE{Set \(\theta_{k+1} \leftarrow \begin{cases} 
				\theta' & \text{with probability } \alpha \\ 
				\theta_k & \text{otherwise} 
			\end{cases}\)}
		\ENDFOR
	\end{algorithmic}
\end{algorithm}

\subsubsection{Hamiltonian Methods}
\label{subsubsec:hamiltonian}

The limitations of overdamped dynamics become apparent in high-dimensional spaces with complex correlations. Hamiltonian Monte Carlo (HMC) addresses this by introducing auxiliary momentum variables, transforming the problem into Hamiltonian dynamics on phase space:
\begin{align}
	\label{eq:hamiltonian}
	&H(\theta,p) = V(\theta) + \frac{1}{2}p^\top M^{-1}p, \\
	&d\theta = M^{-1}p dt, \quad dp = -\nabla V(\theta) dt.
\end{align}

The leapfrog integrator provides a symplectic discretization that preserves volume and reversibility:
\begin{align}
	\label{eq:leapfrog}
	& p_{t+\epsilon/2} = p_t - \frac{\epsilon}{2} \nabla V(\theta_t), \\
	&\theta_{t+\epsilon} = \theta_t + \epsilon M^{-1} p_{t+\epsilon/2}, \\
	&p_{t+\epsilon} = p_{t+\epsilon/2} - \frac{\epsilon}{2} \nabla V(\theta_{t+\epsilon}).
\end{align}

HMC uses a lifted, conservative dynamics rather than a pure gradient flow. Momentum refreshment and Metropolis correction preserve the canonical target, while ergodicity requires additional conditions on the potential and proposal (\cite{ma2015complete,brooks2011handbook,girolami2011riemann,bou2018geometric,betancourt2017conceptual,hoffman2014no,xifara2014langevin}).

\begin{algorithm}
	\caption{Hamiltonian Monte Carlo (HMC)}
	\label{alg:hmc}
	\begin{algorithmic}
		\STATE{Initialize \(\theta_0\), step size \(\epsilon\), trajectory length \(L\), mass matrix \(M\)}
		\FOR{\(k = 0, 1, 2, \ldots\)}
		\STATE{Sample \(p_k \sim \mathcal{N}(0, M)\)}
		\STATE{\((\theta', p') \leftarrow (\theta_k, p_k)\)}
		\FOR{\(i = 1\) to \(L\)}
		\STATE{Perform leapfrog step: \((\theta', p') \leftarrow \text{Leapfrog}(\theta', p', \epsilon)\)}
		\ENDFOR
		\STATE{Reverse the proposed momentum: \(p'\leftarrow-p'\)}
		\STATE{Compute \(\alpha = \min\left\{1, \exp(-H(\theta',p') + H(\theta_k,p_k))\right\}\)}
		\STATE{Set \(\theta_{k+1}=\theta'\) with probability \(\alpha\), and \(\theta_{k+1}=\theta_k\) otherwise}
		\ENDFOR
	\end{algorithmic}
\end{algorithm}

The No-U-Turn Sampler (NUTS) (\cite{hoffman2014no}) adapts the trajectory length by building a binary tree of leapfrog steps and applying a no-U-turn stopping criterion, together with a proposal selection rule that preserves the target.

\subsubsection{Stochastic Gradient Methods: Scaling to Large Datasets}
\label{subsubsec:stochastic_gradient}

Stochastic Gradient MCMC (SGMCMC) methods use minibatch estimates \(\tilde{\nabla}V(\theta)\) of the full gradient (\cite{nemeth2021stochastic,ma2015complete}). With an unbiased gradient estimate, the Stochastic Gradient Langevin Dynamics (SGLD) update is
\begin{equation}
	\label{eq:sgld}
	\theta_{k+1} = \theta_k - h_k \tilde{\nabla} V(\theta_k) + \sqrt{2h_k} \xi_k
\end{equation}
with independent \(\xi_k\sim\mathcal{N}(0,I)\). At a fixed step size, discretization and gradient noise generally introduce bias. Asymptotic consistency requires suitable regularity, stability, and step-size assumptions; common decreasing-step conditions are \(\sum_k h_k=\infty\) and \(\sum_k h_k^2<\infty\).

Stochastic Gradient HMC (SGHMC) incorporates momentum and friction (\cite{chen2014stochastic}). In a diffusion approximation to the minibatch error, its noise-compensated form is
\begin{align}
	d\theta &= M^{-1}p dt, \\
	dp &= -\nabla V(\theta)dt-CM^{-1}p\,dt
	       +\sqrt{2B}\,dW_t^{g}+\sqrt{2(C-\widehat B)}\,dW_t,
\end{align}
where \(W^g\) and \(W\) are independent Brownian motions. The matrices \(M\) and \(C\) are positive-definite mass and friction matrices, and \(\widehat B\) estimates the effective diffusion \(B\) induced by minibatch noise; \(C-\widehat B\) must be positive semidefinite. If the minibatch gradient error has covariance \(\Sigma_g\) and step size \(h\), then \(B\approx h\Sigma_g/2\). Matching \(\widehat B=B\) balances friction and total diffusion in this approximation. The discrete algorithm below uses the actual minibatch gradient, so only the additional covariance \(2h(C-\widehat B)\) is injected. Finite step sizes and imperfect noise estimates still require bias control (\cite{nemeth2021stochastic,ma2015complete}).

\begin{algorithm}
	\caption{Stochastic Gradient HMC (SGHMC)}
	\label{alg:sghmc}
	\begin{algorithmic}
		\STATE{Initialize \(\theta_0\), \(p_0\), step size \(h\), mass matrix \(M\), friction \(C\), and noise estimate \(\widehat B\)}
		\FOR{\(k = 0, 1, 2, \ldots\)}
		\STATE{Sample minibatch, compute \(\tilde{\nabla} V(\theta_k)\)}
		\STATE{\(p_{k+1} \leftarrow p_k - h \tilde{\nabla} V(\theta_k) - h C M^{-1}p_k + \mathcal{N}(0, 2h(C-\widehat B))\)}
		\STATE{\(\theta_{k+1} \leftarrow \theta_k + h M^{-1} p_{k+1}\)}
		\ENDFOR
	\end{algorithmic}
\end{algorithm}

\subsection{Spatial Averaging Methods: Particle Approximations of Transport and Stein Flows}
\label{subsec:spatial_averaging}

The numerical solution of Wasserstein gradient flows presents unique challenges that transcend conventional PDE discretization. While time-averaging methods exploit the ergodic connection to stochastic processes, spatial averaging approaches directly discretize the gradient flow structure in the probability space itself through interacting particle systems. These particle-based variational methods provide a geometric perspective on sampling, framing inference as the optimization of particle dynamics that approximate the continuum gradient flow.

\subsubsection{Wasserstein Gradient Flows and Particle Approximations}
\label{subsubsec:wasserstein_particles}

The fundamental challenge in solving the Wasserstein gradient flow:
\begin{equation}
	\label{eq:wasserstein_flow}
	\partial_t \rho_t = \nabla \cdot \left(\rho_t \nabla \frac{\delta \mathcal{F}}{\delta \rho}(\rho_t)\right)
\end{equation}
lies in its nonlinear dependence on the evolving density \(\rho_t\). Particle methods overcome this by representing the distribution empirically:
\begin{equation}
	\label{eq:particle_approximation}
	\rho^N_t = \frac{1}{N}\sum_{i=1}^N \delta(\theta - \theta_i(t)),
\end{equation}
transforming the PDE into a system of ODEs for particle positions.

For energies whose variational derivative is defined on empirical measures, sufficiently regular interaction fields permit a mean-field justification as \(N\to\infty\). The formal particle dynamics are
\begin{equation}
	\label{eq:particle_dynamics_wasserstein}
	\frac{\mathrm{d}\theta_i}{\mathrm{d}t} = -\nabla \frac{\delta \mathcal{F}}{\delta \rho}(\rho^N_t)(\theta_i)
\end{equation}
and approximate the corresponding continuum transport flow under those assumptions. For entropy-containing energies, the derivative \(\log\rho\) cannot be evaluated at a Dirac measure; additional diffusion, density regularization, or a different geometry is required.

\subsubsection{Stein Variational Gradient Descent as Kernelized Gradient Flow}
\label{subsubsec:svgd_wasserstein}

Stein Variational Gradient Descent (SVGD) addresses a related distribution-learning problem using a kernel-dependent geometry. Its empirical dynamics remain meaningful for targets with a smooth score, even though the entropy derivative in \eqref{eq:particle_dynamics_wasserstein} is undefined on empirical measures. The continuum limit is the Stein flow introduced in Section~\ref{subsec:stein-gf}.

SVGD constrains transport velocities to a reproducing kernel Hilbert space (RKHS). The Stein identity is
\begin{equation}
	\label{eq:stein_identity}
	\mathbb{E}_{\pi}[\mathcal{A}_{\pi}\phi] = 0 \quad \text{for all } \phi \text{ in the Stein class of }\pi,
\end{equation}
where \(\mathcal{A}_{\pi}\phi = \nabla \log \pi \cdot \phi + \nabla \cdot \phi\). The Stein class imposes the integrability and boundary conditions needed for integration by parts. Evaluating the same expectation under a candidate distribution yields a discrepancy without requiring its density.
Given a target distribution \(\pi(\theta) \propto e^{-V(\theta)}\), SVGD seeks to transform an initial distribution \(q_0\) toward \(\pi\) through a sequence of smooth transformations:
\begin{equation}
	\label{eq:svgd_transform}
	T(\theta) = \theta + \epsilon \phi(\theta),
\end{equation}
where \(\phi: \mathbb{R}^d \to \mathbb{R}^d\) is a smooth vector field in an RKHS \(\mathcal{H}\) with kernel \(k(\cdot,\cdot)\).

The unnormalized RKHS descent direction \(\phi^*\) is obtained from the quadratically penalized variational problem
\begin{equation}
	\label{eq:svgd_optimization}
	\phi^* = \arg\max_{\phi \in \mathcal{H}^d} \left\{ -\frac{\mathrm{d}}{\mathrm{d}\epsilon} D_{\text{KL}}(q_{[\epsilon\phi]} \| \pi) \Big|_{\epsilon=0} -\frac12\|\phi\|_{\mathcal{H}^d}^2 \right\},
\end{equation}
where \(q_{[\epsilon\phi]}\) denotes the pushforward of \(q\) under the map \(T\).

The solution to this optimization problem yields the celebrated Stein gradient:
\begin{equation}
	\label{eq:stein_gradient}
	\phi^*(\theta) = \mathbb{E}_{\eta \sim q}[\nabla_\eta \log \pi(\eta) k(\eta,\theta) + \nabla_\eta k(\eta,\theta)].
\end{equation}
This result follows from the Stein identity and the reproducing property of RKHS. The first term drives particles toward high-probability regions of \(\pi\), while the second term acts as a repulsive force that encourages diversity among particles.
The SVGD dynamics:
\begin{equation}
	\label{eq:svgd_dynamics_wasserstein}
	\frac{\mathrm{d}\theta_i}{\mathrm{d}t} = \frac{1}{N}\sum_{j=1}^N [k(\theta_j,\theta_i)\nabla \log \pi(\theta_j) + \nabla_{\theta_j} k(\theta_j,\theta_i)]
\end{equation}
are the particle approximation of the Stein gradient flow, where:
\begin{itemize}
	\item The first term averages the target score through the kernel.
	\item The term \(\nabla_{\theta_j} k(\theta_j,\theta_i)\) provides a repulsive force that maintains particle diversity.
\end{itemize}

For a finite set of particles \(\{\theta_i\}_{i=1}^N\), the SVGD update rule becomes (\cite{liu2016stein,liu2017stein,duncan2023geometry}):
\begin{equation}
	\label{eq:svgd_update}
	\theta_i \leftarrow \theta_i + \epsilon \hat{\phi}^*(\theta_i),
\end{equation}
where the empirical Stein gradient is:
\begin{equation}
	\label{eq:empirical_stein}
	\hat{\phi}^*(\theta_i) = \frac{1}{N}\sum_{j=1}^N \left[\nabla_{\theta_j} \log \pi(\theta_j) k(\theta_j,\theta_i) + \nabla_{\theta_j} k(\theta_j,\theta_i)\right].
\end{equation}

\begin{algorithm}
	\caption{Stein Variational Gradient Descent (SVGD)}
	\label{alg:svgd}
	\begin{algorithmic}
		\STATE{Initialize particles \(\{\theta_i^0\}_{i=1}^N\), step size \(\{\epsilon_t\}\), kernel \(k(\cdot,\cdot)\)}
		\FOR{\(t = 0, 1, 2, \ldots, T-1\)}
		\FOR{\(i = 1\) to \(N\)}
		\STATE{Compute particle gradient: \(g_i \leftarrow \nabla_{\theta_i} \log \pi(\theta_i^t)\)}
		\ENDFOR
		\FOR{\(i = 1\) to \(N\)}
		\STATE{Compute Stein gradient: 
			\(\phi^*(\theta_i^t) \leftarrow \frac{1}{N}\sum_{j=1}^N \left[g_j k(\theta_j^t, \theta_i^t) + \nabla_{\theta_j} k(\theta_j^t, \theta_i^t)\right]\)}
		\STATE{Update particle: \(\theta_i^{t+1} \leftarrow \theta_i^t + \epsilon_t \phi^*(\theta_i^t)\)}
		\ENDFOR
		\ENDFOR
		\RETURN Final particles \(\{\theta_i^T\}_{i=1}^N\)
	\end{algorithmic}
\end{algorithm}

\subsubsection{Connections to Other Particle Methods}
\label{subsubsec:connections}

SVGD can be viewed as a sophisticated instance of more general particle variational inference frameworks:

\noindent\emph{Blob Methods and Regularized Optimal Transport.} 
SVGD relates to the "blob" method for solving PDEs, where particles are smoothed via kernel density estimation. For the Fokker-Planck equation (a special case of Wasserstein gradient flow) (\cite{carrillo2019blob,craig2016blob}):
\begin{equation}
	\label{eq:fokker_planck}
	\partial_t \rho = \nabla \cdot (\rho \nabla V) + \kappa^{-1} \Delta \rho,
\end{equation}
a kernel density estimate can regularize the empirical measure:
\begin{equation}
	\label{eq:blob_regularization}
	\rho^N_t(\theta) = \frac{1}{N}\sum_{i=1}^N k_h(\theta - \theta_i(t)),
\end{equation}
leading to the following score-based particle approximation:
\begin{equation}
	\label{eq:blob_dynamics}
	\frac{\mathrm{d}\theta_i}{\mathrm{d}t} = -\nabla V(\theta_i) - \kappa^{-1} \frac{\nabla \rho^N_t(\theta_i)}{\rho^N_t(\theta_i)}.
\end{equation}
This plug-in score equation is related to blob methods, but it is not generally the exact discrete gradient of a regularized entropy; variational blob schemes differentiate their chosen regularized energy consistently (\cite{carrillo2019blob,craig2016blob}).

\noindent\emph{Interacting Particle Systems as Mean-Field Limits.}
From the perspective of mean-field theory, SVGD implements an interacting particle system where the interaction is mediated through the kernel:
\begin{equation}
	\label{eq:interacting_system}
	\frac{\mathrm{d}\theta_i}{\mathrm{d}t} = \frac{1}{N}\sum_{j=1}^N \left[-\nabla V(\theta_j)k(\theta_j,\theta_i) + \nabla_{\theta_j}k(\theta_j,\theta_i)\right].
\end{equation}
This system approximates the McKean-Vlasov process associated with the gradient flow.

The spatial averaging perspective connects particle variational inference with transport and kernel-dependent gradient flows. The energy, metric, and treatment of singular empirical measures determine which continuum equation a method approximates. In particular, SVGD combines transport maps and kernel interactions to approximate the Stein flow, and its finite-particle accuracy must be assessed independently of Wasserstein-flow convergence results.

\subsection{Proximal Methods and JKO Scheme}
\label{subsec:jko}

The Jordan-Kinderlehrer-Otto (JKO) scheme (\cite{jordan1998variational}) represents a foundational approach to discretizing Wasserstein gradient flows that preserves their variational structure. Unlike particle methods that discretize in space, the JKO scheme provides a temporal discretization based on the implicit Euler method in the Wasserstein space, leading to a sequence of optimization problems that inherit the geometric properties of the continuous flow.

Given a functional $\mathcal{F}: \mathcal{P}_2(\mathbb{R}^d) \to \mathbb{R}$ and a time step $\tau > 0$, the JKO scheme defines a sequence of measures $\{\rho_k\}_{k\geq0}$ through the recursive minimization:
\begin{equation}
	\label{eq:jko_scheme}
	\rho_{k+1} \in \arg\min_{\rho \in \mathcal{P}_2(\mathbb{R}^d)} \left\{ \mathcal{F}(\rho) + \frac{1}{2\tau} W_2^2(\rho, \rho_k) \right\}.
\end{equation}
This is a proximal point method in Wasserstein space. When a minimizer exists and \(\mathcal{F}(\rho_k)<\infty\), comparison with \(\rho_k\) gives energy dissipation for any \(\tau>0\): \(\mathcal{F}(\rho_{k+1})+W_2^2(\rho_{k+1},\rho_k)/(2\tau)\leq\mathcal{F}(\rho_k)\). Existence, convergence, uniqueness, and contraction estimates require their respective compactness and convexity assumptions (\cite{han2024convergence,carrillo2024structure,carrillo2022primal}). Several computational representations are available:

\noindent\emph{Discretization in Space.}
For computational tractability, the space of probability measures is typically discretized. Common approaches include:
\begin{itemize}
	\item \textbf{Finite elements on a grid}: Represent $\rho$ as a piecewise linear function on a fixed grid
	\item \textbf{Particle methods}: Optimize particle positions for energies defined on empirical measures; entropy terms require a density representation or regularization.
	\item \textbf{Histograms}: Use a histogram representation on a fixed partition of space
\end{itemize}

\noindent\emph{Entropic Regularization.}
To overcome the computational challenges of exact optimal transport, the squared Wasserstein distance is often replaced by its entropically regularized version:
\begin{equation}
	\label{eq:entropic_regularization}
	W_{2,\epsilon}^2(\rho, \nu) = \inf_{\pi \in \Pi(\rho,\nu)} \left\{ \int \|x-y\|^2 d\pi(x,y) + \epsilon D_{\text{KL}}(\pi \| \rho \otimes \nu) \right\},
\end{equation}
where $\epsilon>0$ is a regularization parameter. This is a regularized transport cost, not a metric: its self-cost need not vanish. Sinkhorn iterations evaluate the discretized transport problem, while an outer optimization is still needed for the JKO step. Fixed regularization introduces approximation bias; convergence to the unregularized flow requires control of the regularization and optimization errors together with the time step.

\begin{algorithm}
	\caption{Entropic JKO Scheme for Fokker-Planck Equation}
	\label{alg:entropic_jko}
	\begin{algorithmic}
		\STATE{Initialize $\rho_0$, time step $\tau$, regularization $\epsilon$, number of iterations $K$}
		\FOR{$k = 0, 1, 2, \ldots, K-1$}
		\STATE{Solve the regularized JKO step:
			\(\rho_{k+1} \in \arg\min_{\rho\in\mathcal{P}_2} \left\{ \mathcal{F}(\rho) + \frac{1}{2\tau} W_{2,\epsilon}^2(\rho, \rho_k) \right\}\)}
		\STATE{Use Sinkhorn iterations within the outer optimization and control the inner and outer solution errors}
		\ENDFOR
		\RETURN $\{\rho_k\}_{k=0}^K$
	\end{algorithmic}
\end{algorithm}

\section{Open Problems and Future Directions}
\label{sec:open-problems}

The integration of AI with mechanistic modeling presents both new opportunities and challenges for uncertainty quantification in dynamical systems. While the framework developed in this paper provides a unified perspective on gradient flows in probability space under differential equation constraints, several intriguing problems remain open. These problems not only highlight the limitations of current approaches but also suggest promising avenues for future research.

A fundamental question arises in practice: given a specific problem class (e.g., a particular family of dynamical systems and observational constraints), which choice of energy functional and metric geometry yields the most effective gradient flow? The Wasserstein, Fisher-Rao, Wasserstein-Fisher-Rao, Sobolev, and Stein geometries each emphasize different aspects of probability space---transport, statistical efficiency, reaction-diffusion, smoothing, and kernel-based interaction, respectively. However, a systematic understanding of their relative strengths and weaknesses is lacking. Future work should establish a comprehensive benchmark suite that evaluates these geometries on key criteria: convergence speed, robustness to noise, scalability in high dimensions, and quality of uncertainty quantification. Such a benchmark would guide practitioners in selecting appropriate geometries and inform the development of adaptive metrics that combine the advantages of multiple geometries.

The gradient flows derived in this paper are primarily first-order methods in probability space. Yet, just as in Euclidean optimization, there is potential for acceleration by incorporating second-order information (\cite{lin2020accelerated}). In the context of constrained gradient flows, this translates to leveraging the Hessian of the energy functional or the curvature of the underlying probability manifold (\cite{wang2022accelerated,chen2025accelerating,li2019hessian}). For example, one could develop a \emph{Newton flow} in Wasserstein space, where the velocity field is preconditioned by the Hessian of the functional. However, the differential equation constraint complicates the definition and computation of such second-order structures. Investigating accelerated flows that respect dynamical constraints---potentially drawing inspiration from Nesterov acceleration, Riemannian Newton methods, or the natural gradient---could significantly improve convergence rates for complex inference problems (\cite{boumal2023introduction}).

While this paper focuses on formulation and methodology, a rigorous theoretical analysis of the proposed constrained gradient flows remains open.  Addressing these questions will not only strengthen the theoretical foundations but also provide practical guidance for algorithm design and hyperparameter selection \cite{sinho2025log}. Key questions include:
\begin{itemize}
	\item Convergence rates: Under what conditions (e.g., convexity, smoothness, identifiability) do the gradient flows converge exponentially, polynomially, or at some other rate? How does the rate depend on the chosen geometry?
	\item Computational complexity: What is the computational cost of achieving an $\epsilon$-accurate approximation of the posterior, particularly in high-dimensional parameter spaces? How do particle-based discretizations scale with dimension and sample size?
	\item Consistency: Do the discrete approximations (particle systems, JKO schemes) converge to the continuous flow as the number of particles increases and the time step decreases? Establishing such mean-field limits with differential equation constraints is non-trivial and requires careful analysis.
\end{itemize}

Recent advances in generative modeling---notably diffusion models and flow matching---offer powerful tools for sampling and density estimation \cite{lai2025principles,hosseini2025conditional,trigila2016data}. These methods are intrinsically linked to optimal transport and gradient flows. In particular, conditional optimal transport provides a natural framework for learning distributions that depend on auxiliary variables (e.g., initial conditions or control parameters). Integrating these techniques into the constrained gradient flow framework could enable more efficient simulation of high-dimensional posteriors and better handling of complex likelihoods. Future research should explore how diffusion models or flow matching can be adapted to respect differential equation constraints, potentially leading to a new class of physics-informed generative samplers.

Real-world biological and physical data often exhibit complexities beyond the noisy, sparse observations considered here. Three important directions are:
\begin{itemize}
	\item Data privacy: When data are distributed across multiple sources with privacy constraints (e.g., hospitals or research institutions), federated learning or transfer learning approaches are needed. How can constrained gradient flows be adapted to learn from such distributed data without sharing raw observations?
	\item Chaotic data: Many dynamical systems of interest are chaotic, leading to sensitive dependence on initial conditions and parameters. Quantifying uncertainty in such systems requires methods that are robust to the exponential divergence of trajectories. This may necessitate novel divergence measures or regularization techniques that account for Lyapunov exponents (\cite{yang2023optimal}).
	\item Streaming data: In many applications, data arrive sequentially over time. Online versions of the gradient flow framework are needed to update the posterior in real time without reprocessing all previous data. This could involve combining the JKO scheme with filtering techniques or developing incremental gradient flows.
\end{itemize} 

Finally, a significant challenge is extending the framework to parameters that are functions of time or space. In Bayesian inversion for partial differential equations, the unknown is often a distributed parameter field (\cite{stuart2010inverse,idier2013bayesian}). While neural network representations can, in principle, convert such problems into finite-dimensional ones, the inherent infinite-dimensional nature of the problem calls for careful treatment. Future work should focus on developing gradient flows on spaces of functions, leveraging the rich structure of reproducing kernel Hilbert spaces or Gaussian processes, while ensuring consistency with the underlying differential operators. This direction bridges the gap between classical Bayesian inversion and the flexible, geometry-aware methods proposed in this paper.

These open problems underscore the richness of the interface between optimal transport, information geometry, and dynamical systems. Addressing them will not only advance the theory of constrained gradient flows but also expand their applicability to increasingly complex and data-rich scientific domains.

\bibliographystyle{doi_only}
\bibliography{references}

\end{document}